\documentclass{article}
\usepackage{filecontents}
\usepackage{graphicx}
\usepackage{authblk}
\usepackage{verbatim}
\usepackage{array, tabularx}
\usepackage{subfigure}

\usepackage[a4paper, total={6in, 8in}]{geometry}
\usepackage{xr}
\usepackage[utf8]{inputenc}
\usepackage[T1]{fontenc}
\usepackage{amsthm}
\usepackage{amsmath}
\usepackage{amssymb}
\usepackage{siunitx}

\usepackage{url}
\usepackage{booktabs} 
\usepackage{xspace}
\usepackage{float}
\usepackage{multirow}
\usepackage{alltt}
\usepackage{color}
\usepackage{pifont}
\usepackage{makecell}
\usepackage{multicol}
\usepackage[super]{nth}

\let\origref\ref
\def\ref#1{\textnormal{\origref{#1}}}

\newcommand{\cmark}{\ding{51}}
\newcommand{\xmark}{\ding{55}}

\newtheorem*{question}{Factorial Question}

\begin{document}

\title{Collective Privacy Recovery: Data-sharing Coordination via Decentralized Artificial Intelligence}

\author[1]{Evangelos Pournaras*}
\author[2]{Mark Christopher Ballandies}
\author[2]{Stefano Bennati}
\author[3]{Chien-fei Chen}

\affil[1]{School of Computing, University of Leeds, Leeds, UK, E-mail: e.pournaras@leeds.ac.uk}
\affil[2]{Computational Social Science, ETH Zurich, Zurich, Switzerland, E-mails: mark.ballandies@ethz.ch, stefano@bennati.me}
\affil[3]{Institute for a Secure and Sustainable Environment, University of Tennessee, Knoxville, E-mail: cchen26@utk.edu, University of Tennessee, Knoxville, E-mail: cchen26@utk.edu}

\renewcommand\Authands{ and }

\maketitle

\begin{abstract}

Collective privacy loss becomes a colossal problem, an emergency for personal freedoms and democracy. But, are we prepared to handle personal data as scarce resource and collectively share data under the doctrine: as little as possible, as much as necessary? We hypothesize a significant privacy recovery if a population of individuals, the data collective, coordinates to share minimum data for running online services with the required quality. Here we show how to automate and scale-up complex collective arrangements for privacy recovery using decentralized artificial intelligence. For this, we compare for first time attitudinal, intrinsic, rewarded and coordinated data sharing in a rigorous living-lab experiment of high realism involving >27,000 real data disclosures. Using causal inference and cluster analysis, we differentiate criteria predicting privacy and five key data-sharing behaviors. Strikingly, data-sharing coordination proves to be a win-win for all: remarkable privacy recovery for people with evident costs reduction for service providers. 

\end{abstract}

%
%

\section{Introduction}\label{sec:introduction}


Control over sharing or giving access to personal data from pervasive devices, such as smartphones, turns out to be complex, involving critical decisions for privacy with impact on society. How to run data-intensive online services to improve everyday life without compromising personal values and freedoms? For instance, four apps~\cite{Sekara2021} or spatio-temporal points~\cite{Montjoye2013} are enough to identify 91.2\% and 95\% of individuals. In practice, the data-sharing doctrine `as little as possible, as much as necessary' has not yet found a systematic and scalable applicability. The quality of online services is often a result of collective data-sharing decisions made by individuals consuming these services, for instance, traffic predictions using mobility data~\cite{Montjoye2013,Bennati2022}. To achieve a minimum quality of service for a population of individuals while maximizing their privacy, a collective arrangement (i.e. coordination) of their data-sharing decisions is required to minimize both excessive and insufficient levels of data sharing~\cite{Wathieu2007,Ghosh2015,Sweeney2002}. Although a recent survey finds a 58\% of individuals willing to balance data sharing case-by-case~\cite{BCG2020}, it proves cognitively and computationally hard to achieve~\cite{Acquisti2015} even when using state-of-the-art privacy preservation techniques such as differential privacy~\cite{Jorgensen2015,Asikis2020}, secure multi-party computation~\cite{Evans2018} and k-anonymization~\cite{Sweeney2002,Meyerson2004}. The absence, failure or inefficiency in this coordination exhibit a tragedy of the (data sharing) commons, making privacy easier to compromise than quality of service. As a result, studies show that 90\% of individuals tend to give up privacy of their data, often without any added value~\cite{Acquisti2015}, although 76\% intend to protect it~\cite{BCG2012,BCG2014}. This insight is fundamental to several studies on the willingness to accept rewards for giving up privacy or willingness to pay a cost for preserving privacy~\cite{Acquisti2015,Acquisti2013,Tamir2012,Beresford2012}. Implications of giving up excessive personal data include energy-intensive and expensive data centers with unprocessed data growing faster than Moore's law predictions, stress and anxiety, algorithmic biases, discrimination, censorship and influence of election results~\cite{Jobin2019,Jones2018,Aral2019,Acquisti2015,Oulasvirta2012,Manheim2019}. Therefore, establishing a coordinated data sharing is a collective action to recover privacy with an immense impact for the environment, health, society and democracy. 

\begin{figure}[!htb]
	\centering
	\includegraphics[width=0.85\textwidth]{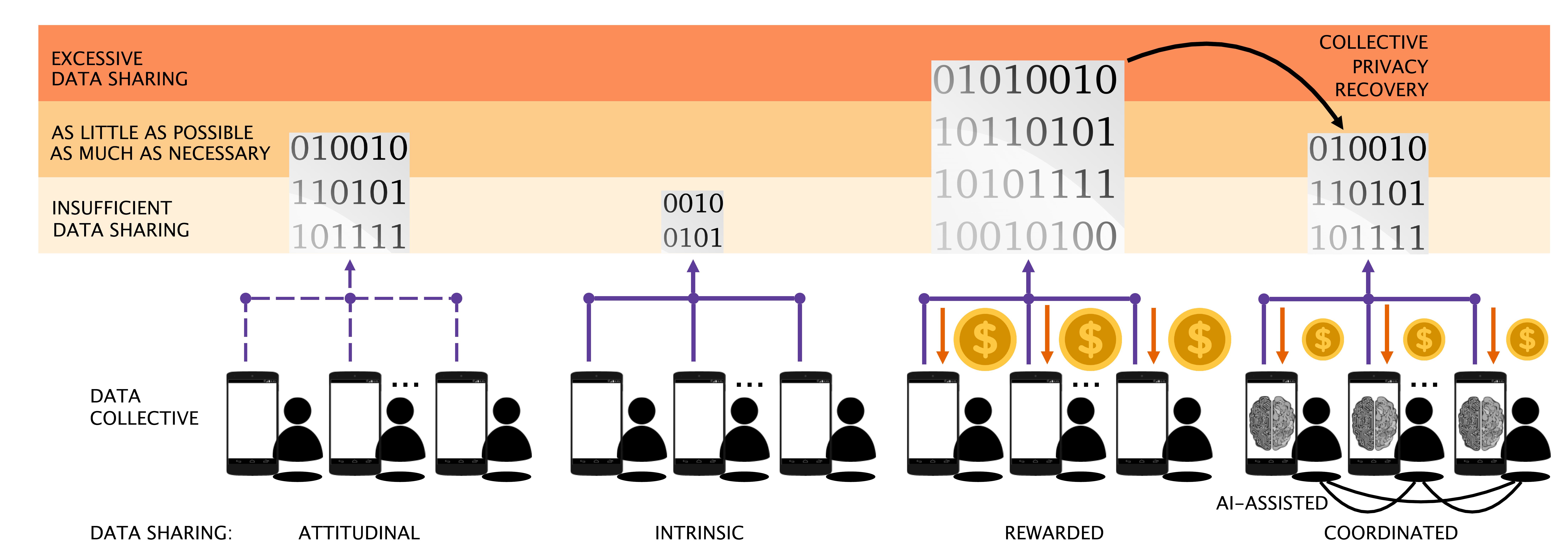}\\
	\caption{\textbf{Tragedies of data-sharing commons showing a coordination deficiency}. We hypothesize that while individuals may rationally intend to share a sufficient level of data, they end sharing intrinsically an insufficient level. If rewarded, data sharing is excessive with significant privacy loss. When coordination is introduced via a trustworthy AI-based decision-support system, significant privacy is recovered while achieving the desired quality of service. These studied hypotheses are formalized into four data-sharing conditions: (i) attitudinal, (ii) intrinsic, (iii) rewarded and (iv) coordinated.}\label{fig:conditions}
\end{figure}

\begin{figure}[!htb]
	\centering
	\includegraphics[width=0.65\textwidth]{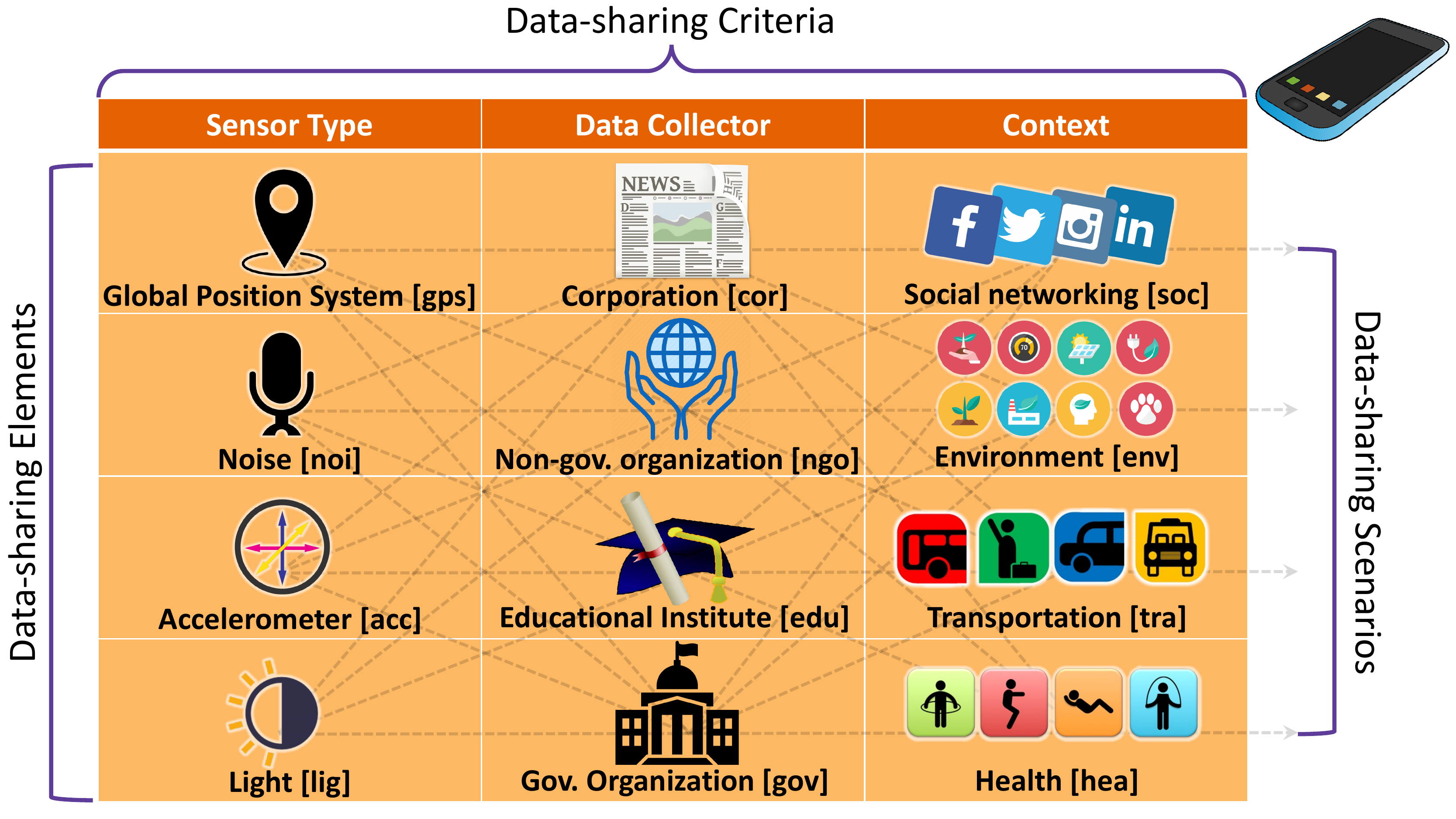}
	\caption{\textbf{The studied 4x4x4 full factorial design for smartphone data sharing}. It consists of 3 data-sharing criteria, each with 4 elements creating 64 combinations of data-sharing scenarios. Each scenario involves a choice of what data to share, to which data collector and for what purpose. The choice of the exact sensors, collectors and contexts is outlined in Section~\ref{subsubsec:entry-phase} of SI. The labels are used in the plots of this paper.}\label{fig:scenarios}
\end{figure}

\textbf{How to make coordinated data sharing feasible}. While privacy control is found essential for 82\% of individuals in an earlier study~\cite{BCG2020,Korff2014}, so is convenience for 63\%. The computational and communication load to coordinate data-sharing decisions at scale is overwhelming for humans alone. Instead, a scalable decision support can be provided by interactive personal assistants using cooperative artificial intelligence (AI) to cope with such complexity~\cite{Dafoe2021}. These assistants can run on (mobile) devices of individuals who form a community (i.e. data collective) to consume an online service that relies on data they share as a result of a collective arrangement. In practice, the remote personal assistants interact in the background to coordinate \emph{how much and what data to share, to which data collector and for what purpose} (see Fig.~\ref{fig:conditions} and~\ref{fig:scenarios}). These multi-agent interactions and calculations self-organize into fully decentralized unsupervised learning process~\cite{Pournaras2018} that optimizes data-sharing efficiency: maximizing quality of service and minimizing privacy cost. Compared to other AI approaches for personalized privacy assistants~\cite{Lippi2019} applied to legal document analytics~\cite{Joshi2016} and pervasive devices~\cite{Anupam2018}, this decision-support system is itself privacy-preserving and does not rely on any centralized third party (UNESCO IRCAI outstanding~\cite{CollectiveLearningIRCAI}). Therefore, the interactive personal assistants are trustworthy by design to serve as the privacy enabler of the data collective. This comes in stark contrast to the mainstream use of supervised AI algorithms that often require large concentrations of sensitive personal data for training~\cite{Montjoye2017,Manheim2019,Jobin2019}. The proposed decision-support system can also operate as a trustworthy collective access control to local data by federated learning algorithms to train models in a privacy-preserving way~\cite{Kaissis2020,Montjoye2017}. 

\textbf{Hypotheses for understanding data-sharing conditions}. The overarching aim of this study is to assess the capacity of this novel AI-based system to steer the data collective into more efficient and privacy-preserving trajectories for data sharing. Fig.~\ref{fig:conditions} illustrates the main studied hypotheses. These hypotheses are formalized into four experimental conditions for data sharing under repeated measures (within-subjects design). They are rigorously compared with each other under high realism in a novel living-lab experiment, see Section~\ref{subsec:experiment}, Fig.~\ref{fig:experimental-design} and~\ref{fig:app-core}. Over 27,403 high-quality records of real data-disclosure decisions are collected by a novel platform developed for this purpose (see Section~\ref{subsec:infrastructure}). It encompasses a smartphone app, a server to collect experimental data as well as a web portal with which the involved data collectors can access the shared data according to the privileges that participants give. The four studied experimental conditions shown in Fig.~\ref{fig:scenarios} are the following: 

\noindent \textbf{1. Attitudinal data sharing} assesses how privacy-sensitive individuals perceive each of the $ 3 \text{ criteria} \cdot  4 \text{ elements/criterion} = 12$ data-sharing elements, see Table~\ref{table:entry-phase}, Questions B.9 to B.12 in Supplementary Information (SI). 

\noindent \textbf{2. Intrinsic data sharing} assesses actual decisions made for voluntarily data sharing (without rewards) in a complete factorial design of $4 \text{ sensors} \cdot 4 \text{ collectors} \cdot 4 \text{ contexts} = 64$ data-sharing scenarios. 

\noindent \textbf{3. Rewarded data sharing} introduces an accumulated privacy-reward balance that individuals initially influence with their choices over the 64 data-sharing scenarios (see Fig.~\ref{fig:app-core}a). The built up balance can be further calibrated by making on-demand and repeated (unlimited within 24 hours) choices among the 64 data-sharing scenarios retrieved automatically. Each retrieved scenario is calculated to improve the individual's choice: privacy or rewards, see Fig.~\ref{fig:app-core}b. To account for threats to validity and trace any order effects, this experimental condition is repeated twice ($2 \cdot 24$ hours) by clearing the privacy-reward balance and collecting new data from sensors to share (Fig.~\ref{fig:experimental-design}). To challenge privacy preservation, the rewards are personalized by inflating and deflating the amounts based on each individual's privacy perception derived from attitudinal data sharing, see Section~\ref{sec:model} in SI. This design choice is also expected to engage participants more effectively by rewarding the data-sharing scenarios fairly, according to their personal values~\cite{Acquisti2015}, while discouraging dropouts. 

\noindent \textbf{4. Coordinated data sharing} relies on the AI-based personal assistants. They use the intrinsic and rewarded data-sharing levels as discrete options to choose from (ex-post condition). Each assistant makes an optimized choice among these that recovers the collective privacy loss of the rewarded data sharing, while reducing the \emph{mismatch} (discrepancy/fitness measure) between the shared and the required data by a service provider. This is a quality-of-service indicator that finds general applicability in adaptive sensor selection and flexible data fusion for several smart city and industrial applications~\cite{Ding2019,Eick2020,Kim2021}. Matching can also be applied by a coordinated data collective to preserve $k$-anonymity in a bottom-up way, i.e. no more than $k$ individuals share any combination of personal data~\cite{Sweeney2002,Meyerson2004,Wahida2021}. 

\textbf{Smartphone sensor data play a pivotal role on privacy}. This paper studies sharing of smartphone sensor data with five discrete choices to choose from (uniform sampling of 100\% to 0\% of sensor data with a step of 25\%), see Fig.~\ref{fig:app-core}b. These choices are applied to the total sensor data collected with a fixed frequency of 30 sec (100\% of data). This is a simple and general discrete-choice model that serves complexity of the experiment. It can be extended to more complex spatio-temporal models as discussed in Section~\ref{sec:discussion}. The study of smartphone sensor data is particularly impactful for both privacy and quality of online services. Sensor fusion has a paramount role in applications of smart homes, grids and transportation~\cite{Ding2019}. There is evidence that smartphone app developers delegate privacy to end-users as the former face challenges in providing privacy solutions at the design and implementation phase~\cite{Balebako2014}. In practice though, it is the powerful data intermediaries that leverage the terms of data-sharing agreements~\cite{BCG2020,Sekara2021}. Sharing smartphone sensor data can be regulated via privacy-protection mechanisms with a natural utility-driven interpretation (buy-sell) such as differential privacy~\cite{Ghosh2015}. Given the symbiotic relationship of individuals with their smartphones, capturing high-dimensional and diverse sensor data for different application scenarios, the study comes with a universal scope on privacy. 

\textbf{A novel approach to understanding data-sharing decisions}. The performed living-lab experiment is the first of its kind: (i) It brings together all four data-sharing conditions for comparison, including the novel one of coordinated data sharing. This is distinguished from earlier survey studies and empirical observations focusing on the two dimensions of intentions vs. behavior that comprise the privacy paradox~\cite{Christin2013,Norberg2007}. (ii) The experimental design uses mixed modalities to achieve rigor within a controlled lab environment as well as realism, scale and external validity by tracing behavior out of the lab using a smartphone platform developed for this purpose (see Section~\ref{subsec:infrastructure}). (iii) The 4x4x4 factorial design results in 64 data-sharing scenarios (see Fig.~\ref{fig:scenarios}). They involve the three data-sharing criteria that model the involved trust (data collectors) and risks (data type and context), and they are the ones that explain malleable data-sharing behaviors~\cite{Acquisti2013,Acquisti2015,Adams2001}. This large spectrum comes in contrast to earlier experiments and field tests made within a context and involving a specific data-sharing scenario such as online social lending~\cite{Bohme2011}, crowdfunding~\cite{Burtch2015} and commerce~\cite{Beresford2012,Acquisti2013,Tsai2011,Ballandies2022}.

\section{Results}\label{sec:results} 

\textbf{Three key results are illustrated in this paper}: (i) Coordinated data sharing is efficient--it recovers privacy for people and reduces costs for service providers by accessing less but better quality of data. (ii) Data collector and context are the most important criteria with which individuals makes data-sharing choices. For rewarded choices with privacy loss though, the type of shared data becomes the most important criterion. (iii) Individuals exhibit five key group-behavior changes from intrinsic to rewarded data sharing. They are stable, yet reinforcing.

\subsection{Coordinated data sharing recovers privacy and lowers costs}\label{subsec:privacy-data-sharing-scenarios}

\begin{figure}[!htb]
	\centering
	\subfigure[Privacy ($P$, mean normalized data-sharing level) sorted from lowest to highest according to intrinsic data sharing.]{\includegraphics[width=0.9\textwidth]{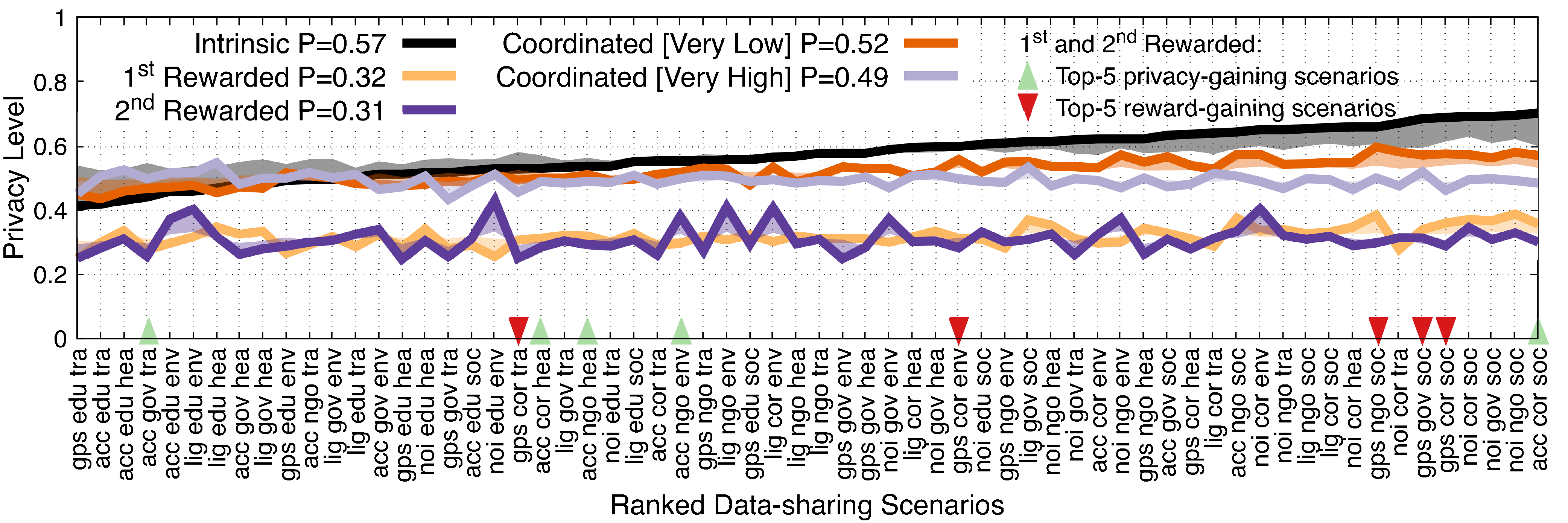}}
	\subfigure[Data-sharing mismatch ($\varepsilon$, absolute error of standardized signals) between three data-sharing conditions and the privacy-preservation goal signals of very high and very low. Values are sorted from lowest to highest mismatch according to coordinated data sharing.]{\includegraphics[width=0.9\textwidth]{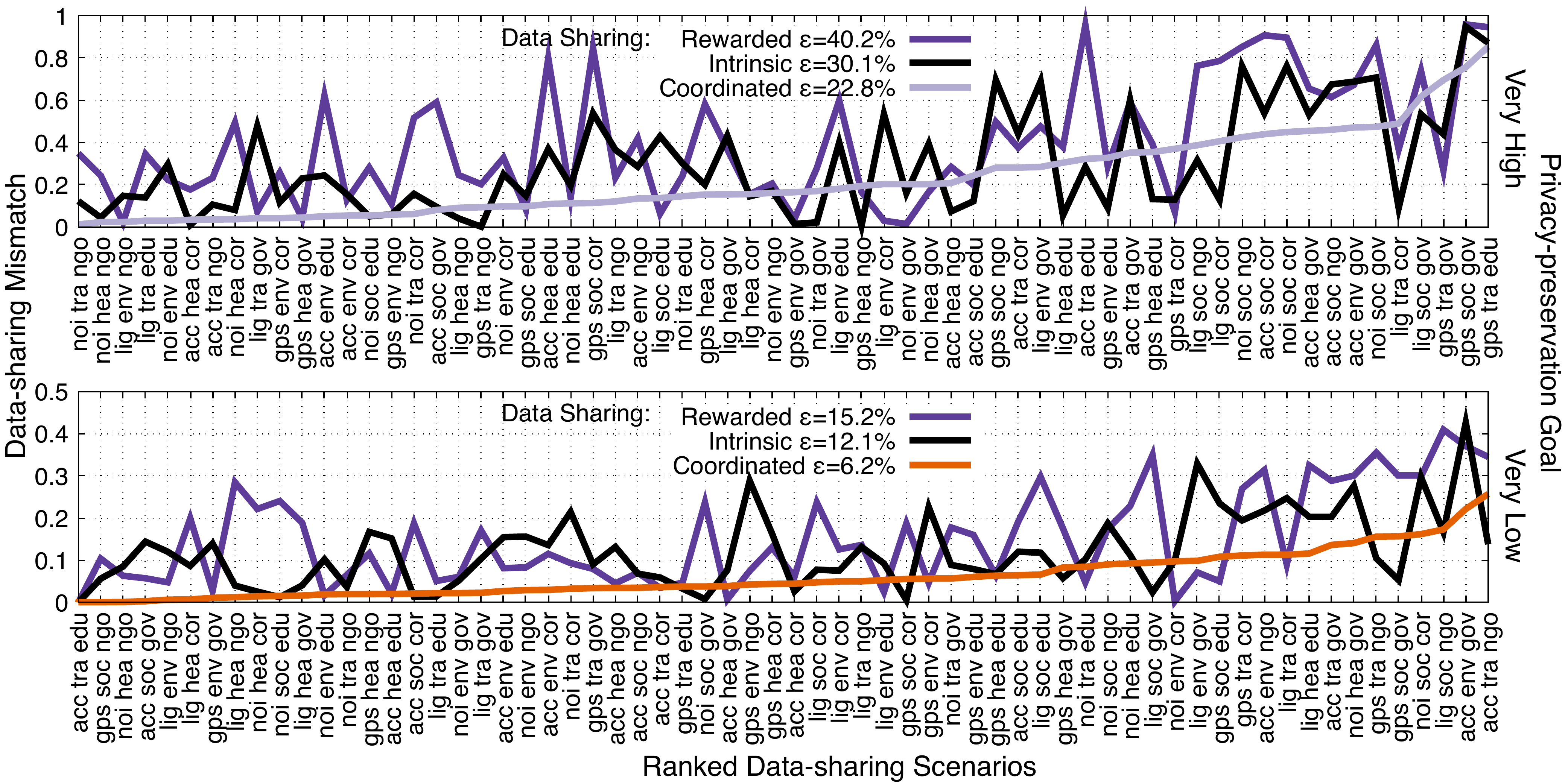}}
	\caption{\textbf{Coordinated data sharing shows higher efficiency than intrinsic and rewarded data sharing}. Privacy and mismatch for the 64 data-sharing scenarios. }\label{fig:privacy-mismatch-scenarios}
\end{figure}

\begin{figure}[!htb]
	\centering
	\subfigure[Privacy level.]{\includegraphics[width=0.52\textwidth]{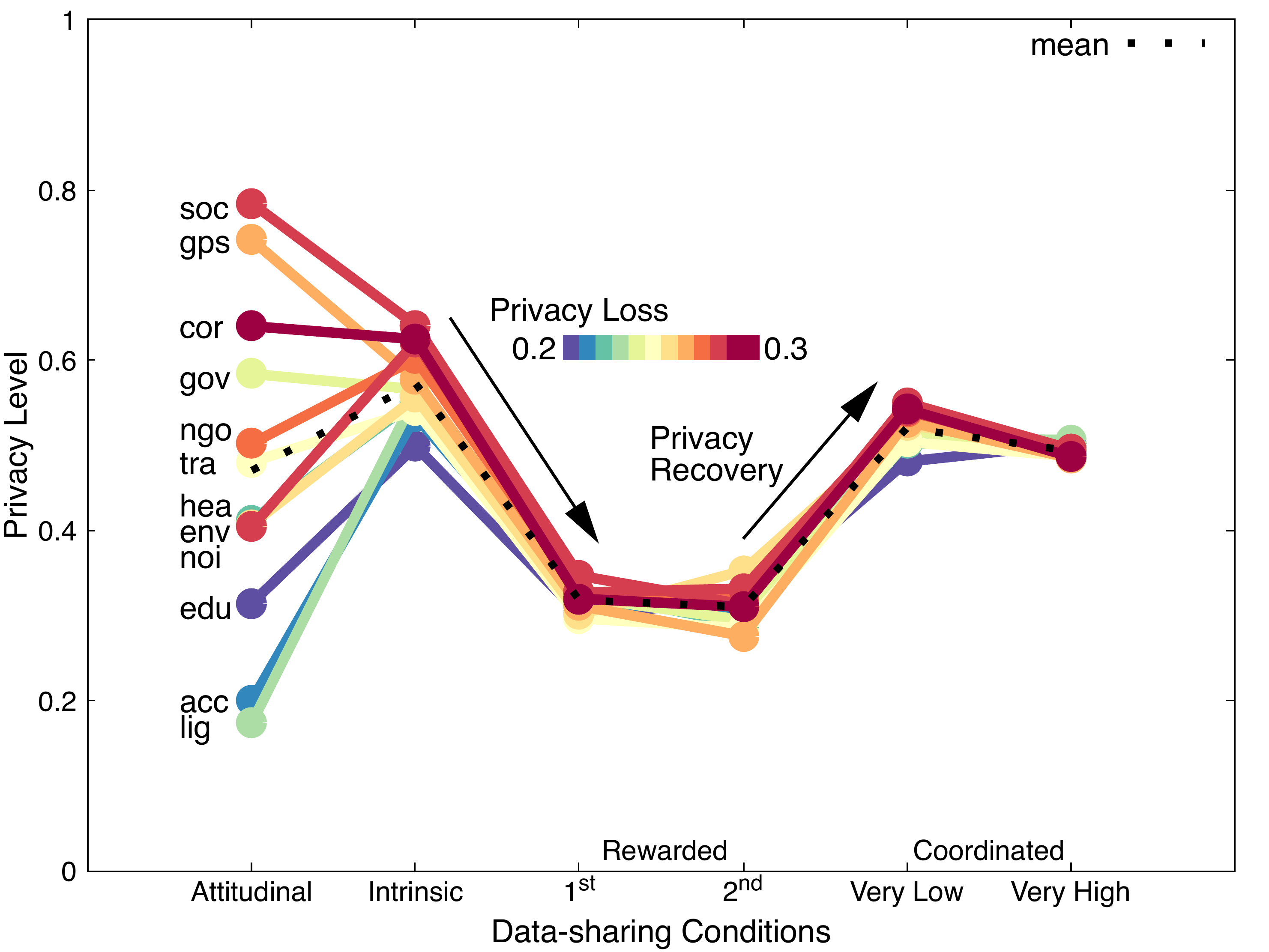}}
	\subfigure[Data-sharing mismatch between three data-sharing conditions and the privacy-preservation goal signals of very low (left) and very high (right).]{\includegraphics[width=0.4\textwidth]{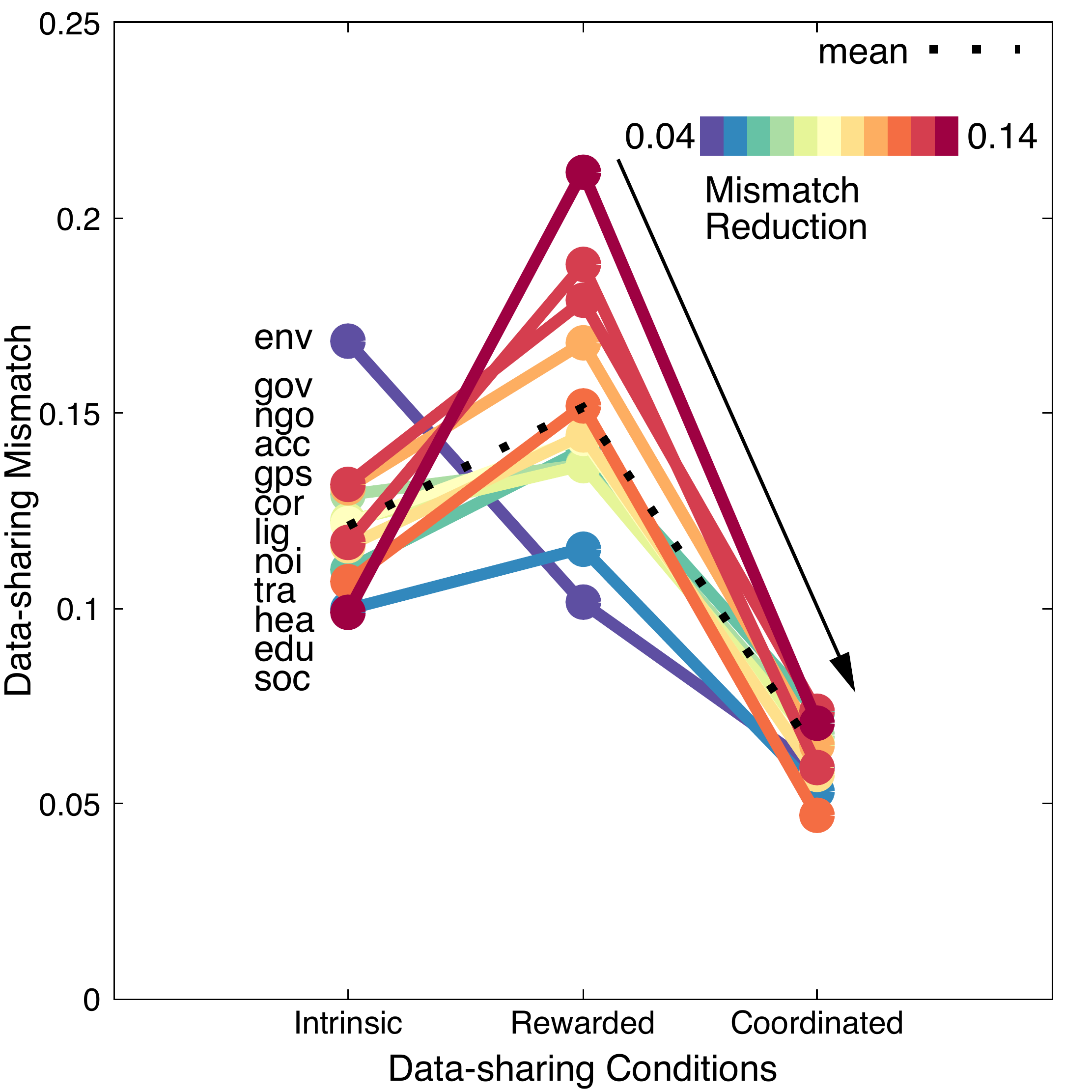}
	\includegraphics[width=0.4\textwidth]{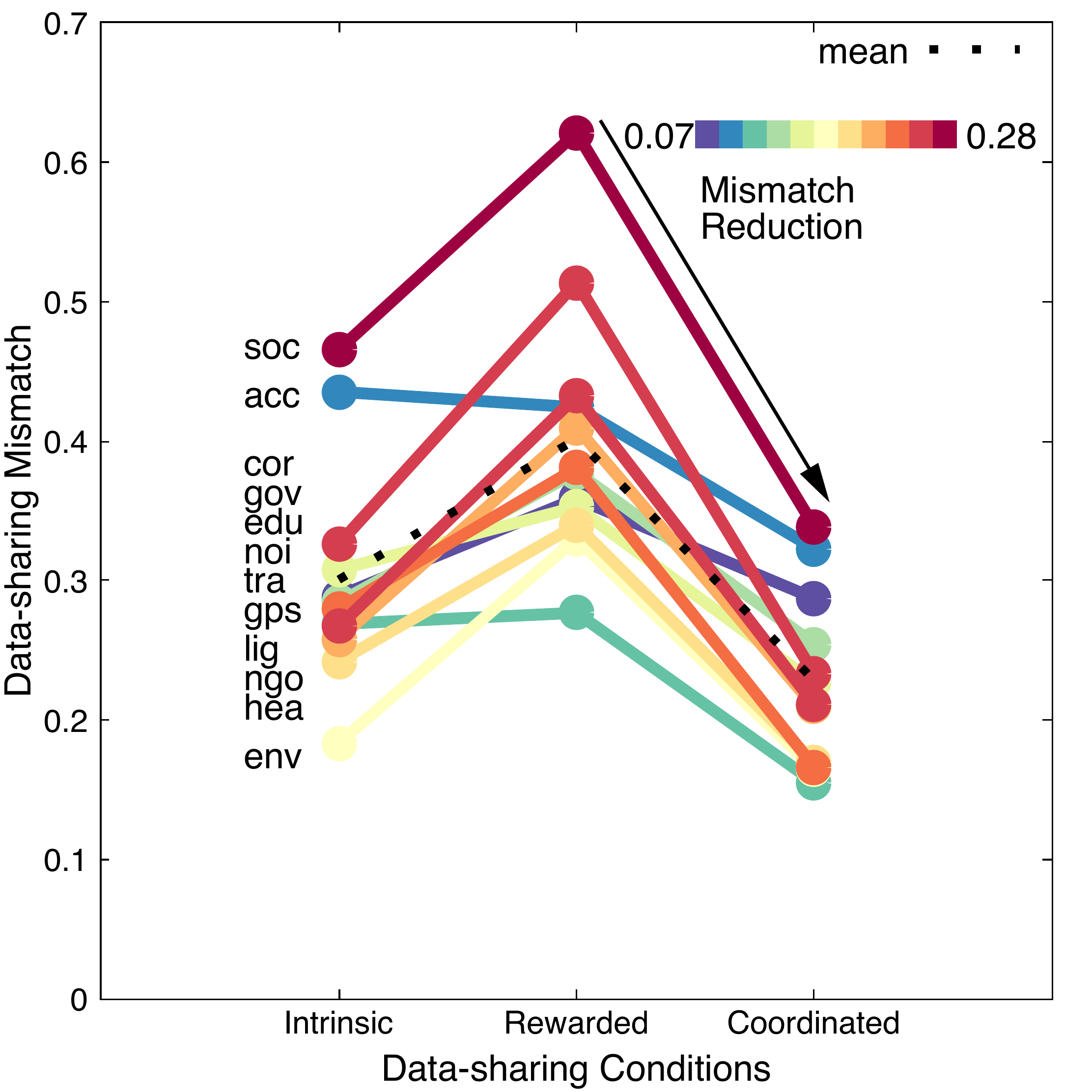}}
	\caption{\textbf{Privacy and data-sharing mismatch level of the different sensors, collectors and contexts under intrinsic, rewarded and coordinated data sharing}. The privacy level of attitudinal data sharing is also shown. The 12 colored lines are ranked according to the privacy loss (intrinsic - \nth{1} rewarded data sharing) and mismatch reduction (\nth{1} rewarded data sharing - coordinated).}\label{fig:privacy-mismatch-elements}
\end{figure}

The privacy level and data-sharing quality (mismatch) are shown in Fig.~\ref{fig:privacy-mismatch-scenarios} for the 64 data-sharing scenarios and the different experimental conditions. Fig.~\ref{fig:privacy-mismatch-elements} aggregates these measurements for each of the four sensors, data collectors and contexts. The shaded areas in Fig.~\ref{fig:privacy-mismatch-scenarios}a illustrate the expected privacy level. It is derived by the mean privacy level of the sensor, collector and context that comprise each data-sharing scenario (see Section~\ref{subsec:privacy-level-reinforcement} for exact calculations).

\begin{figure}[!htb]
	\centering
	\includegraphics[width=0.9\textwidth]{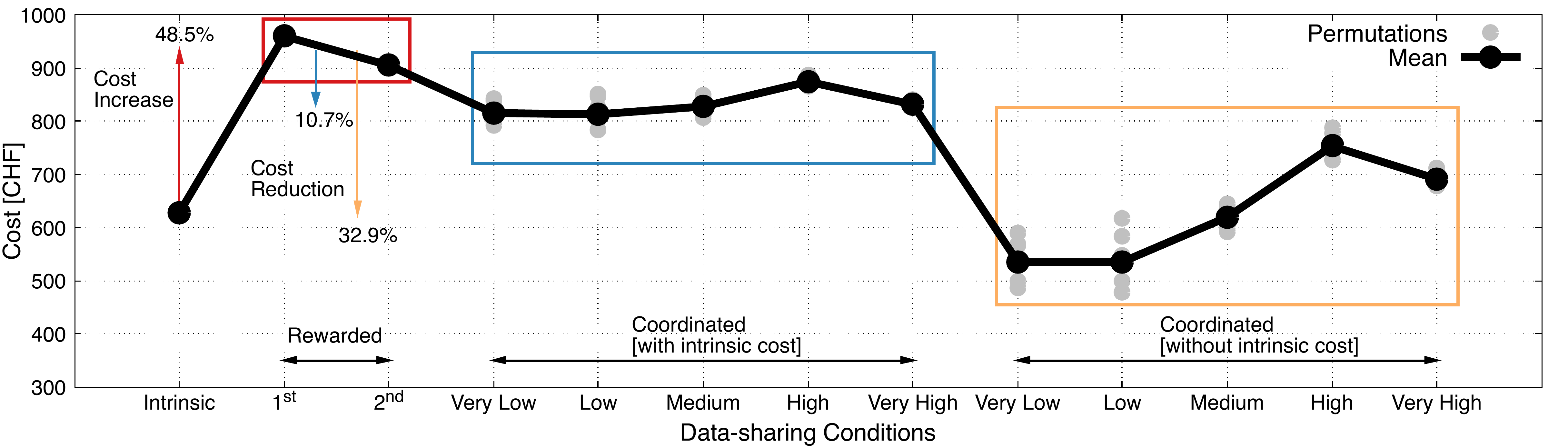}
	\caption{\textbf{Coordinated data sharing reduces data-collection cost 10.7\%-32.9\% compared to rewarded data sharing. This cost is comparable to intrinsic data sharing.} Rewarded data sharing results in excessive data with 48.5\% higher cost than intrinsic data sharing. Coordinated data sharing is calculated with and without the intrinsic cost. The gray points are random permutations of the initial conditions in the optimization process.}\label{fig:costs}
\end{figure}

The key observations are summarized as follows: (i)  Coordinated data sharing results in significant privacy recovery (Fig.~\ref{fig:privacy-mismatch-scenarios}a and~\ref{fig:privacy-mismatch-elements}a) as well as more efficient data sharing (Fig.~\ref{fig:privacy-mismatch-scenarios}b and~\ref{fig:privacy-mismatch-elements}b) at a lower cost for service providers (Fig.~\ref{fig:costs}). (ii) Intrinsic data sharing positively correlates to attitudinal data sharing but has a narrower range (Fig.~\ref{fig:privacy-mismatch-elements}a). (iii) Consecutive rewarded data sharing results in significant (and similar) privacy loss via, though, different data-sharing choices (Fig.~\ref{fig:privacy-mismatch-scenarios}a and~\ref{fig:privacy-mismatch-elements}a). (iv) The privacy loss, rather than the privacy level, under rewarded data sharing is correlated to the perceived privacy sensitivity (Fig.~\ref{fig:privacy-mismatch-elements}a). (v) Individuals improve their privacy by sharing data with lower privacy sensitivity than when improving rewards,	 while they keep sharing data to privacy-intrusive collectors under privacy-intrusive contexts (Fig.~\ref{fig:privacy-mismatch-scenarios}a).

\noindent \textbf{Coordinated data sharing for efficiency and privacy recovery}. Fig.~\ref{fig:privacy-mismatch-scenarios}b illustrates the mismatch (absolute error) between a privacy-goal signal (very low and very high privacy preservation) and the aggregated data-sharing choices made via the AI approach (both standardized). Coordinated data sharing has a lower average mismatch than intrinsic and rewarded data sharing for both goal signals: 22.8\%<30.1\%<40.2\% for very high and 6.2\%<12.1\%<15.2\% for very low privacy preservation respectively. With the very high privacy-preservation goal, matching is harder as there is mainly one data-sharing plan (intrinsic), out of three ones to choose from, containing data-sharing choices with high privacy preservation. On the contrary, with the very low privacy-preservation goal, mismatch is minimal by combining data-sharing plans from both the \nth{1} and \nth{2} rewarded data-sharing conditions. This trend is also confirmed in the other three privacy-goal signals (see Fig.~\ref{fig:privacy-mismatch-scenarios-L-M-H}, Section~\ref{sec:mismatch} of SI). For the very low and very high privacy-preservation goal, health (4.7\%, 16.5\%) and noise (5.7\%, 16.6\%) show a low mismatch on average, while government (7.3\%, 32.3\%) and social networking (7.1\%, 33.8\%) show a high one, see Fig.~\ref{fig:privacy-mismatch-elements}b. Via coordinated data sharing, social networking shows the highest mismatch reduction of 66.6\% and 45.5\% under the very low and very high privacy privacy-preservation goals. The overall average privacy recovery from rewarded to coordinated data sharing is 77\%. These results demonstrate the unprecedented potential of coordinated data sharing to protect privacy, while retaining a data-sharing efficiency (see also Fig.~\ref{fig:privacy-cost-valuations}, Section~\ref{sec:privacy-valuation schemes} of SI illustrating different privacy-recovery valuations). Coordinated data sharing operates close to intrinsic data sharing with a minor (but significant: $t(63)=9.64, p=\num{1.00e-5}$ for the very low and $t(63)=7.81, p=\num{1.00e-5}$ for the very high privacy-preservation goal) additional privacy sacrifice that is a benefit for data-sharing efficiency and as a result, the data collective as a whole. 

\noindent \textbf{Coordinated data sharing reduces data-collection costs}. Fig.~\ref{fig:costs} shows the incurred data-collection costs. The monetary cost of the \nth{1} and \nth{2} rewarded data sharing for data collectors is 960.18 CHF and 905.14 CHF respectively. This cost is higher than the monetary value of the data shared intrinsically, which is 628.22 CHF. Strikingly, the cost of coordinated data sharing is on average 832.56 CHF ($\sigma=15.93$), which is on average 10.7\% lower than the rewarded data sharing. These costs include the monetary value of intrinsic data sharing. If this value is excluded assuming that this data is shared for free (as happened in the experiment), the cost drops further down to 626.77 CHF, which is on average 32.9\% lower than rewarded data sharing. It is remarkable that the monetary value of coordinated data sharing is similar to the one of intrinsic, however, it yields data of higher utility for service providers. As a result, coordinated data sharing is a win-win for all: lower data collection costs for service providers, higher quality of service via improved data-sharing efficiency and significant privacy recovery for the participants of the data collective. 

\noindent \textbf{Attitudinal-intrinsic data sharing}. Privacy preservation under intrinsic data sharing is 21.7\% higher than the perceived privacy (Fig.~\ref{fig:privacy-mismatch-elements}a). While this difference is not significant ($t(11)=-2.07, p=0.06$), the privacy levels between the 12 elements of attitudinal and intrinsic data sharing are positively correlated ($R=0.63, t(10)=2.54, p=0.029$), despite the significant drop of 95.3\% in the dispersion (variance). This result shows that data sharing operates in a narrower decision space than the perceived privacy. Social networking (0.78, 0.64) and corporation (0.64, 0.62) come with both high privacy sensitivity and preservation, while education (0.31, 0.5) and accelerometer (0.2, 0.53) show low privacy sensitivity and preservation. 

\noindent \textbf{Intrinsic-rewarded data sharing}. Under the two rewarded data-sharing conditions, participants clearly give up privacy by 44\% ($t(63)=-31.35, p=\num{1.00e-5}$) and 45.9\% ($t(63)=-25.49, p=\num{1.00e-5}$) respectively (Fig.~\ref{fig:privacy-mismatch-scenarios}a, see also Fig.~\ref{fig:privacy-loss-groups}a and~\ref{fig:privacy-loss-groups}b in Section~\ref{sec:privacy-loss-groups} of SI). The privacy level of intrinsic data sharing for the different sensors, collectors and contexts is correlated to the one of the \nth{1} rewarded data sharing ($R=0.53, t(62)=4.99, p=\num{5.00e-6}$) but not to the one of the \nth{2} rewarded data sharing ($R=0.12, t(62)=0.94, p=0.79$). Consecutive rewarded data sharing results in equivalent privacy preservation ($t(63)=-1.22, p=0.23$); nevertheless, this effect appears via different choices made within the data-sharing scenarios ($R=0.033, t(62)=0.26, p=0.79$). 

\noindent \textbf{Attitudinal-rewarded data sharing}. Rewarded participants sacrifice privacy by 32.4\% ($t(11)=2.72, p=0.013$) and 34\% ($t(11)=2.85, p=0.009$) compared to attitudinal data sharing (Fig.~\ref{fig:privacy-mismatch-scenarios}a). The privacy level under the two rewarded data-sharing conditions is not correlated to the perceived privacy sensitivity (attitudinal) of the different sensors, collectors and contexts ($R=0.36, t(10)=1.22, p=0.24$ and $R=-0.39, t(10)=1.53, p=0.15$  in Fig.~\ref{fig:privacy-mismatch-elements}a). Striking, though, it is the privacy loss (intrinsic-rewarded data sharing) that correlates to attitudinal data sharing ($R=0.64, t(10)=2.64, p=0.025$ , $R=0.77, t(10)=3.82, p=0.0033$). 

\noindent \textbf{Which data-sharing scenarios improve privacy and rewards}? Under rewards, data-sharing scenarios are automatically retrieved to fulfill participants' goal, i.e. data-sharing options with the highest improvement of privacy or rewards, see Fig.~\ref{fig:app-core}. Fig.~\ref{fig:privacy-mismatch-scenarios}a marks the top-5 scenarios that result in the highest mean privacy and reward gain (all ranked scenarios are presented in Fig.~\ref{fig:privacy-reward-gain-scenarios} and Table~\ref{tab:privacy-reward-gain-elements} of SI). The most highly privacy-gaining scenarios involve non-privacy-sensitive sensor data such as accelerometer, which are shared though with privacy-intrusive data collectors and contexts such as social networking and corporation. In contrast, the most highly reward-gaining scenarios involve privacy-sensitive sensor data such as GPS, which are also shared with the privacy-intrusive data collectors and context of social networking and corporations. These observations reveal the following: Individuals improve privacy or rewards by sharing data under privacy-sensitive contexts to privacy-intrusive collectors. Nonetheless, compared to improving rewards, individuals change to sharing data with lower privacy sensitivity when improving their privacy. 

\subsection{Rewarded individuals better distinguish data than collectors/contexts}\label{subsec:causal-inference}

\begin{figure}[!htb]
	\centering
	\subfigure[Coefficients of the different regression models. The type of sensor data contributes positively to privacy preservation and rewards gain. Data collectors and context contribute negatively to privacy preservation and rewards gain.] {\includegraphics[width=0.9\textwidth]{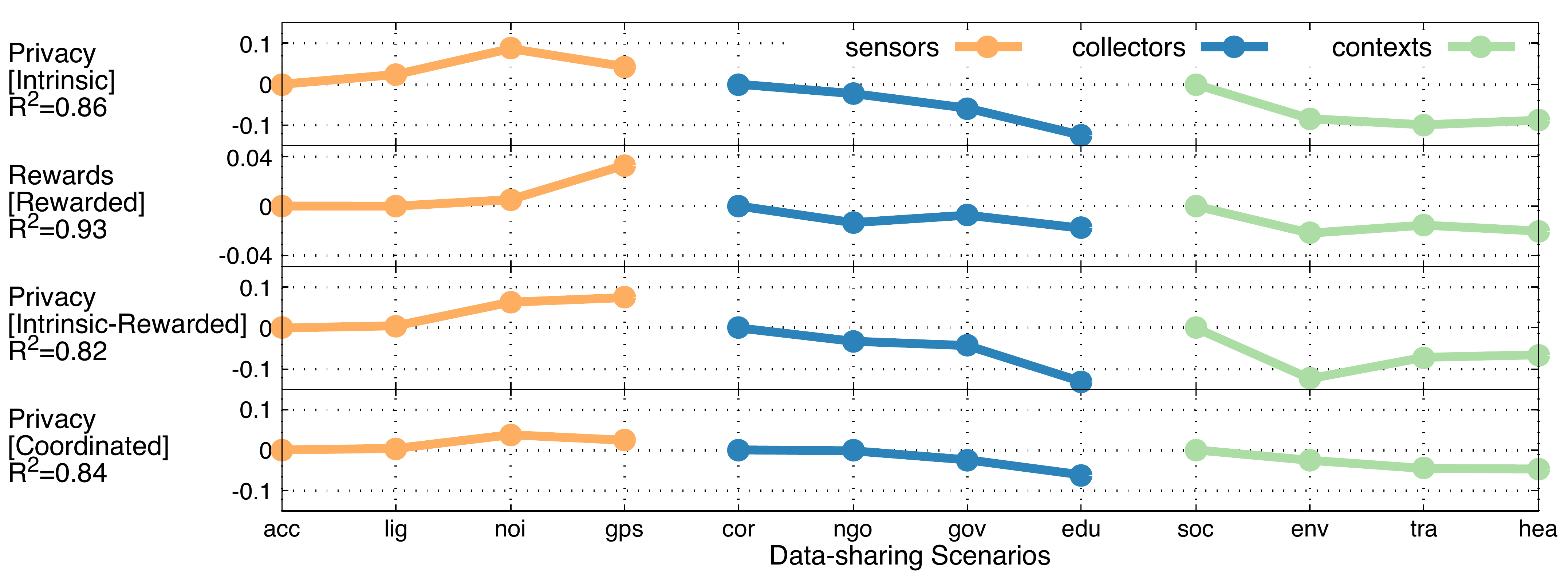}}
	\subfigure[The relative importance (partworth utilities) of the data-sharing criteria and elements (relative within each criterion ) derived from the different regression models of conjoint analysis and the perceived privacy sensitivity. The data collector is the most important criterion for the models based on privacy. In contrast, the sensor type is the most important criterion for the model based on rewards gain.]{\includegraphics[width=0.9\textwidth]{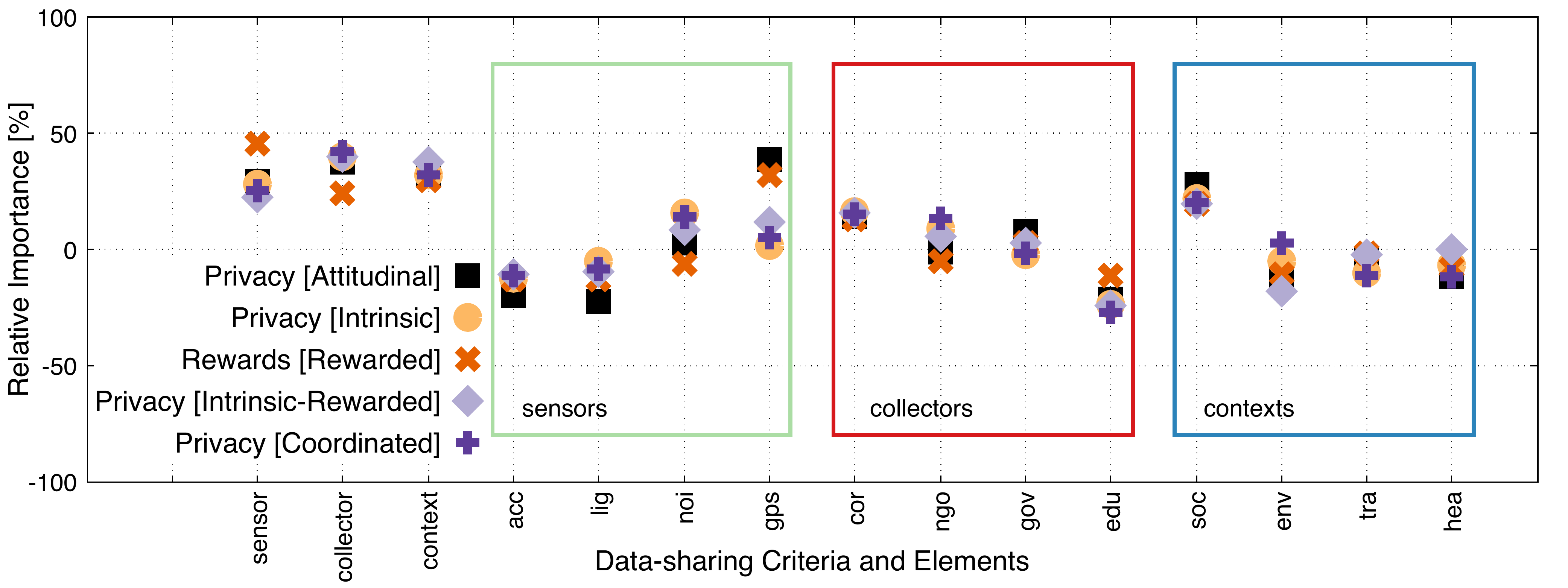}}
	\caption{\textbf{Rewarded individuals, who share data shift the importance from collectors and contexts to data}. Via a conjoint analysis, four multiple linear regression models are compared. It explains how the different data-sharing criteria and elements influence different key data-sharing behaviors.}\label{fig:conjoint-analysis-per-criterion}
\end{figure}

Here we study the causal link between the data-sharing criteria/elements (independent variables) and the privacy/reward gains (dependent variables) in different experimental conditions. Four explanatory models based on a conjoint analysis are outlined in Section~\ref{subsec:conjoint-analysis}. Fig.~\ref{fig:conjoint-analysis-per-criterion}a illustrates the regression coefficients of the models, while Fig.~\ref{fig:conjoint-analysis-per-criterion}b shows the relative importance of the data-sharing criteria and their elements calculated from these coefficients. All models come with $R^{2}$>0.8 and with statistically significant values of relative importance ($p<0.05$) for the vast majority of data-sharing elements as shown in Table~\ref{tab:conjoint-results}, Section~\ref{sec:conjoint-analysis} of SI. Fig.~\ref{fig:conjoint-analysis-per-criterion}b also shows the perceived relative importance derived from the self-reported entry survey questions. 

The data collector is the most important criterion (40.73\% on average, Fig.~\ref{fig:conjoint-analysis-per-criterion}b) for all models that predict privacy, and this criterion explains privacy loss (Fig.~\ref{fig:conjoint-analysis-per-criterion}a). Context follows with a 33.91\% of importance explaining privacy loss, while sensor type shows the lowest importance of 25.36\%, explaining the privacy gains. The consistency of these three privacy models reveals the following: the data collectors to whom individuals share data determine to a high extent (i) the privacy level under intrinsic or coordinated data sharing and (ii) the privacy loss under rewarded data sharing. The type of data they share plays a more minor role, though a positive one for privacy preservation. The models align well with the perception of individuals: 29.4\%, 37.85\% and 32.75\% for sensor type, collector and context respectively (Fig.~\ref{fig:conjoint-analysis-per-criterion}b). In contrast, for data-sharing choices of individuals with reward gains, the dominant criterion is the type of sensor data with a 45.4\% of relative importance over the data collector and context with 24.55\% and 30.01\% respectively. The collectors and contexts explain loss of rewards, while the type of sensor, and in particular the GPS, explains reward gain. GPS, as a privacy-sensitive sensor, provides high gain of rewards, and individuals are likely to be accustomed with apps accessing their GPS data, which is likely to reduce privacy-preservation. Choices that improve rewards suggest a radically different decision frame than the ones that improve privacy: \textit{a shift from protecting to sharing GPS data without strongly distinguishing anymore the data collectors and contexts}. 

Fig.~\ref{fig:conjoint-analysis-per-criterion}b also provides the following observations: The relative importance of the perceived privacy sensitivity over the 12 data-sharing elements is positively correlated with all models based on privacy: $R=0.97, t(10)=12.22, p=\num{2.46e-7}$ for rewarded data sharing, $R=0.84, t(10)=4.87, p=0.00066$ for intrinsic$-$rewarded, $R=0.69, t(10)=3.025, p=0.013$ for the coordinated data sharing and $R=0.67, t(10)=2.89, p=0.016$ for the intrinsic one. All models come with a positive relative importance for GPS (12.67\%), corporation (15.16\%) and social networking (20.42\%), while negative one for accelerometer (-11.85\%), light (-8.9\%), educational institutes (-21.52\%), transportation (-6.13\%) and health (-6.63\%).


\subsection{From intrinsic to rewarded data sharing: five behavior changes}\label{subsec:data-sharing-behavioral-patterns}

\begin{figure}[!htb]
	\centering
	\subfigure[Data-sharing group behaviors for intrinsic vs. rewarded data sharing.]{\includegraphics[width=0.32\textwidth]{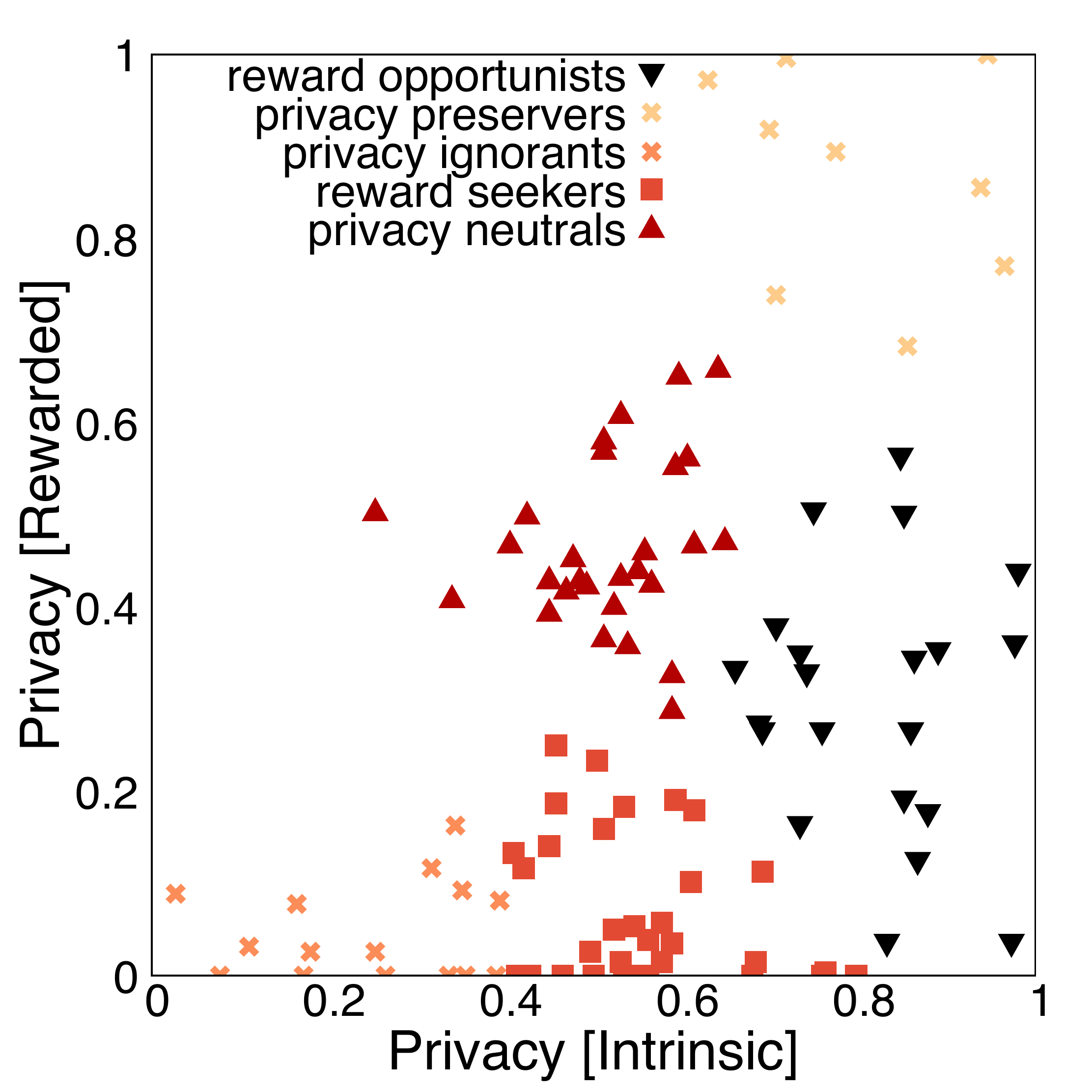}}
	\subfigure[Privacy of groups over consecutive rewarded data-sharing choices.]{\includegraphics[width=0.32\textwidth]{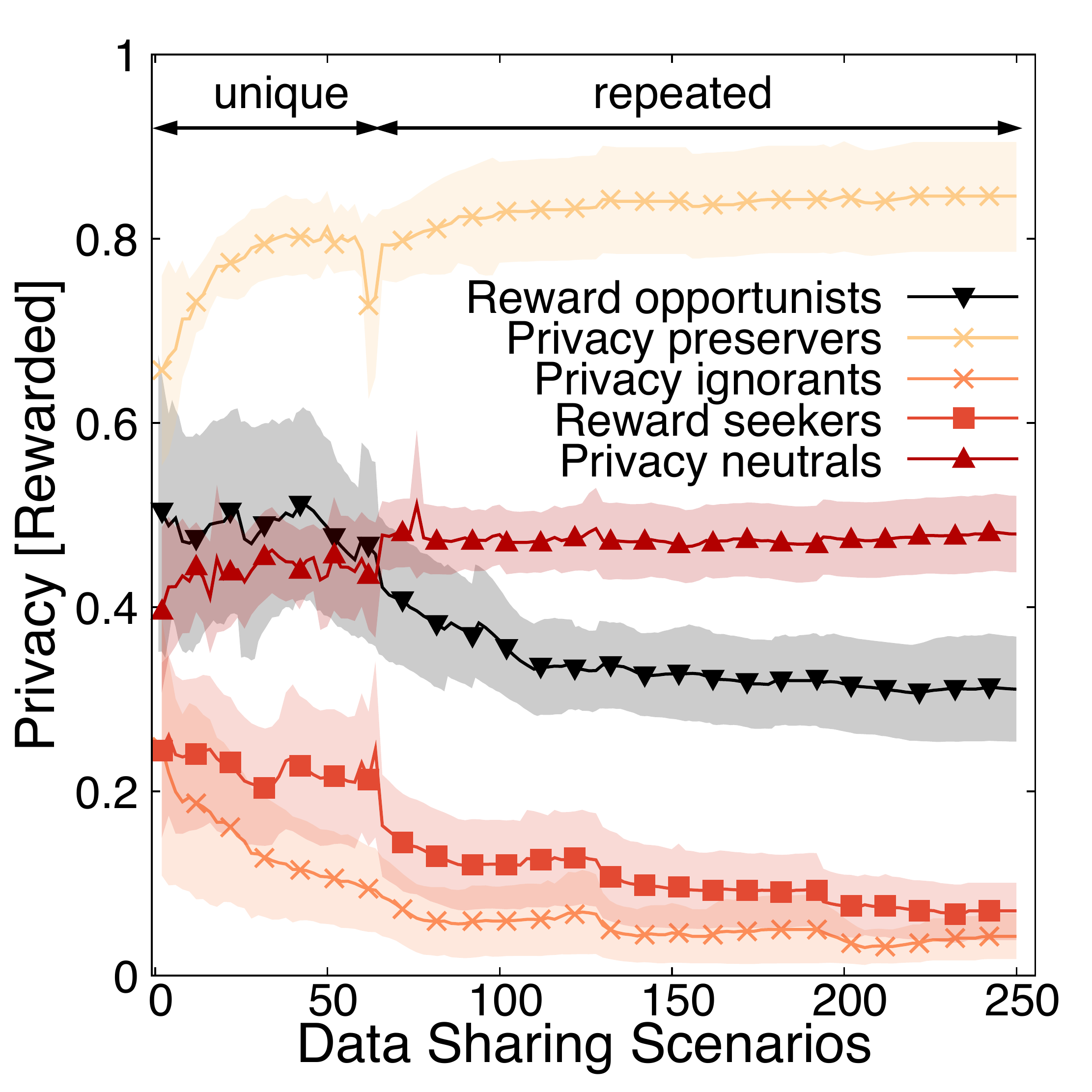}}
	\subfigure[Group pair differences of privacy sensitivity over data-sharing criteria]{\includegraphics[width=0.32\textwidth]{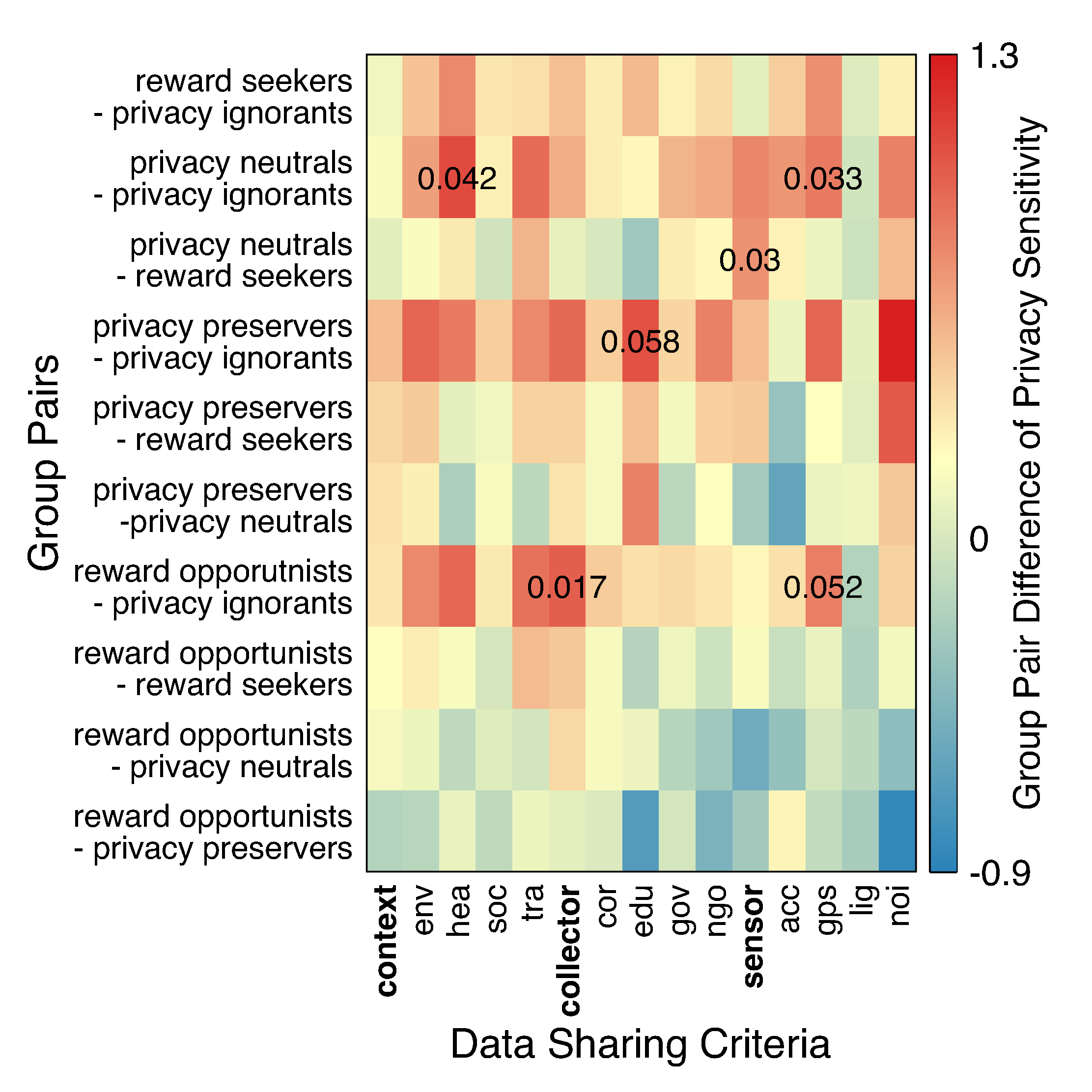}}
	\caption{\textbf{Five key group behaviors in data sharing and their effects}. Revealing the group behaviors under intrinsic vs. rewarded data sharing, how the group behaviors are formed over the passage of consecutive data-sharing decisions and the difference of privacy sensitivity between group pairs for the data-sharing criteria and elements.}\label{fig:behavioral-patterns}
\end{figure}

\noindent \textbf{Identifying group behaviors}. Table~\ref{tab:groups} provides an exemplary of all nine possible behavioral transitions that can happen in data sharing as a result of introducing monetary rewards. A clustering and stability analysis are performed in the experimental data projected in Fig.~\ref{fig:behavioral-patterns}a (intrinsic vs. \nth{1} rewarded), which reveal five robust behavioral patterns out of the 9 possible ones (similar groups are observed for intrinsic vs. \nth{2} rewarded). See Section~\ref{subsec:group-behavior} for more information. Some individuals are oblivious to rewards. Yet, these are the ones who intrinsically share a significant amount of data (\emph{privacy ignorants} and \emph{privacy neutrals}) or do not share data (\emph{privacy preservers}). \emph{Reward seekers} increase the data-sharing level when rewarded, while \emph{reward opportunists} intrinsically preserve privacy but eventually share a significant amount data when rewarded. It is astonishing that a moderate sacrifice of privacy preservation by rewards is not observed (privacy sacrificers in Table~\ref{tab:groups}), meaning that rewards significantly polarize individuals to keep protecting privacy or give up significant privacy. There are also no cases observed in which rewards motivate change to privacy protection; however, rewards reinforce privacy protection for privacy preservers. 
 
\noindent \textbf{Groups behavior converges to stable, while boundary ones polarize}. The behavioral pattern of privacy sacrificer (Table~\ref{tab:groups}) is found to be a transient one and observed within the reward opportunists during the first unique responses to the 64 data-sharing scenarios (see Fig.~\ref{fig:behavioral-patterns}b). When though these individuals get more involved in reevaluating their decisions, they converge to a further privacy sacrifice of 30.9\%. The minimum number of questions answered by all groups is 250. This incremental privacy decline in reoccurring decision-making is also observed in reward seekers and privacy ignorants that decrease their privacy level by 55.7\% and 64.8\% respectively. On the contrary, privacy preservers show a further increase in their privacy by 8.7\% as they reevaluate their data-sharing decisions. Such a privacy increase of 8.1\% is also observed for privacy neutrals. 

Strikingly, the two boundary behavioral patterns of privacy preservers and privacy ignorants show polarization from the very first data-sharing decisions. These individuals reinforce the privacy preservation and privacy ignorance respectively throughout the choices they make and regardless of whether these choices are the primary ones (the first 64 questions) or the reassessments (the follow up reinvoked questions). A similar behavior is documented for data sharing in social media~\cite{Stutzman2013,Acquisti2013,Acquisti2015}, though this is the first evidence of such behavior in a broader context, involving both privacy and rewards dilemmas.



 
\noindent \textbf{How privacy sensitivity of data-sharing criteria explains group behaviors}. Fig.~\ref{fig:behavioral-patterns}c shows all group pairs and the differences between these groups in terms of how privacy sensitive they regard each data-sharing criterion (attitudinal). Statistically significant observations ($p\leq0.05$) and those close to the significance threshold are marked in Fig.~\ref{fig:behavioral-patterns}c. These results are derived with a post hoc Tukey's range test ($\alpha=0.05$) after a one-way Analysis of Variance (ANOVA). The independent variable is calculated within the groups by the privacy change from intrinsic to rewarded data sharing. The dependent variables are the privacy sensitivity of the data-sharing criteria and their elements. Several of these criteria explain the data-sharing groups with a statistical significance (see Fig.~\ref{fig:criteria-significance}, Section~\ref{sec:ANOVA} in SI): transportation ($F(4,111) = 2.779, p = 0.03$), data collector ($F(4,110) = 2.463, p = 0.027$), sensor ($F(4,110) = 2.686, p = 0.031$), GPS ($F(4,110) = 2.201, p = 0.033$), noise ($F(4,110) = 3.573, p = 0.056$).

In Fig.~\ref{fig:behavioral-patterns}c, the data collector ($p=0.017$) and the GPS sensor ($p=0.052$) explain the privacy-sensitivity difference between reward opportunists and privacy ignorants: rewarded individuals of these groups share a significant level of data, while reward opportunists preserve privacy without rewards. Compared to privacy ignorants, reward opportunists find data collector and GPS more privacy intrusive by 24.2\% and 20.4\%. Similarly, the context of health ($p=0.042$) and the GPS sensor ($p=0.033$) explain the divergence between privacy neutrals and privacy ignorants. Privacy neutrals find these two data-sharing criteria 26.6\% and 20.9\% more privacy intrusive than privacy ignorants. Privacy neutrals also find sensors ($p=0.033$) more privacy intrusive than reward seekers by 18\%, which explains the higher data sharing of rewards seekers under rewards. Finally, the data-sharing criterion of educational institute determines when individuals share a very high or very low level of data with or without rewards: privacy preservers find the context of education ($p=0.058$) 25.9\% more privacy intrusive than privacy ignorants. 



\section{Discussion}\label{sec:discussion}


The findings reveal that a significant privacy recovery is attainable within the modus operandi of a data collective. This is a radical shift from the mainstream thought of privacy as a personal value to privacy as a collective value~\cite{Veliz2020}, a public good shared within a community of citizens generating data. Coordinated data sharing supported by a trustworthy decentralized AI automates and scales up collective arrangements for sharing under the doctrine `as little as possible as much as necessary'. Such optimized arrangements would be otherwise too complex and expensive to achieve in a transparent way with existing top-down privacy policies and regulations or even with automated data-access committees~\cite{lawson2021}. 

Findings also reveal that data collectives create tangible benefits for online service providers that collect or access data shared in a coordinated way: data collection costs drop down dramatically and data are used more purposefully to deliver the required quality of service. This can create further remarkable cost reductions such as reduced data storage, security, energy and carbon footprint costs as well as costs for solving legal disputes that are more likely to incur when dealing with excessive personal data. 

Within rising information asymmetries and monopolies of knowledge in existing data markets and big tech, the capability of data collectives to coordinate data sharing at large-scale has been so far a gap~\cite{Morozov2019,Muldoon2022}. This is underlined in promising solutions from political and economic theory such as data-owning democracy~\cite{Fischli2022}, digital socialism~\cite{Morozov2019} and peer-to-peer digital commons~\cite{Bauwens2019}. Establishing data collectives at a community or municipality level can create alternative forms of data ownership and control; they can empower citizens participation based on an agenda of using digital assets for priorities such as social welfare and environmental sustainability~\cite{Muldoon2022,Asikis2021}. These blueprints can be the basis of alternative data-market designs that encourage business models based on social innovation without over-relying on excessive free personal data. Data collectives can further benefit from scale, for instance, increasing individuals who coordinate their data-sharing decisions or increasing individuals' contributions by generating more alternative data-sharing options. The AI system based on collective learning has a higher degree of freedom to calculate data-sharing choices that match the required data and recover more privacy in larger populations~\cite{Pournaras2018}. It is also decentralized to make coordination more resilient to computational bottlenecks. 

Science can also benefit from data collectives. They can scale up open data and citizen science initiatives, while improving the transparency and reproducibility of research. Moreover, data collectives can be a response to the current opaque models of generative AI such as ChatGPT. Selective data shared as a result of coordination can be used to train open and more transparent generative AI models, ethically aligned to community values. This could be a new type of `curricula' for training AI, institutionalized in a bottom-up way via data collectives. 

Choices under intrinsic and rewarded data sharing prioritize different criteria. Individuals better distinguish data collectors and contexts than the type of data they share. In contrast, rewarded individuals that give up privacy better distinguish the type of data they share, and in particular the GPS. Thus, rewards diminish the importance of who collects data and for what purpose. In this case, data collectors may have no competitive advantage against each other but instead excessive and irrelevant data that increase their costs and risks. 

The perceived privacy sensitivity of the data-sharing criteria explains different key data-sharing behaviors (groups), for instance, individuals who do not preserve privacy vs. individuals who sacrifice privacy under rewards. Raising awareness about the privacy sensitivity of data collectors can influence data-sharing decisions. This has implications for how privacy policies and data consents are designed to be more transparent and user-friendly. Data-sharing choices that preserve and give up significant privacy tend to polarize, thus highlighting the value of privacy for individuals who have it rather than for the ones who do not~\cite{Acquisti2013}. Coordinated data sharing breaks this vicious cycle by redistributing the privacy cost within the individuals for the benefit of all. This demonstrates opportunities for digitally networked societies without borders to reconcile different cultural norms on privacy. 

Future work can unleash further opportunities to reclaim privacy in the digital age: Spatio-temporal coordinated data sharing can automate and scale up the ``\emph{right to be forgotten}", which improves both privacy control and the willingness to share data, e.g. 10\%-18\%~\cite{BCG2012}. The feasibility of collective learning using optimization scenarios in time and space are earlier demonstrated for Smart City applications~\cite{Pournaras2018}. Nevertheless, defining and conveying to individuals the context of data use is not always straightforward and further work is required in this area, for instance, semantics and ontologies~\cite{lawson2021}. Moreover, beyond purposeful data sharing, speculative data analysis out of a specific context can also encourage innovation and creativity. In such scenarios, data collectors may have a more significant role for trust in data-sharing decisions. The acceptance of coordinated data-sharing recommendations requires a follow up study, in particular, the incentives and the interface design of the AI system for the broader population. Notwithstanding, earlier results demonstrate significant coordination capacity even when large portions of the population are not flexible~\cite{Pournaras2019}. The explainability of coordinated data sharing based on decentralized AI is particularly challenging and is expected to further shield the trust on data collectives.

\section{Methods}\label{sec:methods}

We outline here the experimental design and the developed technical infrastructure. We also illustrate the methods with which we analyzed the experimental data and the AI-based decision-support system with which coordinated data sharing is performed. 

\subsection{Living-lab experimental design}\label{subsec:experiment}
 
\begin{figure}[!htb]
	\centering
	\subfigure[Data-collection infrastructure.]{\includegraphics[width=0.38\textwidth]{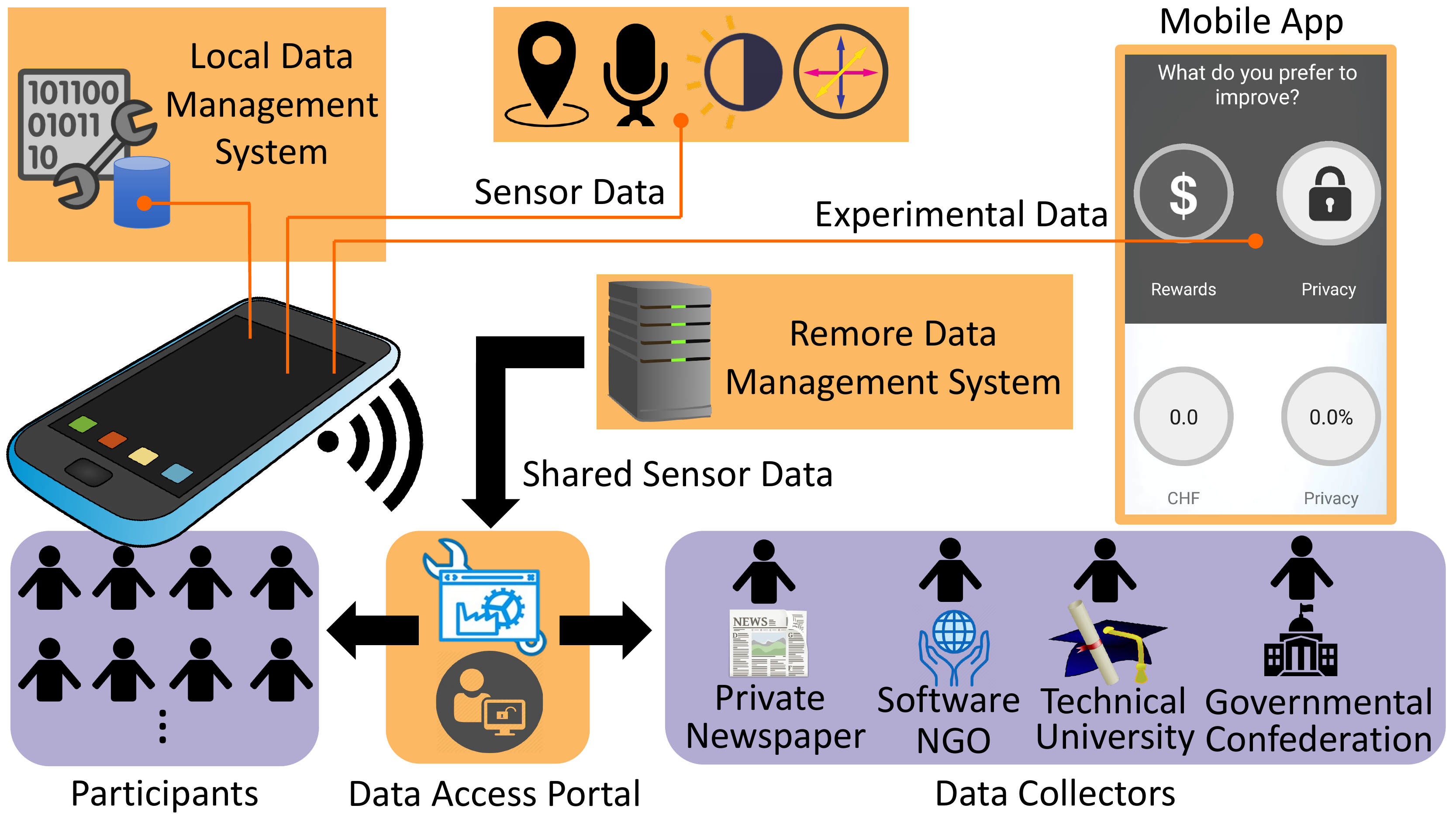}}\hfill
	\subfigure[Experimental design.]{\includegraphics[width=0.55\textwidth]{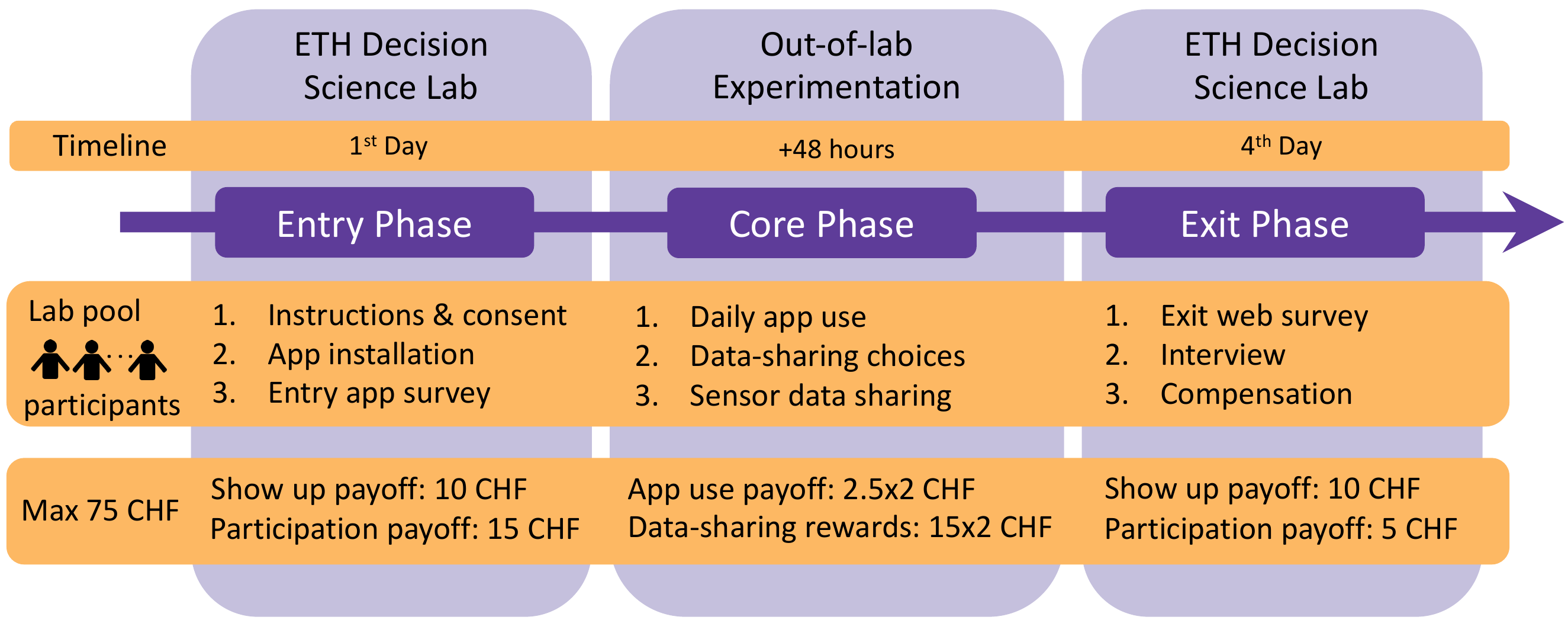}}
	\caption{\textbf{A data-collection infrastructure used for the design of a novel `living-lab' experiment of high realism and rigor}. The experiment consists of three phases in and out of the lab. Data are collected via smartphones and are made accessible to data collectors according to the privileges given by participants.}\label{fig:experimental-design}
\end{figure}

A novel design for a `living-lab' experiment is introduced. It defines a \emph{mixed-mode} experiment that seamlessly integrates in participants' everyday life, while the overall experimental process is orchestrated via the controlled environment and experimental protocols of the Decision Science Laboratory (DeSciL) of ETH Zurich~\cite{DeSciL2021}. The proposed experiment has received ethical approval by DeSciL and the Ethics Commission at ETH Zurich (\#EK 2016-N-40). To improve the realism of the experiment and comply to the non-deceiving policy of DeSciL, letters of support were collected from data collectors to confirm their interest in accessing the collected sensor data of participants. The study consist of three phases: (i) \emph{entry}, (ii) \emph{core} and (iii) \emph{exit}. Fig.~\ref{fig:experimental-design} provides an outline of the overall experimental process and the developed data-collection infrastructure (details are documented in Section~\ref{sec:experimental-design} of SI). 


\noindent \textbf{Recruitment approach and sampling biases}. The living-lab experimentation involves the recruitment of 123 participants during the entry phase, out of which 116 completed the exit phase and 89 participated in all phases. Aggregated privacy-reward records for all experimental conditions is found for 84 participants. Responses to the data-sharing scenarios for all experimental conditions are found for 73 participants. In the context of this study, a higher number of participants is particularly challenging and probably unrealistic as it requires significantly more resources for compensation/infrastructure, sacrifice of rigor, and much looser control of the experimental process. Instead, priority is given to a satisfactory compensation per participant for active participation in all experimental phases (see Section~\ref{subsec:compensation} in SI) and by incentivizing appropriately a large number of data-sharing choices: 27403 in total. Moreover, the development of a data-collection platform, including the data-access web portal and the mixed-mode experimental process, preserves an eminent realism, yet in well-controlled laboratory conditions that result at the end in a novel high-quality dataset to perform causal inference. 

Participants were recruited from the DeSciL pool~\cite{DeSciLPool2021}, mainly consisting of students of ETH Zurich and University of Zurich (see the invitation in Section~\ref{subsec:invitation} of SI). This pool is not representative of the population and is subject to sampling biases. However, smartphone users, who use a broad range of apps that require sharing of sensor data are mainly young people~\cite{Sarraute2014,Dimonte2006,Rice2003}, and therefore the students' profile fits well with the nature of the conducted experiment. Participants with technological literacy are also more likely to be familiar with data-sharing dilemmas involving a privacy cost to gain access to smartphone app services. Studying such a sample of participants can make results more compelling as shown in earlier experiments conducted on such recruitment basis~\cite{Mayer2016}. Only Android smartphone users are recruited, who are a large portion of the population, for instance, 39.8\% in Switzerland, 68.6\% in Europe and 72\% worldwide in 2016 according to StatCounter. Moreover, several smartphone apps with data-sharing decisions are made for both Android and iOS. Therefore there is no substantial evidence to suggest different decision patterns among the market share in the population as also supported in earlier work~\cite{Mayer2016}. Recruitment is performed in 8 sessions on a weekly basis. To eliminate any further temporal bias, each of the three phases in Fig.~\ref{fig:experimental-design} took place on the same day of the week. Table~\ref{tab:sessions} in SI provides an overview of the experimental sessions.

\noindent \textbf{Entry phase}. It takes place at DeSciL and it involves the following: (i) Collection of basic demographics about participants and information about their privacy profile using the survey questions of Table~\ref{table:entry-phase} in SI. (ii) Use of the privacy-intrusion level assigned to each data-sharing criterion and its elements (Questions~B.9-B.12) to calibrate the calculation of the monetary rewards for the core phase according to the model illustrated in Section~\ref{sec:model} of the SI. (iii) Collection of the intrinsic data-sharing decisions by letting participants choose once the data-sharing level for each of the 64 data-sharing scenarios (see Fig.~\ref{fig:app-entry}b in SI). The following question implements the data-sharing scenarios: 

\begin{question}
	Please choose the amount of \textbf{<sensor type>} sensor data shared with \textbf{<data collector>} to be used in the context of \textbf{<context>}.
\end{question}

\noindent There are in total five possible data-sharing levels to choose from (see Fig.~\ref{fig:app-entry}b in SI). 

\begin{figure}[!htb]
	\centering
	\subfigure[Privacy vs. rewards dilemma]{\includegraphics[width=0.223\textwidth]{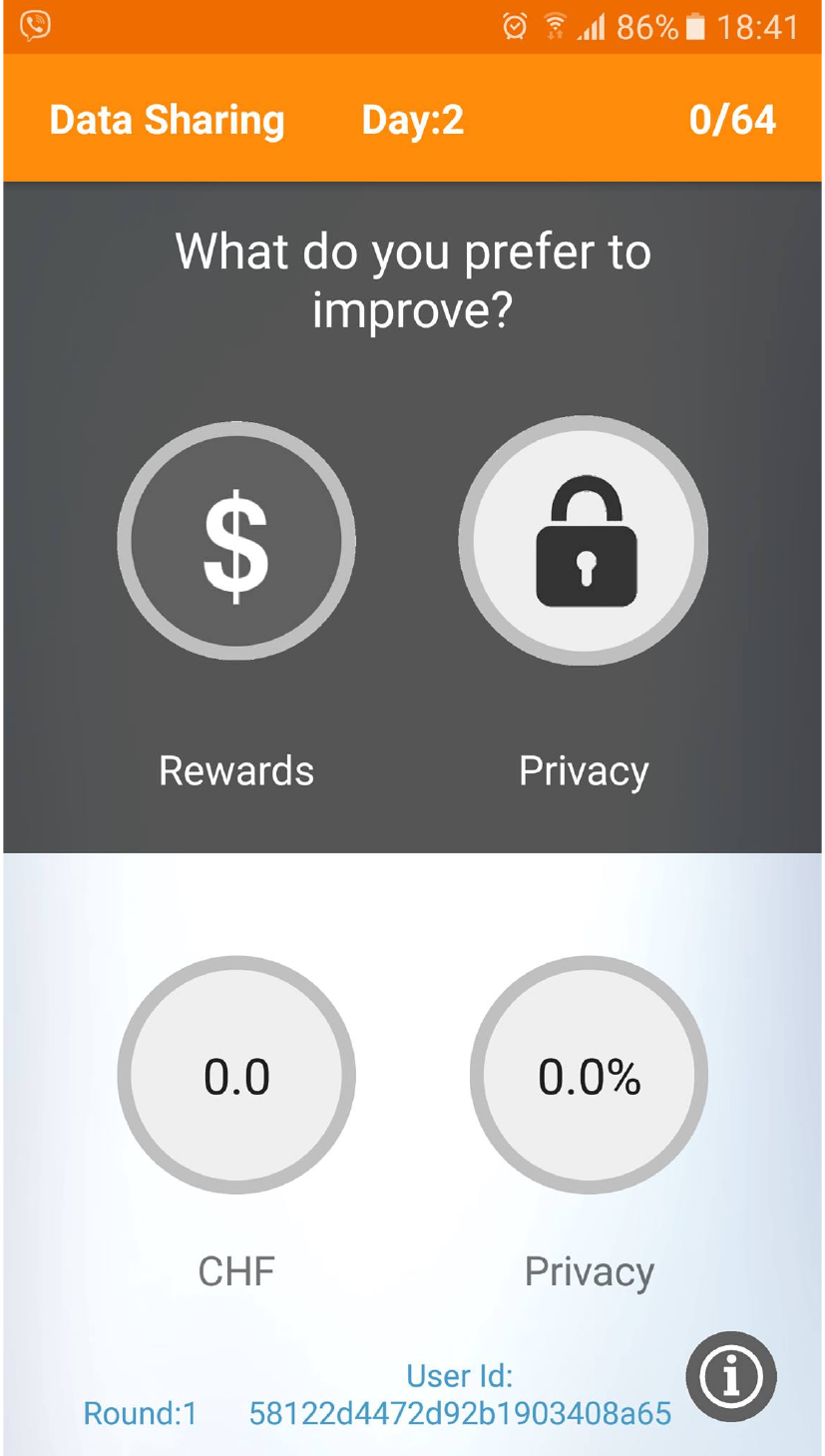}}
	\subfigure[Data-sharing scenario choice]{\includegraphics[width=0.305\textwidth]{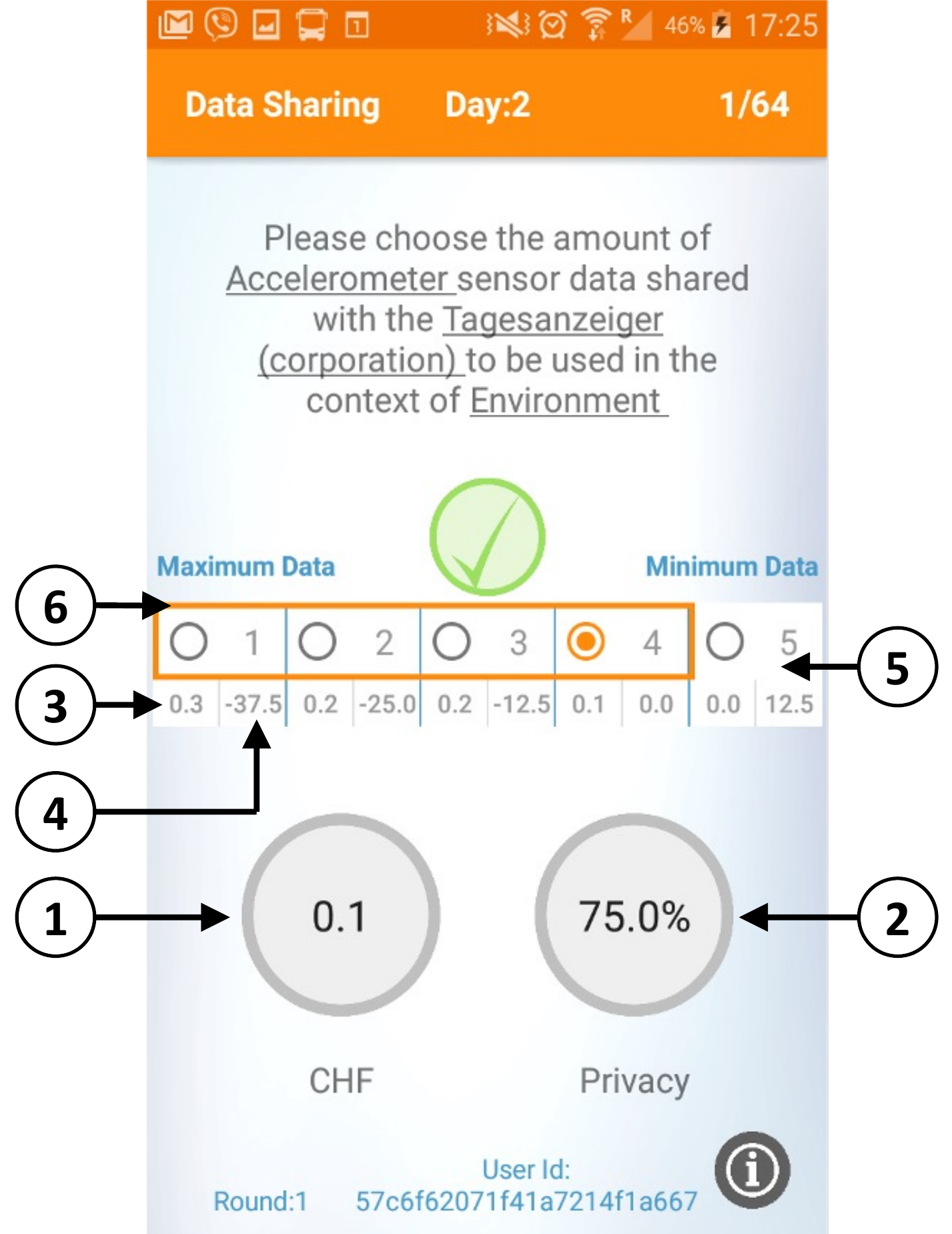}}
	\caption{\textbf{The key experimental functionality of the smartphone app}. Arrows: (1) Accumulated rewards. (2) Privacy level. (3) Gain/loss of rewards for a particular option. (4) Gain/loose of privacy for a certain option. (5) Data-sharing options. (6) Options in the improvement box.}\label{fig:app-core}
\end{figure}

\noindent \textbf{Core phase}. It takes place out of the lab and lasts for two days (48 hours), starting right after the completion of the entry phase. During the 24 hours of each day, participants are voluntarily involved in an (unlimited) sequence of dilemmas of either improving their privacy or rewards by sharing less or more data respectively in a data-sharing scenario. Fig.~\ref{fig:app-core} illustrates the two app screens for the privacy-rewards dilemma and the data-sharing scenario that follows. First, participants decide what to improve based on their privacy-rewards balance they currently have (Fig.~\ref{fig:app-core}a). Next, a data-sharing scenario is automatically retrieved with the latest choice made (Fig.~\ref{fig:app-core}b), marking the options that fulfill their goal (the improvement box, see Arrow 6). The retrieved scenario is the one that maximizes the improvement of the chosen goal, i.e. privacy or rewards. For each option, the app informs participants about the rewards and privacy they gain or lose (Arrows 3 and 4 respectively). After a choice, the participant moves back to the main screen of Fig.~\ref{fig:app-core}a with an updated privacy-rewards balance.

The first unique 64 data-sharing scenarios are the ones that participants have decided about during the entry phase. The difference in this core phase is that data sharing is rewarded based on two factors defined in the data-sharing model (see Section~\ref{sec:model} of SI): (i) the data-sharing level (the higher, the more rewards) and (ii) how privacy-intrusive the data-sharing scenario is according to each participant. More rewards are allocated to data-sharing scenarios involving criteria regarded highly privacy-intrusive by a participant. The latter personalization is derived from the responses of the entry phase (Questions B.9-B.12 in Table~\ref{table:entry-phase} of SI) without explicitly making participants aware of this. 

Within the 24 hours, participants can change their goal based on their privacy-reward balance. They continue responding to further retrieved data-sharing scenarios that can satisfy their goal, i.e. improve privacy or rewards, see Fig.~\ref{fig:app-core}a. This allows studying how data-sharing decisions evolve. Each decision in a data-sharing scenario overwrites the previous one for the calculation of the privacy-reward balance. At the end of the 24 hours, the process completes by locking the decisions of the 64 scenarios and sharing the data to the data-access web portal. This process runs for two days to validate the results, confirming similar data-sharing behavior at both days (see Fig.~\ref{fig:privacy-mismatch-scenarios}a as well as Fig.~\ref{fig:privacy-loss-groups}a and~\ref{fig:privacy-loss-groups}b in SI). 

\noindent \textbf{Exit phase}. The participants of each experimental session return to DeSciL on the 4th day. They answer a survey questionnaire, participate in an interview and receive their calculated compensation. The survey consists of questions that cover the following aspects (see Table~\ref{table:exit-phase-general} to~\ref{table:exit-phase-experiment} in SI): (i) smartphone use, (ii) user interface and functionality of the app, (iii) rewards and privacy, (iv) experimental process. The data collected during this phase have a supportive role serving the validation and interpretation of the results produced during the entry and core phase. See Section~\ref{subsubsec:exit-phase} of SI for further details.

\noindent \textbf{Compensation and monetary incentives}. Participants are compensated for their engagement in the experiment as well as for the sensor data they share. The engagement covers (i) showing up in the lab ($2\cdot 10=20$ CHF), (ii) completing the lab activities ($15+5=20$ CHF) and (iii) using the app in terms of answering at least once all 64 data-sharing scenarios ($2\cdot 2.5=5$ CHF). The rewards for the app use is distributed with a geomentric progression over the data-sharing scenarios to eliminate dropout effects (see Section~\ref{subsec:compensation} of SI). Those who successfully complete all experimental phases receive the total fixed compensation of 45 CHF and an additional maximum reward of $2\cdot 15=30$ CHF based on the amount of shared data. Fig.~\ref{fig:experimental-design} shows how the total maximum amount of 75 CHF is allocated over the experimental process. Section~\ref{subsec:compensation} of SI further motivates the allocation of these compensations.

\subsection{Technical infrastructure}\label{subsec:infrastructure}

 Fig.~\ref{fig:experimental-design} outlines the technical infrastructure developed to serve the designed experimental process. Two types of data are collected by the smartphone app: (i) the sensor data that participants explicitly choose to share and (ii) all data from participants' choices and survey answers used for the analysis. These data are stored on a remote server and locally on the smartphone for redundancy so that they can be restored during the exit phase by moderators in case of software or communication failures. 

The developed infrastructure consists of the following interactive systems: (i) the \emph{local} and (ii) \emph{remote data-management system}, (iii)  the \emph{smartphone app} and (iv) the \emph{data-access web portal}. The two data-management systems synchronize and secure the shared sensor data as well as the experimental data. The smartphone app is developed to run on Android devices. The data-access web portal stores the shared data and provides authorized access to the registered participants of the experiment as well as the data collectors involved in the data-sharing scenarios. Making available this system improves the realism of the experiment by realizing the actual data-sharing decisions, while allowing the experimental design to comply with the non-deceiving policy of DeSciL. See Section~\ref{sec:infrastructure} of SI for further details.

\subsection{Privacy calculations for sensors, collectors and contexts}\label{subsec:privacy-level-reinforcement}

The privacy measurements in Fig.~\ref{fig:privacy-mismatch-elements}a are made as follows: In the case of the attitudinal data-sharing condition, the mean privacy level is calculated by normalizing (in $[0,1]$ over all participants) the privacy sensitivity reported in the Questions B.10-B.12 during the entry phase. In the intrinsic, rewarded and coordinated data-sharing conditions, the privacy level of a certain sensor, data collector or context is the normalized privacy mean across all participants for 16/64 data-sharing scenarios that contain this respectively (see Fig.~\ref{fig:privacy-mismatch-scenarios}a). In the coordinated data-sharing conditions, this is calculated using the mean privacy level of the data-sharing scenarios selected over all 10 repetitions of the coordination with a random positioning of the agents (see Section~\ref{subsec:learning} for more information). 

The expected privacy level of a data-sharing scenario (see  shaded areas in Fig.~\ref{fig:privacy-mismatch-scenarios}a) is calculated by the mean privacy level of the sensor, collector and context that comprise the data-sharing scenario. The expected privacy level of a certain sensor, data collector, or context is the mean expected privacy level over 16/64 data-sharing scenarios containing this. The relative difference between the actual privacy level and the expected one defines the \emph{privacy reinforcement}. Detailed measurements are illustrated in Fig.~\ref{fig:privacy-reinforcement}, Section~\ref{sec:privacy-reinforcement} of SI. 

\subsection{Coordinated data-sharing via decentralized AI}\label{subsec:learning}

Coordinated data sharing is modeled as a decentralized discrete-choice multi-agent combinatorial optimization problem. It is designed to recover excessive privacy loss by rewarded data sharing. A decision-support system implements the optimization that achieves the coordination. The discrete choice model and the coordination method are outlined below. 

\noindent \textbf{Data-sharing plans and elicitation of privacy sensitivity}. Each participant comes with three data-sharing plans extracted from the living-lab experiment as follows: each plan is a sequence of 64 real values that represent the data-sharing choices made at each scenario and each experimental condition: intrinsic, \nth{1} rewarded and \nth{2} rewarded. Each plan has a privacy cost represented by a real value. It is calculated by the mean normalized level (in $[0,1]$) of shared data over the data-sharing scenarios. Alternative privacy valuation schemes are assessed in Section~\ref{sec:privacy-valuation schemes} of SI.

\noindent \textbf{Steering data sharing using privacy-preservation goal signals}. A goal signal represents a data-collection scenario with the minimum required data to enable a data-driven service or application~\cite{Ding2019,Eick2020,Kim2021}. Five privacy-preservation goal signals for data sharing are generated using the intrinsic data-sharing choices of participants. Each goal signal is a sequence of 64 values corresponding to the data-sharing scenarios. For each data-sharing option out of the five possible ones, a goal signal is calculated with the 64 values representing the probability of participants choosing this data-sharing option without rewards. Similarly with the data-sharing options, the five goal signals are referred to within the range of very low to very high privacy preservation. Fig.~\ref{fig:coordinated-data-sharing}, Section~\ref{sec:goal-signals} in SI illustrates the five goal signals. 

\noindent \textbf{Coordinated data sharing}. The goal of the data collective is to choose and aggregate (sum up element-wise) the data-sharing plans of all individuals such that the resulting signal matches a given goal signal for privacy preservation. This matching is measured here with the residual sum of squares between these two signals (standardized). As this goal cannot be satisfied by letting individual participants choosing independently the plan with the best matching (minimizing a non-linear cost function), coordination between participants' choices is required. This discrete-choice coordination problem is combinatorial NP-hard and requires approximating solutions~\cite{Pournaras2018}. The coordination capability can be generalized to a multi-objective combinatorial optimization problem in which the data collective minimizes the following cost function: 

\begin{equation}\label{eq:cost-function}
	(1-\alpha-\beta)\cdot \text{privacy inefficiency}+ \alpha \cdot \text{privacy unfairness} + \beta \cdot \text{privacy cost},
\end{equation}

\noindent where privacy inefficiency is the residual sum of squares between the aggregated data-sharing plans and the goal signal, privacy cost is mean cost of the selected plans and the privacy unfairness is the dispersion (variance) of privacy cost over individuals. The parameters $\alpha$ and $\beta$, for $\alpha+\beta=1$ and $\alpha, \beta \in [0,1]$, are self-determined by each individual and model a behavioral continuum between selfish vs. altruistic behavior in terms of data sharing. A selfish individual that minimizes privacy without coordinating its data sharing with other individuals is determined by $\beta=1, \alpha=0$. An individual that minimizes the collective privacy inefficiency without counting its personal privacy cost is an altruistic one by $\beta=0, \alpha=0$. And these altruistic individuals can balance for privacy unfairness by increasing the $\alpha$ parameter. 

\noindent \textbf{A decentralized computational approach for coordination}. The collective learning method of I-EPOS is used to cope with the computational and communication complexity of the coordinated data-sharing problem~\cite{Pournaras2018}. This algorithm is used as a decision-support system that automates and scales up the coordination, which would otherwise be too complex and infeasible for humans to perform without digital assistance. As featured by UNESCO IRCAI~\cite{CollectiveLearningIRCAI}, this method is particularly fitting in this privacy context: (i) The algorithm itself is privacy-preserving by design as it exclusively relies on exchanging aggregated (and not individual) information. The use of differential privacy and homomorphic encryption can also enhance the overall security of information aggregation. (ii) The algorithm is highly cost-effective with a low computational and communication complexity compared to other multi-agent approaches for combinatorial optimization problems~\cite{Pournaras2018}. The data-sharing choices calculated by the algorithm can rapidly match the goal signal with a low communication exchange between the agents. (iii) The algorithm is open-source, decentralized and can scale up without relying on a trusted third party, which makes it particularly applicable for bottom-up data collectives. (iv) The algorithm can operate in different faulty environments and application scenarios~\cite{Pournaras2020}. 

\noindent \textbf{Collective learning parameterization}. Agents are self-organized in a binary balanced tree within which they are positioned randomly. Coordination repeats 10 times, each with a different random positioning of the agents. For each random positioning, collective learning runs for 50 learning iterations. Each iteration proceeds from leaves to root and back to leaves. It results in the selection of data-sharing plans that minimize at an aggregate level the cost function in Equation~\ref{eq:cost-function}. More information about the algorithm can be found in earlier work~\cite{Pournaras2018}.

\subsection{Causal inference with conjoint analysis}\label{subsec:conjoint-analysis}

The complete factorial design of $3$ data-sharing criteria each with $4$ elements results in 64 scenarios encoded by a sequence of $12-3=9$ dummy variables. These represent the membership of a certain sensor, collector and context in a data-sharing scenario. Multiple linear regression models are constructed using as independent variables the nine dummy variables ($4-1=3$ variables per data-sharing element are used to resolve the linear dependency problem in multiple regression). The dependent variables that distinguish the regression models include the following (Fig.~\ref{fig:conjoint-analysis-per-criterion}): privacy (intrinsic, intrinsic$-$\nth{2} rewarded, coordinated with very low privacy-preservation goal) and gained rewards (\nth{1} and \nth{2} rewarded data sharing with those individuals who intend and do improve rewards as in Fig.~\ref{fig:app-core}). These privacy and reward values across the 64 data-sharing scenarios of the full factorial design are used for a rating-based conjoint analysis. Other regression models with lower statistical power are assessed and further illustrated in Fig.~\ref{fig:conjoint-coefficients}, Table~\ref{tab:conjoint-results}, Section~\ref{sec:conjoint-analysis} of SI. 

The regression models result in the 12 coefficients for each data-sharing element as shown in Fig.~\ref{fig:conjoint-analysis-per-criterion}a. Together with a constant (Table~\ref{tab:conjoint-results} in SI), they predict the depend variable. Using the coefficients, the partworth utilities are estimated that calculate the relative importance of each data-sharing criterion and element (Equations~\ref{eq:criterion-utility} and~\ref{eq:element-utility-per-criterion} in SI). For each data-sharing element, the relative importance is calculated across the elements of the criterion it belongs (Equation~\ref{eq:element-utility-per-criterion}) or across all elements (Equation~\ref{eq:element-utility}). The latter is shown in Fig.~\ref{fig:conjoint-analysis-per-element} of SI. The conjoint analysis models are compared to the mean relative perceived privacy sensitivity as declared by participants in the Questions~B.9-B.12 in Table~\ref{table:entry-phase} of SI. 

\subsection{Extraction and validation of group behavior}\label{subsec:group-behavior}

\noindent \textbf{How groups are extracted}. To extract the data-sharing group behaviors, the participants' privacy level under intrinsic and \nth{1}/\nth{2} rewarded data sharing are clustered using three clustering techniques of R: (i) k-means~\cite{Jain2010} (\texttt{kmeans}), (ii) hierarchical clustering~\cite{Langfelder2012,Murtagh2014} (\texttt{hclust}) and (iii) partitioning around medoids~\cite{Kaufman2009} (\texttt{pamkCBI}). A subset of 110 participants were clustered that made both intrinsic and rewarded data-sharing decisions.
An optimum number of five clusters is confirmed in all three methods that correspond to the data-sharing groups marked in Fig.~\ref{fig:behavioral-patterns}a. An exemplary of observed and unobserved group behaviors is outlined in Table~\ref{tab:groups}. 

\begin{table}[!htb]
	\caption{\textbf{Exemplary of possible group behaviors with and without rewards in data sharing}. A low, moderate and high level of data sharing is assumed for illustration purposes. \cmark: denotes the observed group behaviors. \xmark: denotes the unobserved group behaviors.}\label{tab:groups}
	\centering
	\begin{tabular}{lcccccc}\toprule
		& \multicolumn{3}{c}{Without Rewards} & \multicolumn{3}{c}{With Rewards}
		\\\cmidrule(lr){2-4}\cmidrule(lr){5-7}
		\multicolumn{1}{r}{Data Sharing:} & $Low$  & Moderate & High    & Low  & Moderate & High\\\midrule
		Privacy ignorants    &  &  & \cmark &  &  & \cmark \\
		Privacy neutrals &  & \cmark &  &  & \cmark &  \\
		Privacy preservers & \cmark &  &  & \cmark &  & \\
		Rewards seekers   &  & \cmark &  &  &  & \cmark \\
		Rewards opportunists   & \cmark &  &  &  &  & \cmark\\\bottomrule
		Privacy sacrificers   & \xmark &  &  &  & \xmark & \\
		Reward opposers (sharer)  & &  & \xmark & \xmark &  & \\
		Reward opposers  (neutral) & & \xmark &  & \xmark &  & \\
		Reward sacrificer  (sharer) & &  & \xmark &  & \xmark & \\\bottomrule
	\end{tabular}
\end{table}


\noindent \textbf{How groups are validated}. In the case of k-means and hierarchical clustering, the optimum number of five clusters is derived by performing a bootstrap evaluation (\texttt{clusterboot} of R) of the clusters~\cite{Hennig2007}. It assesses both the stability of the clusters and the stability of different clustering algorithms. The pamkCBI algorithm performs partitioning around medoids. The number of clusters is estimated by the optimum average silhouette width~\cite{Rousseeuw1987,Reynolds2004}. However, a bootstrap evaluation is also performed for pamkCBI for a complete comparison of the three algorithms. An outline of the clusters stability (mean Jaccard similarity) and the number of dissolved clusters for 100 bootstrap iterations is given in Table~\ref{tab:bootstrap} of SI. Visual inspections show that all three algorithms find the same clusters, while k-means achieves a mean Jaccard similarity (\texttt{bootmean}) higher than 0.75 for all clusters, which indicates stable clusters. As such, the groups of k-means are analyzed in this paper (Fig.~\ref{fig:behavioral-patterns}). Note also that the population split over the data-sharing groups matches well to Westin's general population privacy indexes, see further Section~\ref{sec:grouping} of SI.


\section*{Data Availability}

The collected data of the living-lab experiment are made available at: \url{https://doi.org/10.6084/m9.figshare.21750158}. The generated plans are made part of the following planning portfolio: \url{https://doi.org/10.6084/m9.figshare.7806548.v5}.

\section*{Code Availability}

The source code of the AI system is under active development at \url{https://github.com/epournaras/epos}. Source code used and developed for this paper is made available at \url{https://doi.org/10.5281/zenodo.7457575}.

\bibliography{ms}
%
\bibliographystyle{naturemag}
%
%

\section*{Acknowldgments}

The authors would like to thank Prof. Dirk Helbing for supporting and encouraging the work on this project. Special thanks go to Stefan Wehrli and the rest of the ETH DeSciL staff members for their support to the overall experiment design. The authors would also like to particularly thank: Ramapriya Sridharan for the development of the app, Lewin K\"onemann for the design of user interface as well as all the Nervousnet development team for their support and expertise. Athina Voulgari supported the experimental process, and Stefan Klauser engaged the data collectors for the realism of the experiment. Thanks to Thomas Wellings and Lily Lovingood for reviewing this paper. 

Evangelos Pournaras is supported by a UKRI Future Leaders Fellowship (MR\-/W009560\-/1): `\emph{Digitally Assisted Collective Governance of Smart City Commons--ARTIO}' and the SNF NRP77 `Digital Transformation' project "Digital Democracy: Innovations in Decision-making Processes", \#407740\_187249. This work was also earlier supported by the White Rose Collaboration Fund: `\emph{Socially Responsible AI for Distributed Autonomous Systems}', a 2021 Alan Turing Fellowship, the European Community's H2020 Program under the scheme `INFRAIA-1-2014-2015: Research Infrastructures', grant agreement \#654024 `\emph{SoBigData: Social Mining \& Big Data Ecosystem}' (\url{http://www.sobigdata.eu}) and the European Community's H2020 Program under the scheme `ICT-10-2015 RIA', grant agreement \#688364 `\emph{ASSET: Instant Gratification for Collective Awareness and Sustainable Consumerism}'.

Mark Ballandies has received funding from the European Research Council (ERC) under the European Union’s Horizon 2020 research and innovation programme (project `\emph{Co-Evolving City Life - CoCi}', grant agreement No 833168) and the Swiss National Science Foundation for the EU FLAG ERA project \emph{FuturICT2.eu} under the grant number 170226. 

Chien-fei Chen has been supported by the Engineering Research Center Program of the National Science Foundation (NSF), the Department of Energy in the US under NSF Award Number EEC-1041877 and the CURENT Industry Partnership Program.

\section*{Author Contributions}

E.P. wrote the manuscript, conceived the study, designed and developed the AI models and analyzed the data. M.C.B. edited the manuscript, supported the living-lab data collection and analyzed the data. S.B. edited the manuscript, developed the AI models and analyzed the data. C.C. edited the manuscript and analyzed the data.

\section*{Competing Interests}

The authors declare no competing interests.

\section*{Additional Information}

Supplementary information: The online version contains supplementary material available at \url{https://doi.org/}. Correspondence and requests for materials should be addressed to Evangelos Pournaras. 
\makeatletter\@input{yy.tex}\makeatother
\end{document}


\title{Collective Privacy Recovery: Data-sharing Coordination via Decentralized Artificial Intelligence\\
	\large Supplementary Information}
\author[1]{Evangelos Pournaras*}
\author[2]{Mark Christopher Ballandies}
\author[2]{Stefano Bennati}
\author[3]{Chien-fei Chen}

\affil[1]{School of Computing, University of Leeds, Leeds, UK, E-mail: e.pournaras@leeds.ac.uk}
\affil[2]{Computational Social Science, ETH Zurich, Zurich, Switzerland, E-mails: mark.ballandies@ethz.ch, stefano@bennati.me}
\affil[3]{Institute for a Secure and Sustainable Environment, University of Tennessee, Knoxville, E-mail: cchen26@utk.edu}

\renewcommand\Authands{ and }

\maketitle

%

\tableofcontents

\section{General Data-Sharing Model}\label{sec:model}

This section provides the mathematical formulation of human data-sharing choices under personalized (monetary) incentives. Table~\ref{table:math-symbols} provides an overview of the mathematical notations. 

\begin{table}[!htb]\footnotesize\centering
	\caption{An overview of the mathematical symbols.}
	\centering
	\resizebox{\columnwidth}{!}{%
		\begin{tabular}{l l}
			\hline
			Symbol & Interpretation \\ \hline
			$\numOfCriteria$ & Number of data-sharing criteria\\
			$\criterion$ & A data-sharing criterion\\
			$\numOfCriterionDimensions{\criterion}$ & Number of elements of a criterion \criterion\\
			$\numOfSharingScenarios$ & Number of data-sharing scenarios\\
			$\participant$ & An individual\\
			$\sharingScenarioIndex$ & A data-sharing scenario index\\
			$\sharingOptions$ & Number of data-sharing levels\\
			$\selectedSharingOption{\participant}{\sharingScenarioIndex}$ & The selected data-sharing level of individual \participant for a data-sharing scenario \sharingScenarioIndex\\
			$\sharingScenario{\sharingScenarioIndex}$ & A data-sharing scenario\\
			$\decisionMaking{\participant}{\sharingScenario{\sharingScenarioIndex}}$ & A data-sharing decision function of individual \participant in a data-sharing scenario \sharingScenario{\sharingScenarioIndex}\\
			$\criterionDimension{\sharingScenarioIndex}{\criterion}$ & An element of criterion \criterion in a data-sharing scenario \sharingScenarioIndex \\
			$\numOfParticipants$ & Number of individuals\\
			$\criterionWeight{\participant}{\criterion}$ & The weight of criterion \criterion by an individual \participant\\
			$\criterionDimensionIndex$ & The index of an element of a data-sharing criterion\\
			$\criterionDimensionWeight{\participant}{\criterionDimensionIndex}{\criterion}$ & The weight of an element \criterionDimensionIndex of a criterion \criterion by an individual \participant\\
			$\weightSharingScenario{\participant}{\sharingScenarioIndex}$ & The weight of a data-sharing scenario \sharingScenarioIndex by an individual \participant\\
			$\budget$ & Maximum (monetary) budget\\
			$\participationBudget$ & Rewards for participation\\
			$\sharingBudget$ & Rewards for data sharing\\
			$\maxRewardsSharingScenario{\participant}{\sharingScenarioIndex}$ & The maximum rewards of individual \participant for a data-sharing scenario \sharingScenarioIndex\\
			$\weightSharingScenarios{\participant}$ & The total weight of all data-sharing scenarios by an individual \participant\\
			$\rewardsSharingScenario{\participant}{\sharingScenarioIndex}$ & The actual rewards of an individual \participant for a data-sharing scenario \sharingScenarioIndex\\
			$\privacy{\participant}$ & The privacy level of an individual \participant derived from the data-sharing choices\\
			$\coefficient{\criterion}{\criterionDimensionIndex}$ & The coefficient of a data-sharing element $\criterionDimensionIndex$ in the criterion $\criterion$\\
			$\dummyVariable{\criterion}{\criterionDimensionIndex}$ & The dummy variable for the absence or presence of the data-sharing element $\criterionDimensionIndex$ in the criterion $\criterion$\\
			$\regressionError$ & The error of the regression model \\
			$\criterionUtility{\criterion}$ & The partworth utility (relative importance) of criterion $\criterion$\\
			$\elementUtility{\criterion}{\criterionDimensionIndex}$ & The partworth utility (relative importance) of element $\criterionDimensionIndex$ in criterion $\criterion$ among all criteria \\
			$\elementUtilityPerCriterion{\criterion}{\criterionDimensionIndex}$ & The partworth utility (relative importance) of element $\criterionDimensionIndex$ within criterion $\criterion$\\
			$\privacyScenario{\sharingScenarioIndex}$ & The mean privacy level of a data-sharing scenario $\sharingScenarioIndex$ \\
			$\varepsilon$ & The mismatch (absolute error) of data sharing from a privacy-preservation goal signal \\
			$\rewardScenario{\sharingScenarioIndex}$ & The mean rewards level of a data-sharing scenario $\sharingScenarioIndex$ \\
			$\rewardsSharingScenarios{\participant}$ & The rewards of individual \participant gained over the data-sharing scenarios \\
			$\rewardsSharingScenariosIntrinsic{\participant}$ & The hypothetical rewards of an individual \participant gained over the data-sharing scenarios under intrinsic data sharing \\
			$\privacyCost{\participant}{\rewardsSharingScenarios{\participant}}$ & The privacy cost of a data-sharing plan generated by individual \participant as a function of $r_{i}$ \\
			$\alpha, \beta$ & The weights of privacy unfairness and privacy cost respectively in the optimization cost function \\
			\hline
		\end{tabular}\label{table:math-symbols}
			}
\end{table}

\subsection{Data-sharing criteria}\label{subsec:criteria}

Let \numOfCriteria factors, referred to as \emph{criteria}, govern the level of data sharing that an individual, i.e. a \emph{citizen}, chooses. This ranges from sharing no data to sharing all locally available data in an individual's device such as a smartphone. Each criterion $\criterion \in \{1,...,\numOfCriteria\}$ has a number of possible \emph{elements} \numOfCriterionDimensions{\criterion}. For instance, the type of sensor data is a criterion with the following elements (see Figure~\ref{fig:scenarios} in the main paper): GPS location, light sensor, etc. The former element may be regarded more privacy intrusive than the latter one. The total number:

\begin{equation}\label{eq:scenarios}
	\numOfSharingScenarios=\prod _{\criterion=1}^{\numOfCriteria} \numOfCriterionDimensions{\criterion},
\end{equation}

\noindent of combinations between the \numOfCriterionDimensions{\criterion} elements of the \numOfCriteria criteria define the \emph{scenarios} of data sharing, which are the ones studied in this paper. For each data-sharing scenario $\sharingScenarioIndex \in \{1,...,\numOfSharingScenarios\}$, individuals have a number of \sharingOptions discrete \emph{data-sharing options}, where the first option corresponds to sharing all collected data, whereas the $\sharingOptions$th option corresponds to sharing no data. Each individual \participant selects a \emph{data-sharing level} $\selectedSharingOption{\participant}{\sharingScenarioIndex} \in \{1,...,\sharingOptions\}$ for scenario \sharingScenarioIndex. For simplicity, assume that the actual level of data sharing decreases linearly from $1$ to \sharingOptions by, for instance, averaging, obfuscating or resampling the data to share (e.g. with a period proportional to \selectedSharingOption{\participant}{\sharingScenarioIndex}). The data-sharing level \selectedSharingOption{\participant}{\sharingScenarioIndex} is a result of a function:

\begin{equation}\label{eq:choice}
	\selectedSharingOption{\participant}{\sharingScenarioIndex}=\decisionMaking{\participant}{\sharingScenario{\sharingScenarioIndex}},
\end{equation}

\noindent where \sharingScenario{\sharingScenarioIndex}=($\criterionDimension{\sharingScenarioIndex}{\criterion})_{\criterion=1}^{\numOfCriteria}$ represents the \emph{data-sharing scenario} \sharingScenarioIndex as the sequence of elements $\criterionDimension{\sharingScenarioIndex}{\criterion} \in \{1,...,\numOfCriterionDimensions{\criterion}\}$ over all \numOfCriteria criteria. For the sake of simplicity in the model illustration, the number of criteria \numOfCriteria and the number of elements \numOfCriterionDimensions{\criterion} for each criterion \criterion are assumed finite and fixed for all \numOfParticipants individuals. 

\subsection{A weighting scheme for personalized privacy valuation}\label{subsec:valuation}

Let the weight $\criterionWeight{\participant}{\criterion} \in [0,1]$ denote how privacy-sensitive a criterion \criterion is for an individual $\participant$ relative to the rest of the criteria, such that $\sum_{\criterion=1}^{\numOfCriteria}\criterionWeight{\participant}{\criterion}=1$. Similarly, the weight $\criterionDimensionWeight{\participant}{\criterionDimensionIndex}{\criterion} \in  [0,1]$ denotes how privacy-sensitive an individual \participant finds the element $\criterionDimensionIndex \in \{1,...,\numOfCriterionDimensions{\criterion}\}$ of criterion \criterion relative to the rest of the elements, such that $\sum_{\criterionDimensionIndex=1}^{\numOfCriterionDimensions{\criterion}}\criterionDimensionWeight{\participant}{\criterionDimensionIndex}{\criterion}=1$.

The weight \weightSharingScenario{\participant}{\sharingScenarioIndex} of a data-sharing scenario \sharingScenarioIndex is determined by each criterion weight \criterionWeight{\participant}{\criterion} and each element weight \criterionDimensionWeight{\participant}{\criterionDimensionIndex}{\criterion} it consists of as follows:

\begin{equation}\label{eq:scenario-weight}
	\weightSharingScenario{\participant}{\sharingScenarioIndex}=\sum_{\criterion=1}^{\numOfCriteria}\criterionWeight{\participant}{\criterion}\cdot \criterionDimensionWeight{\participant}{\criterionDimensionIndex}{\criterion},
\end{equation}

\noindent where $\criterionDimensionIndex=\criterionDimension{\sharingScenarioIndex}{\criterion}$ is the element of criterion \criterion in the data-sharing scenario \sharingScenarioIndex. 

The weighting scheme is used to model the heterogeneity in the availability of data that stems from the individuals' privacy perception, i.e. it is expected that privacy-sensitive data are more scarce and as a result they are also expected to have higher value in data sharing.

\subsection{Calculating rewards and privacy}\label{subsec:privacy-rewards-calcculations}

The calculation of rewards and privacy relies on the weighting scheme for personalized privacy valuation (Section~\ref{subsec:valuation}). Assume there is a maximum (monetary) budget \budget to incentivize data sharing that is split as follows:

\begin{equation}\label{eq:budget}
\budget=\participationBudget+\sharingBudget,
\end{equation}

\noindent where \participationBudget rewards participation, meaning the cognitive effort required for individuals to make choices for all data-sharing scenarios and \sharingBudget rewards the actual data sharing respectively. Moreover, assume that the weights of each cri\-te\-rion/element represent the actual intrinsic privacy concerns of individuals. The maximum \emph{rewards} \maxRewardsSharingScenario{\participant}{\sharingScenarioIndex} of an individual \participant for each data-sharing scenario \sharingScenarioIndex are allocated according to self-determined privacy-intrusion level of the data-sharing scenario as follows:

\begin{equation}\label{eq:scenario-max-rewards}
	\maxRewardsSharingScenario{\participant}{\sharingScenarioIndex}=\frac{\weightSharingScenario{\participant}{\sharingScenarioIndex}}{\weightSharingScenarios{\participant} \cdot \sharingBudget},
\end{equation}

\noindent where the weight \weightSharingScenarios{\participant} sums up the weights of all scenarios as follows:

\begin{equation}\label{eq:total-weight}
	\weightSharingScenarios{\participant}=\sum_{\sharingScenarioIndex=1}^{\numOfSharingScenarios}\weightSharingScenario{\participant}{\sharingScenarioIndex}. 
\end{equation}

\noindent The actual received rewards of an individual \participant with a data-sharing level \selectedSharingOption{\participant}{\sharingScenarioIndex} under a data-sharing scenario \sharingScenarioIndex are calculated as follows:

\begin{equation}\label{eq:scenario-rewards}
	\rewardsSharingScenario{\participant}{\sharingScenarioIndex}=\frac{\sharingOptions-\selectedSharingOption{\participant}{\sharingScenarioIndex}}{\sharingOptions-1} \cdot \maxRewardsSharingScenario{\participant}{\sharingScenarioIndex}.
\end{equation}

\noindent The \emph{privacy} of an individual \participant over all selections made in the \numOfSharingScenarios data-sharing scenarios is calculated as follows:

\begin{equation}\label{eq:actual-privacy}
	\privacy{\participant}=\frac{1}{\numOfSharingScenarios}\sum_{\sharingScenarioIndex=1}^{\numOfSharingScenarios} \frac{\selectedSharingOption{\participant}{\sharingScenarioIndex}-1}{\sharingOptions-1}.
\end{equation}

\section{Recruitment Process}\label{sec:recruitment}

The split of the recruitment process into multiple sessions as well as the invitation for the recruitment are illustrated in this section. 

\subsection{Recruitment sessions}\label{subsec:sessions}

Splitting the recruitment of participants and the experiment into multiple sessions serves the following: (i) Guaranteeing enough time to recruit participants from the pool. (ii) Having a manageable number of participants to moderate during the experimental process. (iii) Scale up the number of participants incrementally so that potential failures do not influence the overall experiment. The entry phase takes place on Mondays, the core phase during Mondays-Wednesdays and the exit phase on Thursdays.

\begin{table}[!htb]
	\caption{Recruitment during the 8 experimental sessions performed.}\label{tab:sessions}
	\centering
	\resizebox{\columnwidth}{!}{%
		\begin{tabular}{lllllllll}
			\toprule
			\textbf{Session:} & \textbf{1} & \textbf{2}& \textbf{3} & \textbf{4} & \textbf{5} & \textbf{6} & \textbf{7} &\textbf{8} \\
			\midrule
			\textbf{Entry Phase} & 3.10.2016 & 17.10.2016 & 31.10.2016 & 7.11.2016 & 14.11.2016 & 21.11.2016 & 28.11.2016 & 5.12.2016  \\
			\textbf{Core Phase} & 3-5.10.2016 & 17-20.10.2016 & 31.10-2.11.2016 & 7-9.11.2016 & 14-16.11.2016 & 21-23.11.2016 & 28-30.11.2016 & 5-7.12.2016  \\
			\textbf{Exit Phase} & 6.10.2016 & 21.10.2016 & 3.11.2016 & 10.11.2016 & 17.11.2016 & 24.11.2016 & 1.12.2016 & 8.12.2016  \\
			\textbf{Num. of Participants} & 15 & 13 & 11 & 16 & 15 & 13 & 19 & 21  \\
			\textbf{Compensations} (CHF) & 666.0 & 813.0 & 746.0 & 840.0 &  943.0 & 805.0 & 1259.0 & 1283.0  \\
			\bottomrule
		\end{tabular}
	}
\end{table}

A 93.6\% of the participants did not know about the experiment before participating (Question D.28 in Table~\ref{table:exit-phase-experiment}). 

\subsection{E-mail invitation for recruitment}\label{subsec:invitation}

The invitation sent to the DeSciL pool of participants for the recruitment is presented below: 

\begin{quote}
	\begin{alltt}
		Dear <firstname> <lastname>,
		
		We would like to invite you to an upcoming experiment 
		'<experiment name>'. The experiment will be carried out
		in English, so you should be fluent in English in order
		to register for this study. 
		
		The experiment requires your participation at the 
		ETH Decision Science Laboratory at TWO different days and
		the use of your mobile phone (Android only) at other two
		days to answer some questions.
		
		Your participation in the experiment will be maximally
		compensated as follows:
		
		Session 1: CHF 25.-
		Core phase on mobile phone: Up to CHF 35.-
		Session 2: CHF 15.-
		Total: Up to CHF 75.-
		
		You MUST attend both lab sessions in order to receive
		your payment. Furthermore, the following criteria are
		a requirement:
		
		1) have and use an Android mobile phone, version 4.4
		and above
		2) have mobile Internet connection
		3) Keep your phone switched on and adequately charged
		throughout the experiment
		4) officially register for the study
		5) arrive on time for the experiment at both days
		6) install and use a mobile application to answer some
		question at two days
		7) fulfill all experimental criteria including specified
		language proficiency
		8) provide photo identification.
		
		The sessions are scheduled as follows:
		<session list>
		
		If you want to participate, you can register by clicking on the following link:
		<link>
		
		(If you cannot click on the link, copy it to the clipboard by selecting it, 
		right click and choosing "Copy", 
		and then paste it into the address line in your browser
		by right clicking there and choosing "Paste".)
		
		Kind regards
		ETH Decision Science Laboratory (DeSciL)
		\url{http://www.descil.ethz.ch/contact/}
	\end{alltt}
	
\end{quote}

\section{Experimental Design}\label{sec:experimental-design}

The preparatory, entry, core and exit phase of the conducted experiment are outlined here in more detail. The compensation and monetary incentives introduced to engage participants are also illustrated. 

\subsection{Preparatory phase}\label{subsubsec:preparatory-phase}

The preparatory phase has a supportive role in the overall experiment as participants are neither compensated nor selected rigorously. Participants of the preparatory phase are selected from the network of employees at ETH Zurich (convenience sampling). The findings of the preparatory phase are not conclusive and mainly serve the design of the following phases. Nevertheless, this phase was scaled up to approximately 200 participants within 3 months, starting on 19.05.2016. 

The preparatory phase consists of a web survey implemented in Qualtrics~\cite{Qualtrics2021} with the questions outlined in Table~\ref{table:preparatory-phase}. The goal of the preparatory phase is to provide some first insights about the perception of privacy from the perspective of the three studied aspects: sensor type, data collector and context. Questions~A.9-A.14 are designed for this purpose. Questions~A.6-A.8 provide information about the smartphone usage profiles, whereas, Question~A.15 scrutinizes the type of incentives that motivate participants to share mobile sensor data. Questions~A.1-A.5 collect demographic information.

\begin{table}[!htb]
	\caption{Survey questions for the preparatory phase.}\label{table:preparatory-phase}
	\centering
	\resizebox{\textwidth}{!}{%
		\begin{tabular}{p{0.15\textwidth}p{0.9\textwidth}p{0.19\textwidth}p{0.39\textwidth}}
			\toprule
			\textbf{ID} & \textbf{Question} & \textbf{Type}& \textbf{Options} \\
			\midrule
		    A.1 & What is your gender? & multiple choice, one selection & female, male \\\midrule
			A.2 & Which year were you born? & multiple choice, one selection & 81 [1920,2000]  \\\midrule
			A.3 & In which country have you lived most of your life? & multiple choice, one selection & all countries  \\\midrule
			A.4 & What is the highest level of education you have completed? & multiple choice, one selection & less than high school, high school, some college, bachelors degree, masters degree, PhD degree \\\midrule
			A.5 & Which of the following categories best describes your employment status? & multiple choice, multiple selections & employed full time, employed part time, unemployed (looking for work), unemployed (not looking for work), retired, student, disabled\\\midrule
			A.6 & Which types of apps do you usually have on your smartphone? & multiple choice, multiple selections & education, entertainment, finance, game, health \& fitness, medical, music \& audio, news, productivity, shopping, social networking, transportation, travel, utility, weather\\\midrule
			A.7 & How many times do you check your mobile phone during the day (e.g. check notifications/time, open apps, etc.)? & multiple choice, one selection & 1-35, 36-70, 71-100, 101-135, 135+\\\midrule
			A.8 & How concerned are you about the privacy of your mobile sensor data? & ratio scale & 5 [Not at all concerned,extremely concerned]\\\midrule
			A.9 & Which level of privacy intrusion would you assign to the following mobile sensors? & group of questions & 12 questions\\
			\qquad A.9.1 & \qquad Accelerometer (it measures the changes of the velocity of the smartphone) & ratio scale & 5 [very low,very high]\\
			\qquad A.9.2 & \qquad Gyroscope (it measures the rotation/twist of the smartphone) & ratio scale & 5 [very low,very high]\\
			\qquad A.9.3 & \qquad GPS (it measures the geographical location of the smartphone) & ratio scale & 5 [very low,very high]\\
			\qquad A.9.4 & \qquad Proximity Sensor (it measures the physical distance of the smartphone from your body) & ratio scale & 5 [very low,very high]\\
			\qquad A.9.5 & \qquad Ambient Light Sensor (it measures the ambient light level) & ratio scale & 5 [very low,very high]\\
			\qquad A.9.6 & \qquad Battery Sensor (it measures the battery level) & ratio scale & 5 [very low,very high]\\
			\qquad A.9.7 & \qquad Microphone (it measures several sound features. e.g. level of sound frequencies) & ratio scale & 5 [very low,very high]\\
			 \qquad A.9.8 & \qquad Camera & ratio scale & 5 [very low,very high]\\
			 \qquad A.9.9 & \qquad Thermometer (it measures the temperature of the device) & ratio scale & 5 [very low,very high]\\
			 \qquad A.9.10 & \qquad Air Humidity Sensor (it measures the relative humidity in a range 0-100\%) & ratio scale & 5 [very low,very high]\\
			 \qquad A.9.11 & \qquad Barometer (it measures the atmospheric pressure) & ratio scale & 5 [very low,very high]\\
			 \qquad A.9.12 & \qquad Bluetooth (it measures the proximity of the device with other devices) & ratio scale & 5 [very low,very high]\\\midrule
			 A.10 & How important for your privacy is the type of sensor from which you share data? & ratio scale & 5 [not at all important,extremely important]\\\midrule
			 A.11 & Which level of privacy intrusion would you assign to the following stakeholders if you had to share your mobile sensor data with them? & group of questions & 12 questions\\
			 \qquad A.11.1 & \qquad Corporations/companies & ratio scale & 5 [very low,very high]\\
			 \qquad A.11.2 & \qquad Non-profitable/non-governmental organizations & ratio scale & 5 [very low,very high]\\
			 \qquad A.11.3 & \qquad Educational institutes (Public) & ratio scale & 5 [very low,very high]\\
			 \qquad A.11.4 & \qquad Governments and governmental organizations & ratio scale & 5 [very low,very high]\\\midrule
			 A.12 & How important for your privacy is the stakeholder you share your mobile sensor data with? & ratio scale & 5 [not at all important,extremely important]\\\midrule
			 A.13 & Which level of privacy-intrusion would you assign to the following contexts of apps with access to your mobile sensor data? & group of questions & 9 questions\\
			 \qquad A.13.1 & \qquad Education & ratio scale & 5 [very low,very high]\\
			 \qquad A.13.2 & \qquad Entertainment & ratio scale & 5 [very low,very high]\\
			 \qquad A.13.3 & \qquad Environment & ratio scale & 5 [very low,very high]\\
		     \qquad A.13.4 & \qquad Finance & ratio scale & 5 [very low,very high]\\
			 \qquad A.13.5 & \qquad Health & ratio scale & 5 [very low,very high]\\
			 \qquad A.13.6 & \qquad Shopping & ratio scale & 5 [very low,very high]\\
			 \qquad A.13.7 & \qquad Social networking & ratio scale & 5 [very low,very high]\\
			 \qquad A.13.8 & \qquad Training & ratio scale & 5 [very low,very high]\\
			 \qquad A.13.9 & \qquad Transportation/Traveling & ratio scale & 5 [very low,very high]\\\midrule
			 A.14 & How important is for your privacy the context of apps in which you share your mobile sensor data? & ratio scale & 5 [not at all important,extremely important]\\\midrule
			 A.15 & Select one or more incentives which would motivate you to share your mobile sensor data & multiple choice, multiple selections & money, vouchers/discounts on services and stores, free access to additional services (maps, recommended apps, etc.), free access to data, contributing to public good, contributing data if my friends did, contributing data without incentives  \\
			\bottomrule
		\end{tabular}
	}
\end{table}

\subsection{Entry phase}\label{subsubsec:entry-phase}

\begin{figure}[!htb]
	\centering
	\includegraphics[width=0.31\textwidth]{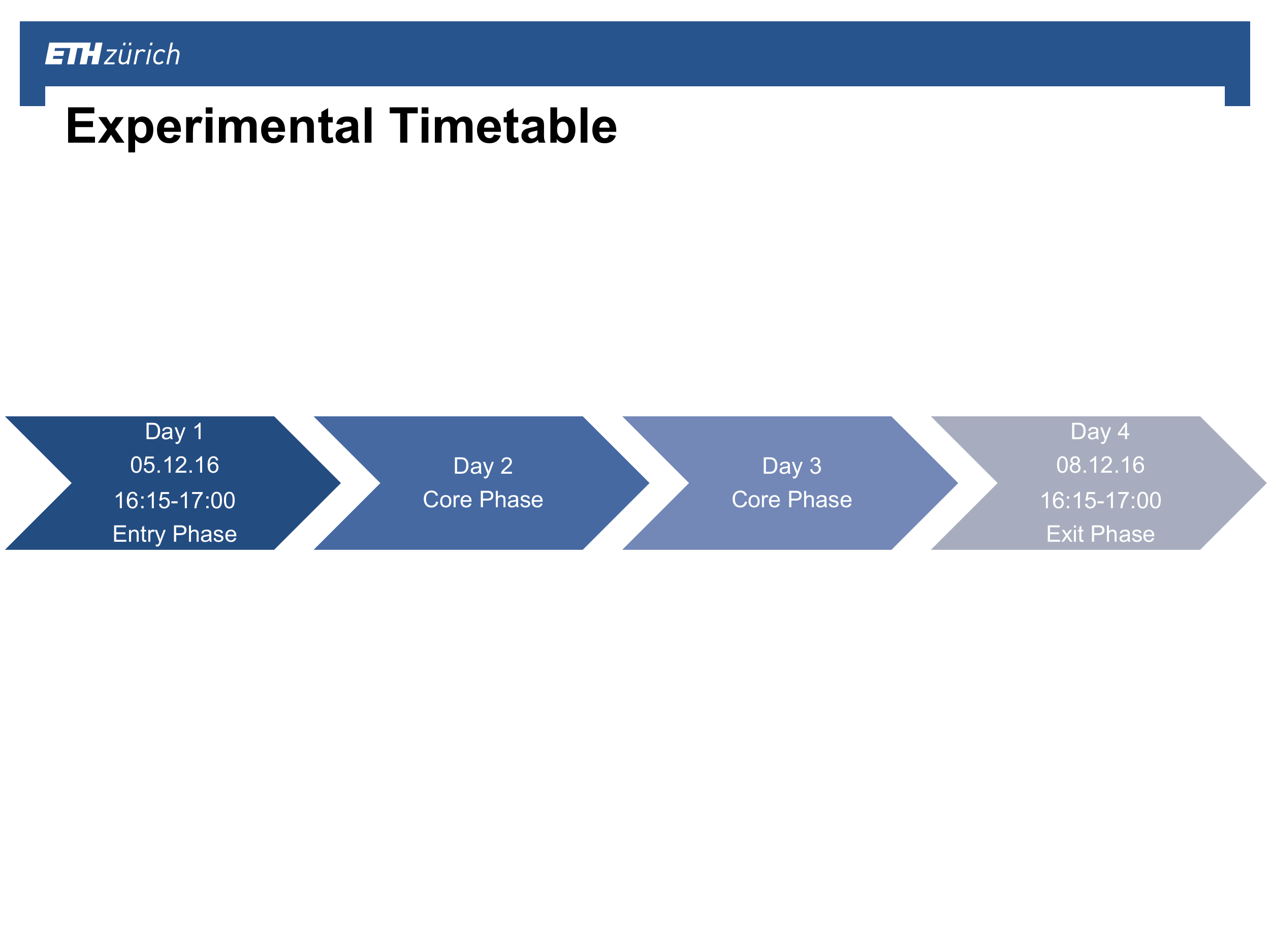}
	\includegraphics[width=0.31\textwidth]{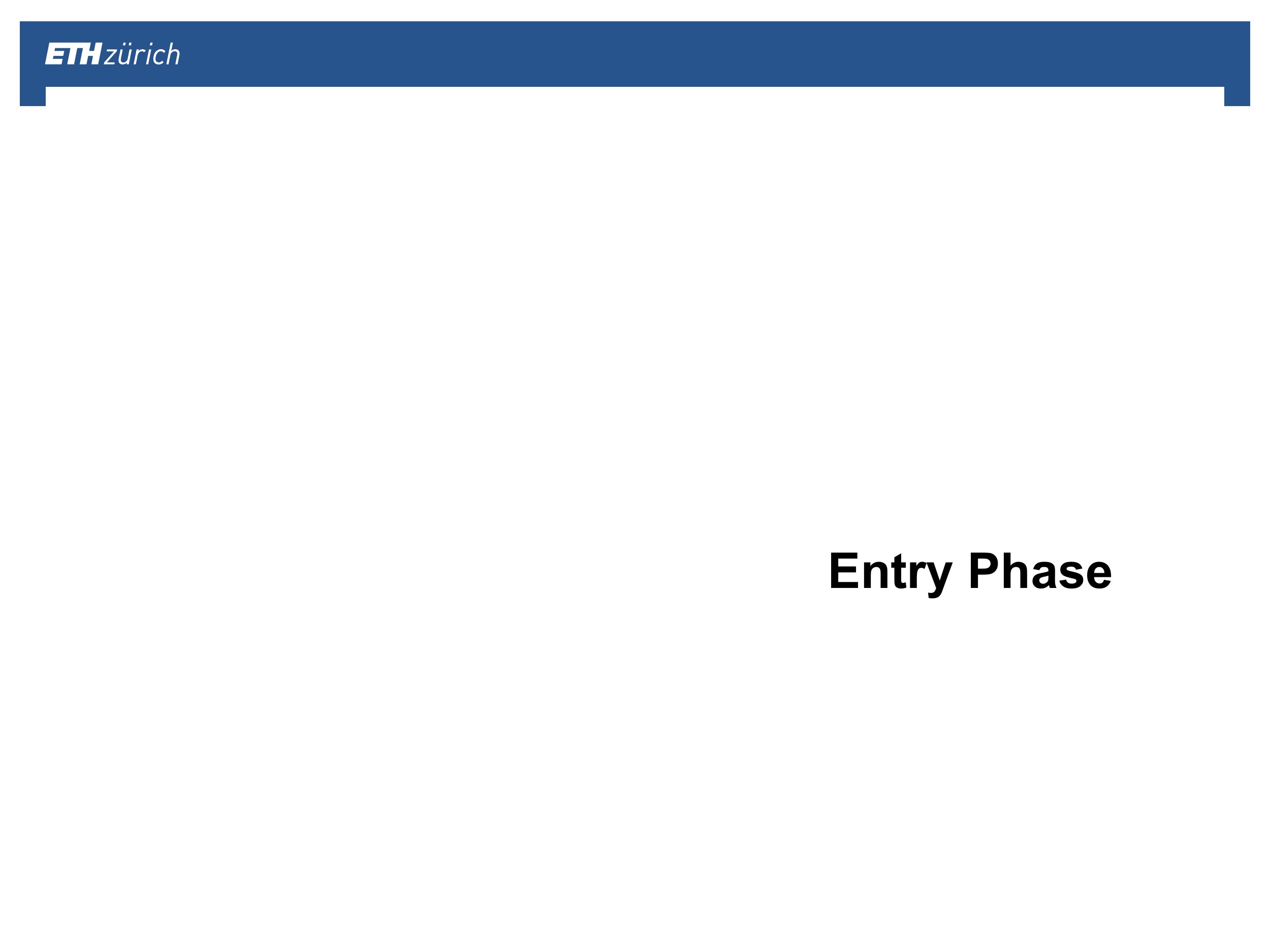}
	\includegraphics[width=0.31\textwidth]{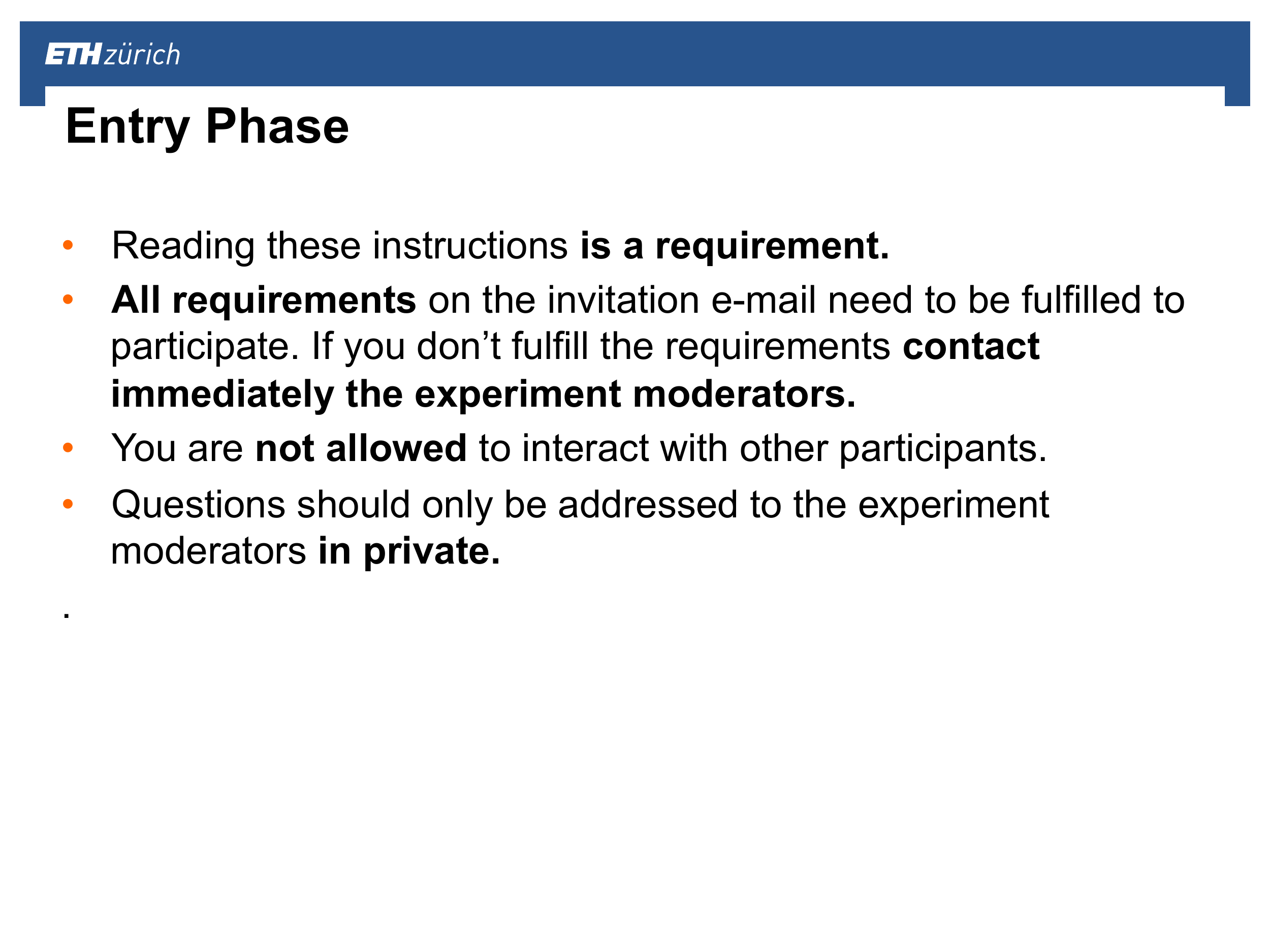}
	\includegraphics[width=0.31\textwidth]{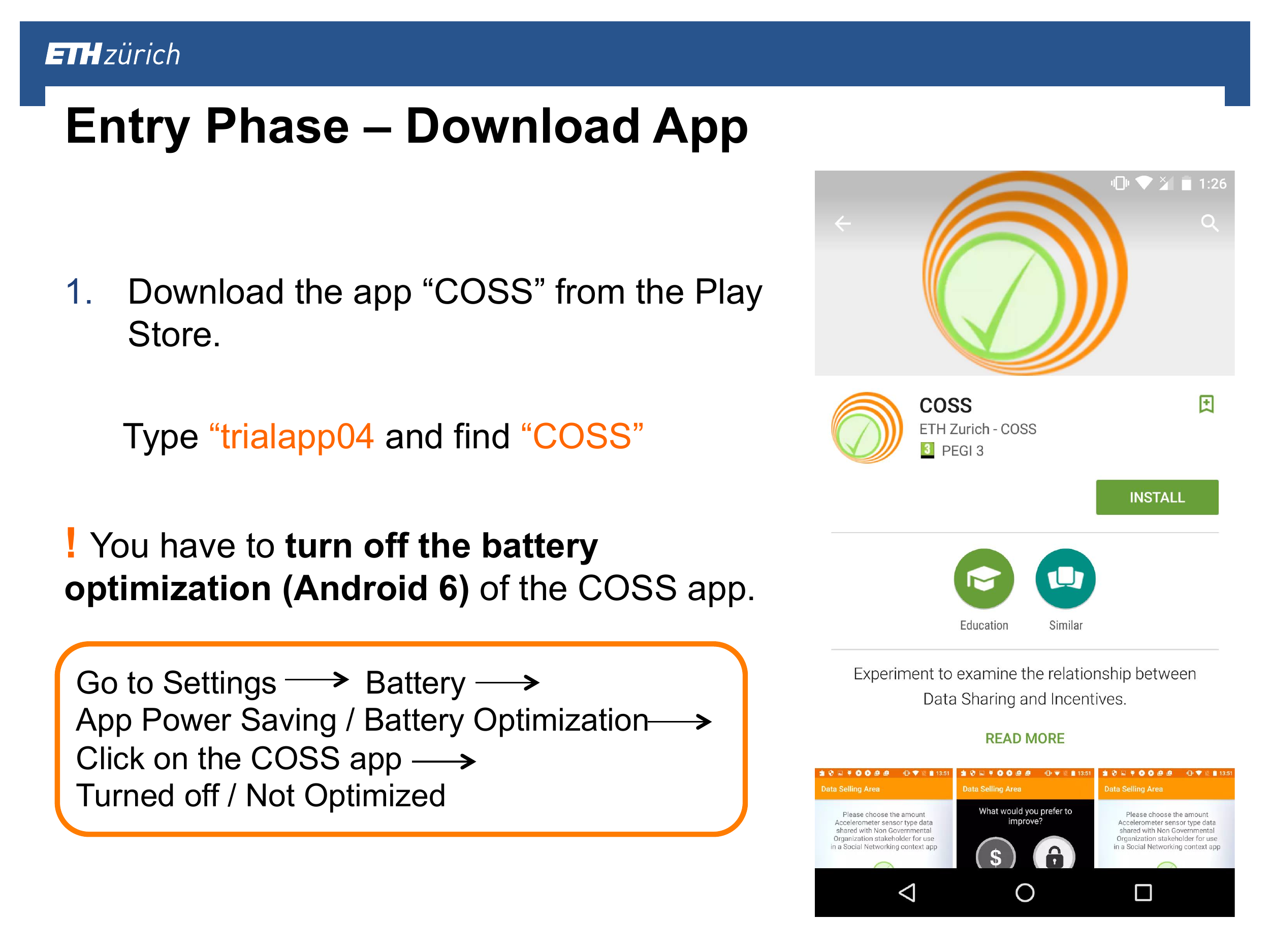}
	\includegraphics[width=0.31\textwidth]{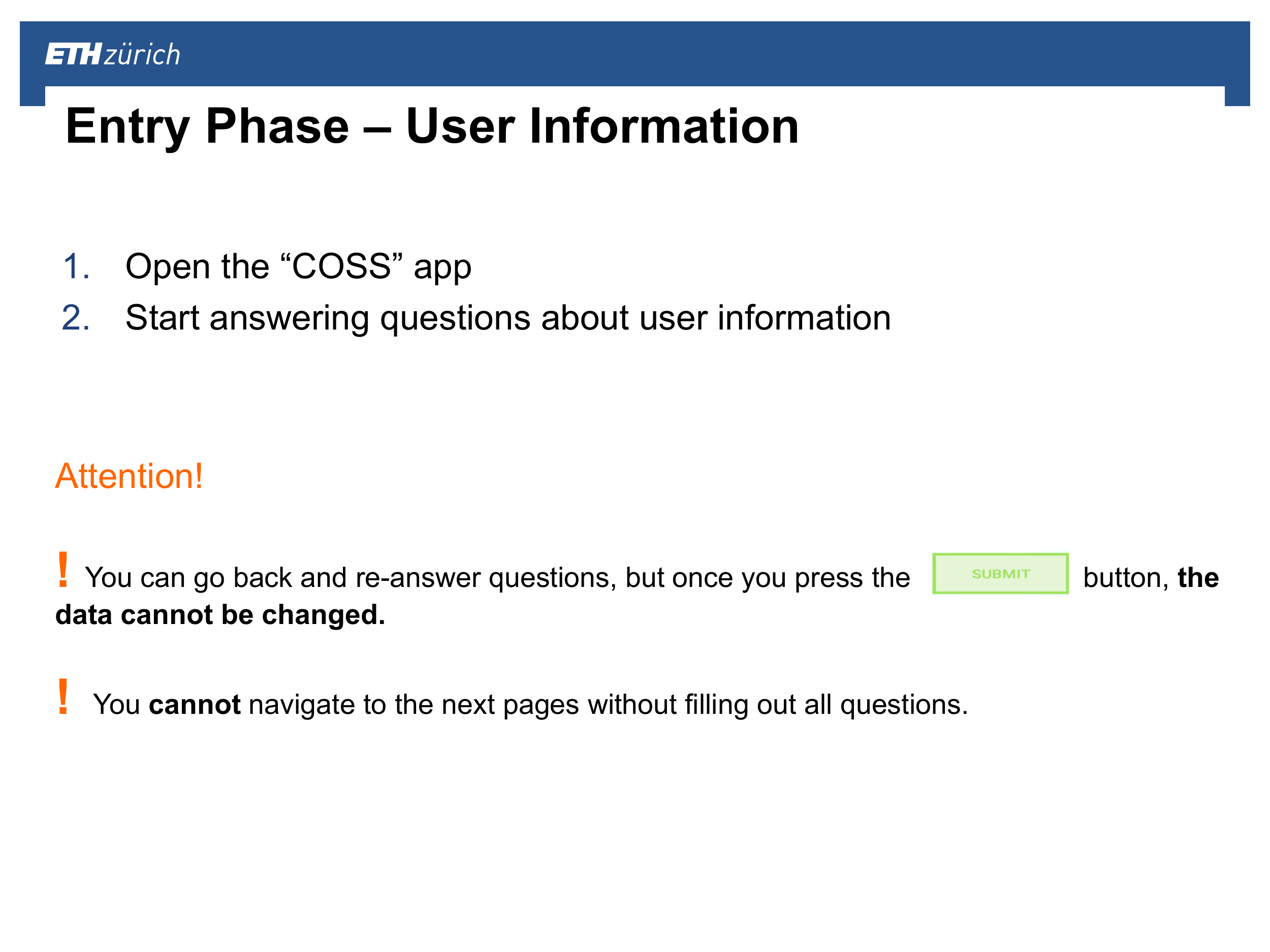}
	\includegraphics[width=0.31\textwidth]{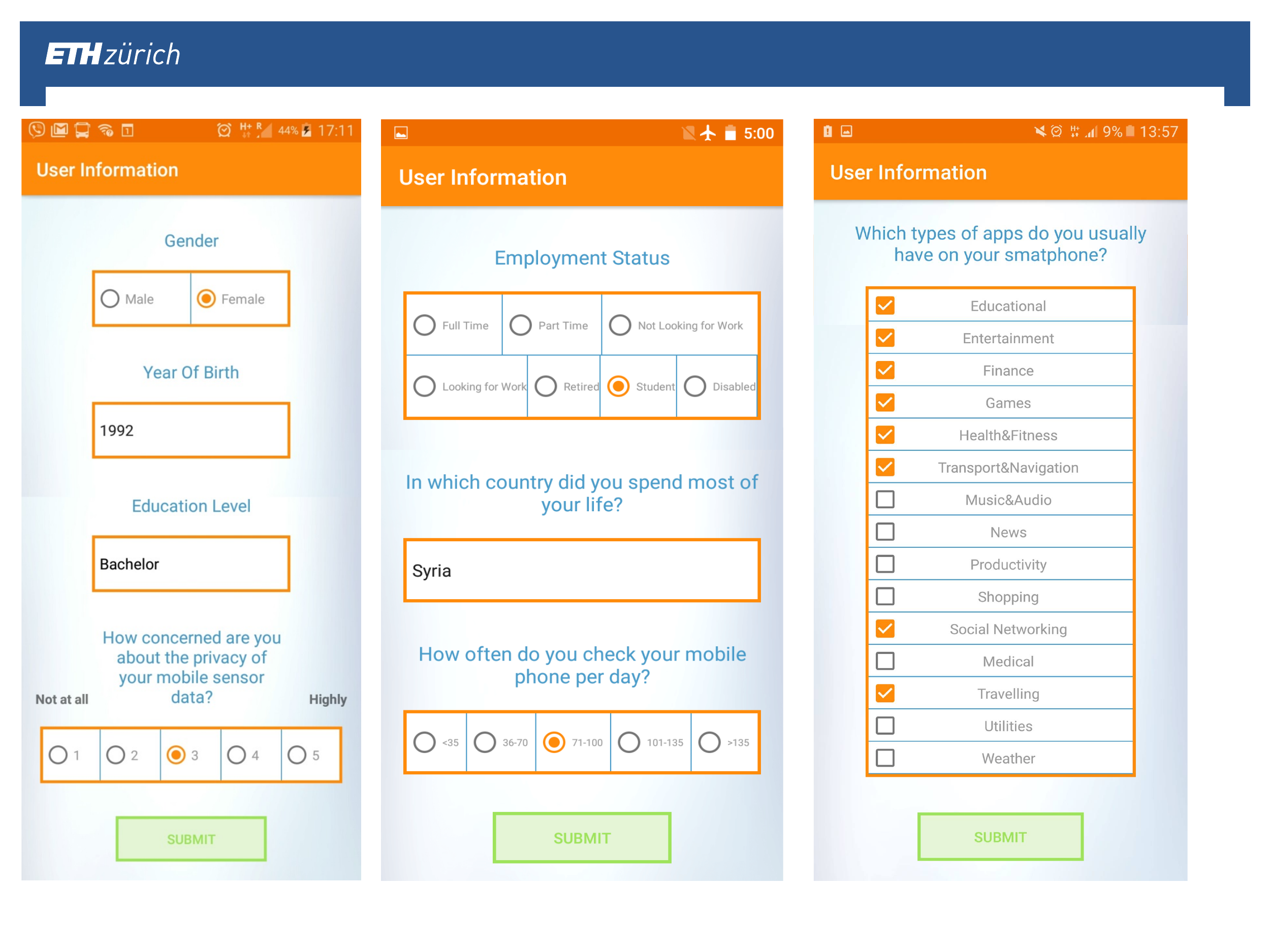}
	\includegraphics[width=0.31\textwidth]{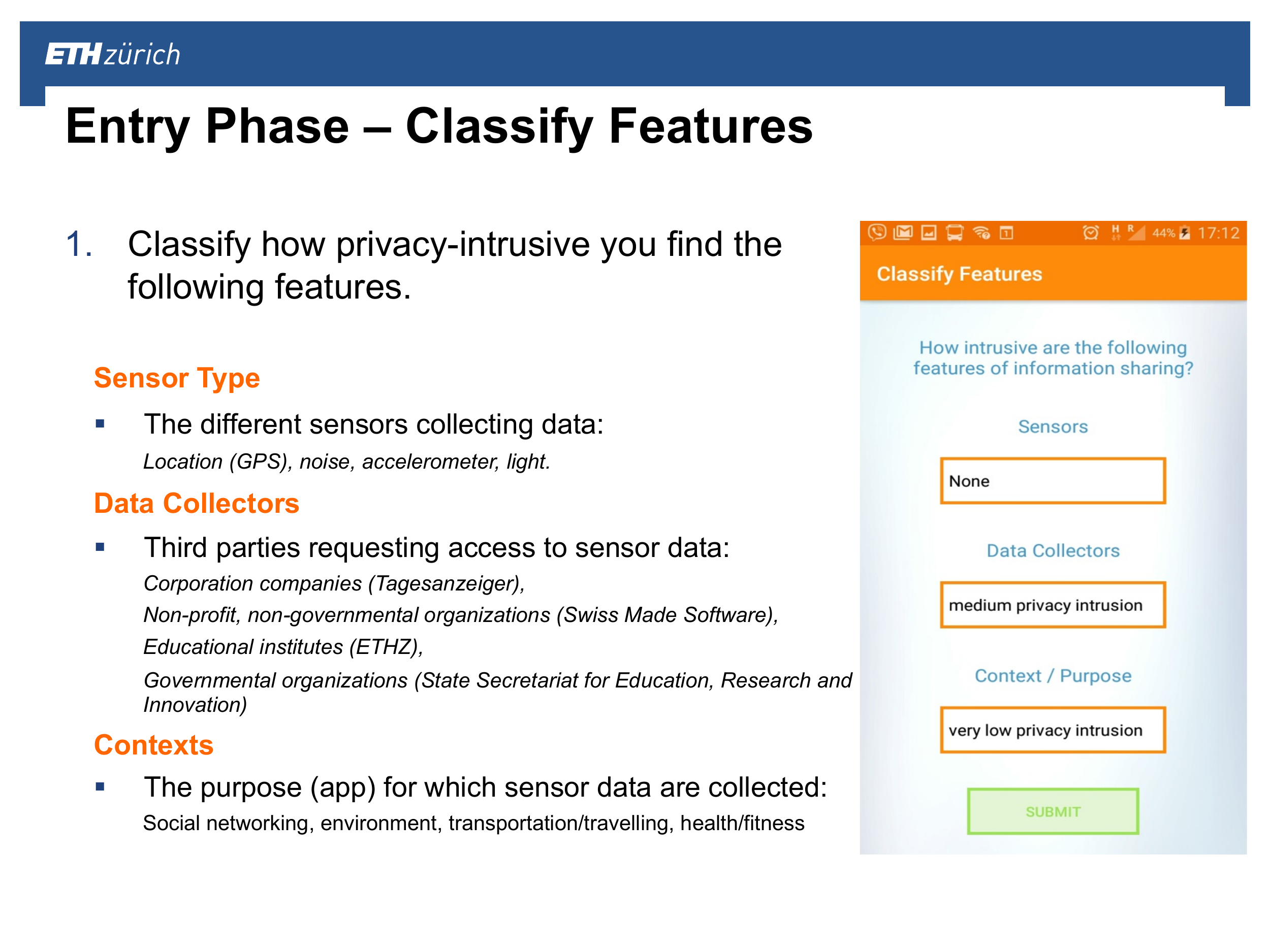}
	\includegraphics[width=0.31\textwidth]{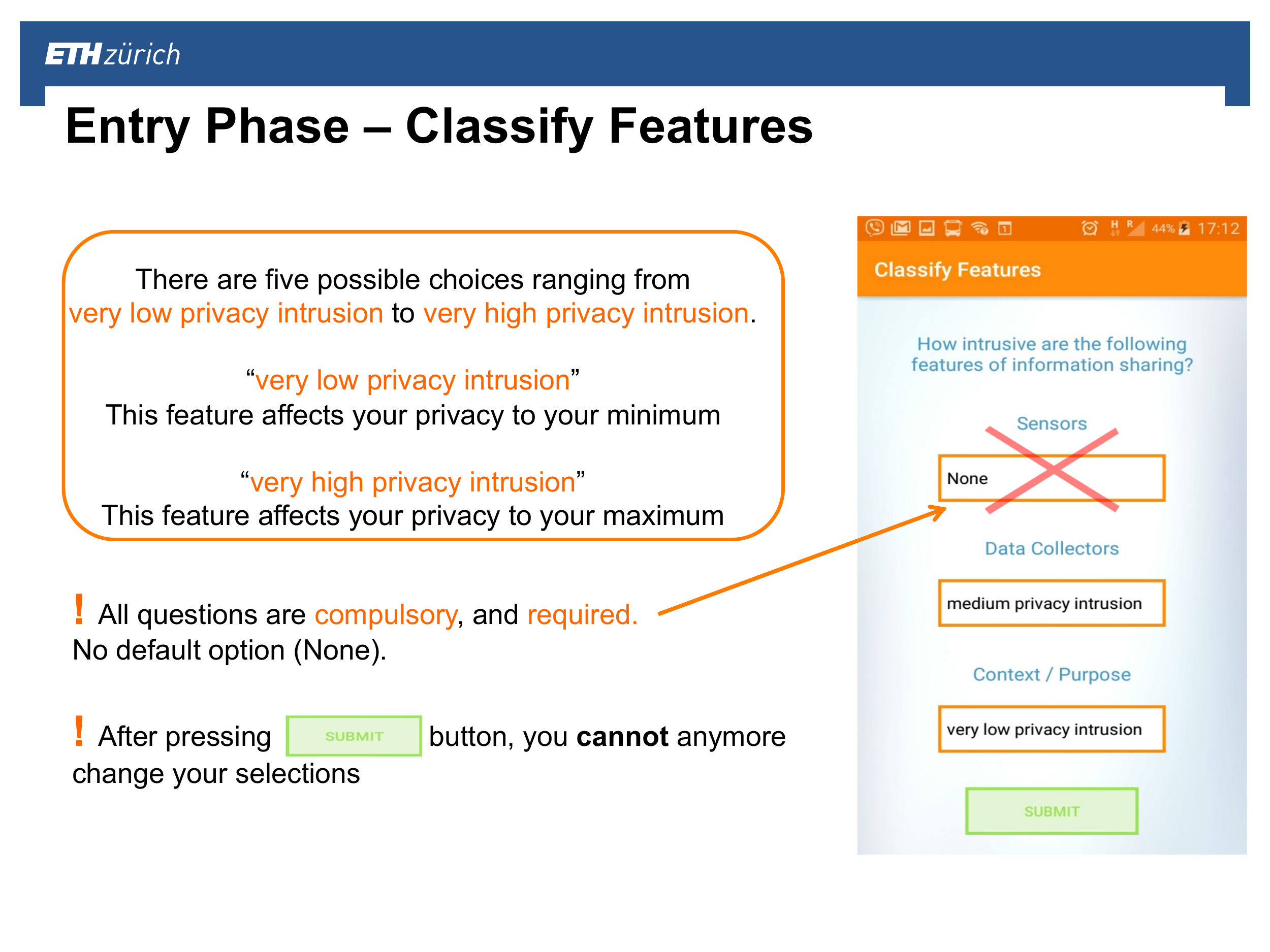}
	\includegraphics[width=0.31\textwidth]{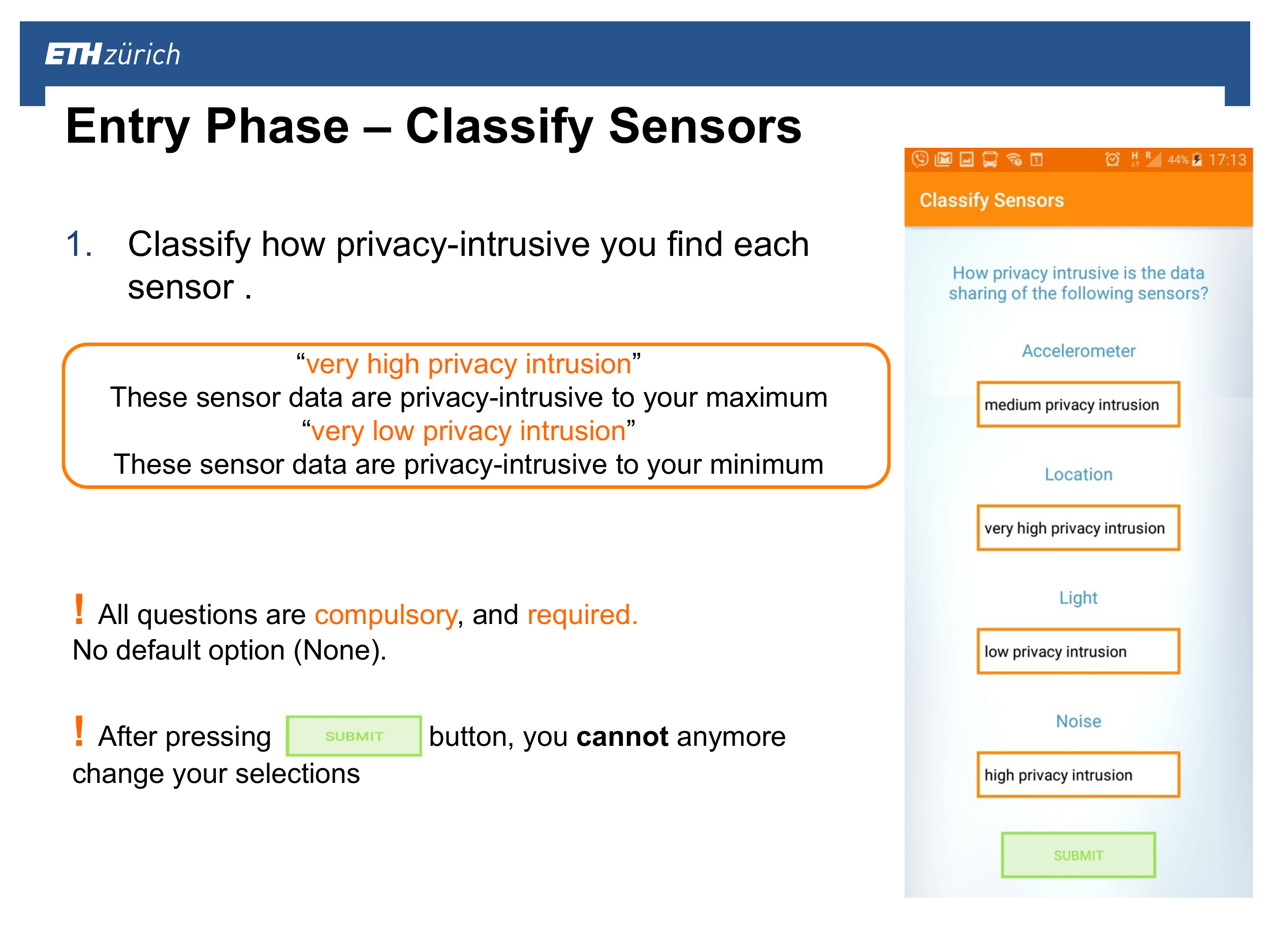}
	\includegraphics[width=0.31\textwidth]{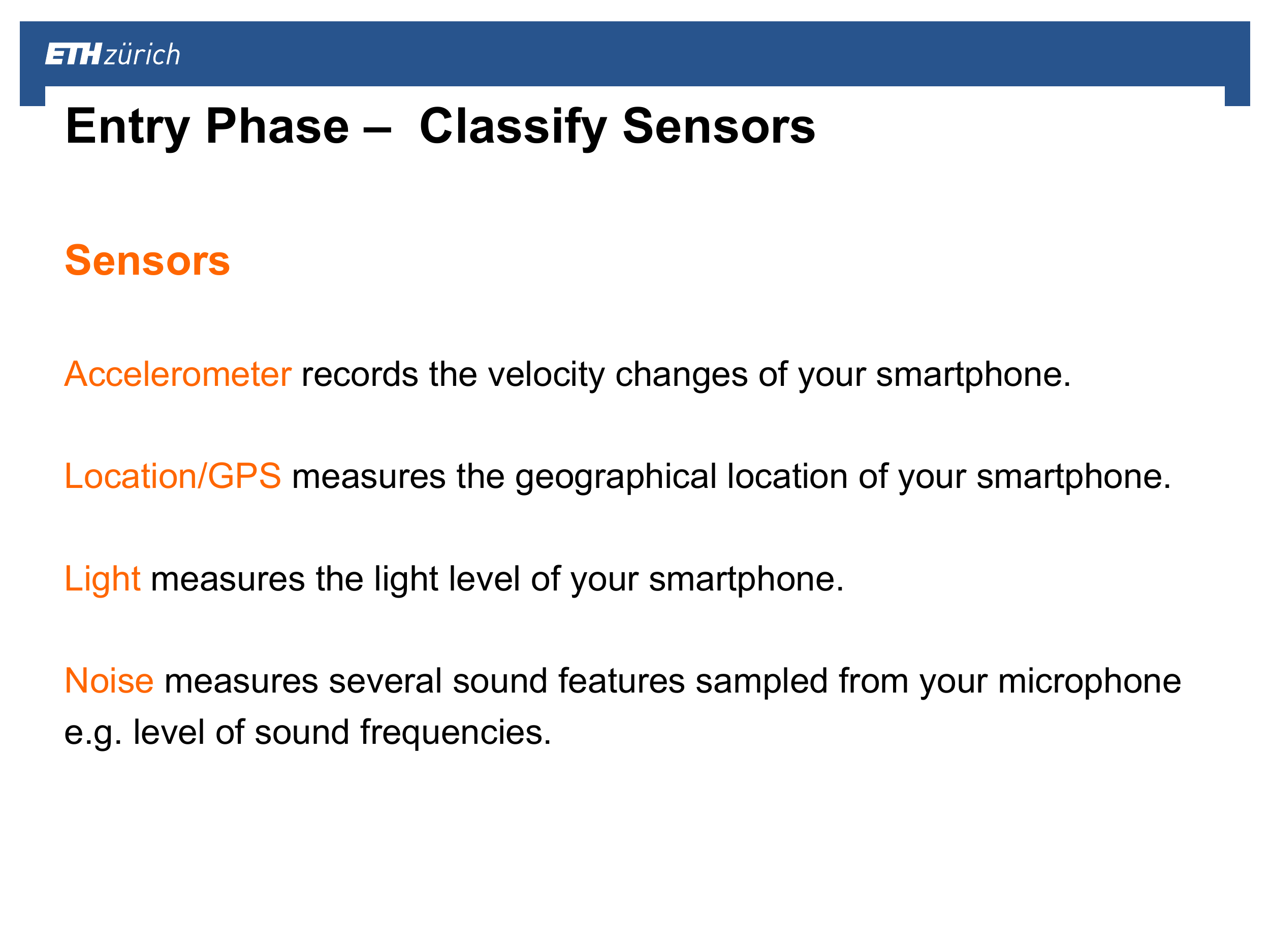}
	\includegraphics[width=0.31\textwidth]{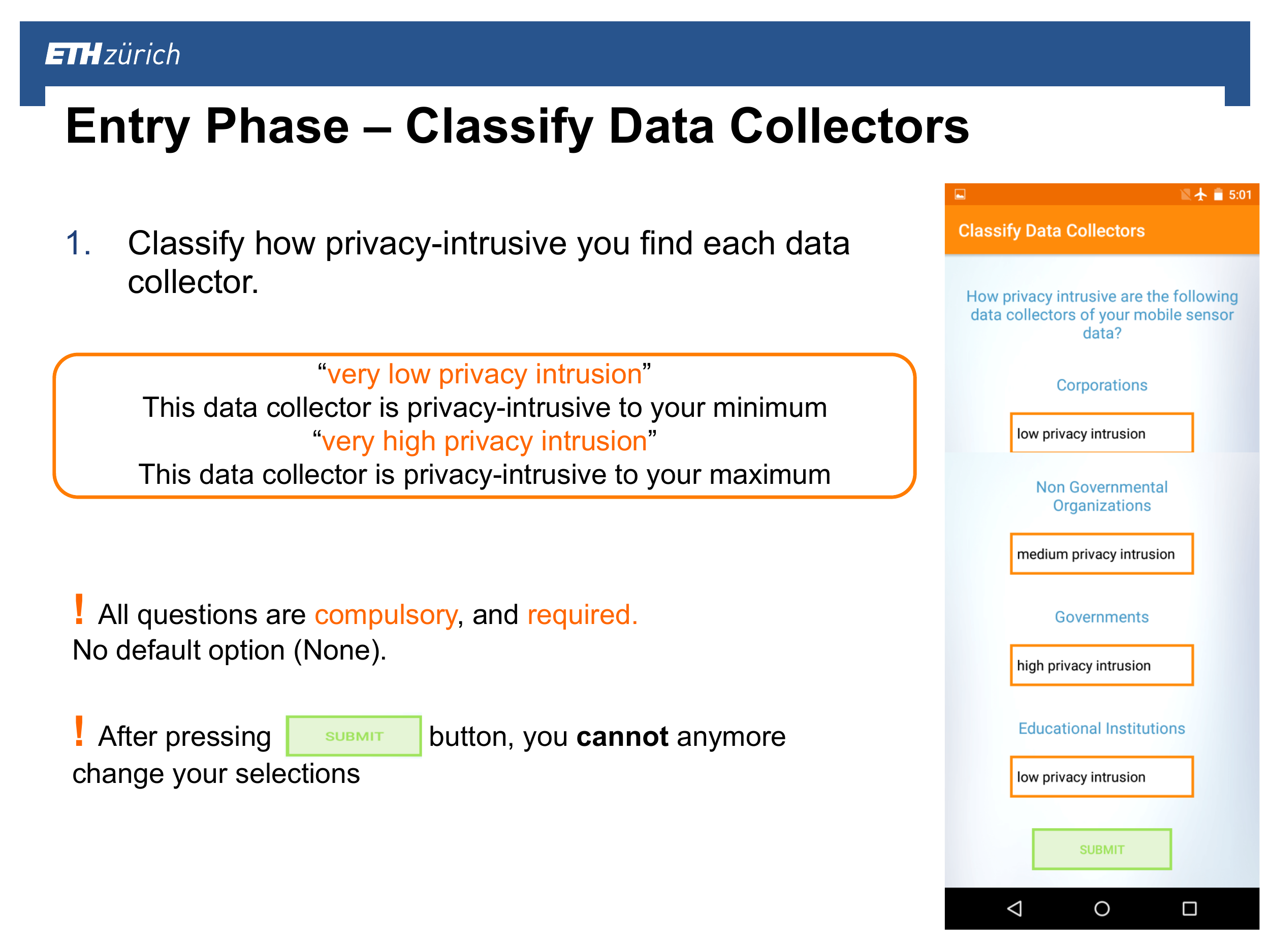}
	\includegraphics[width=0.31\textwidth]{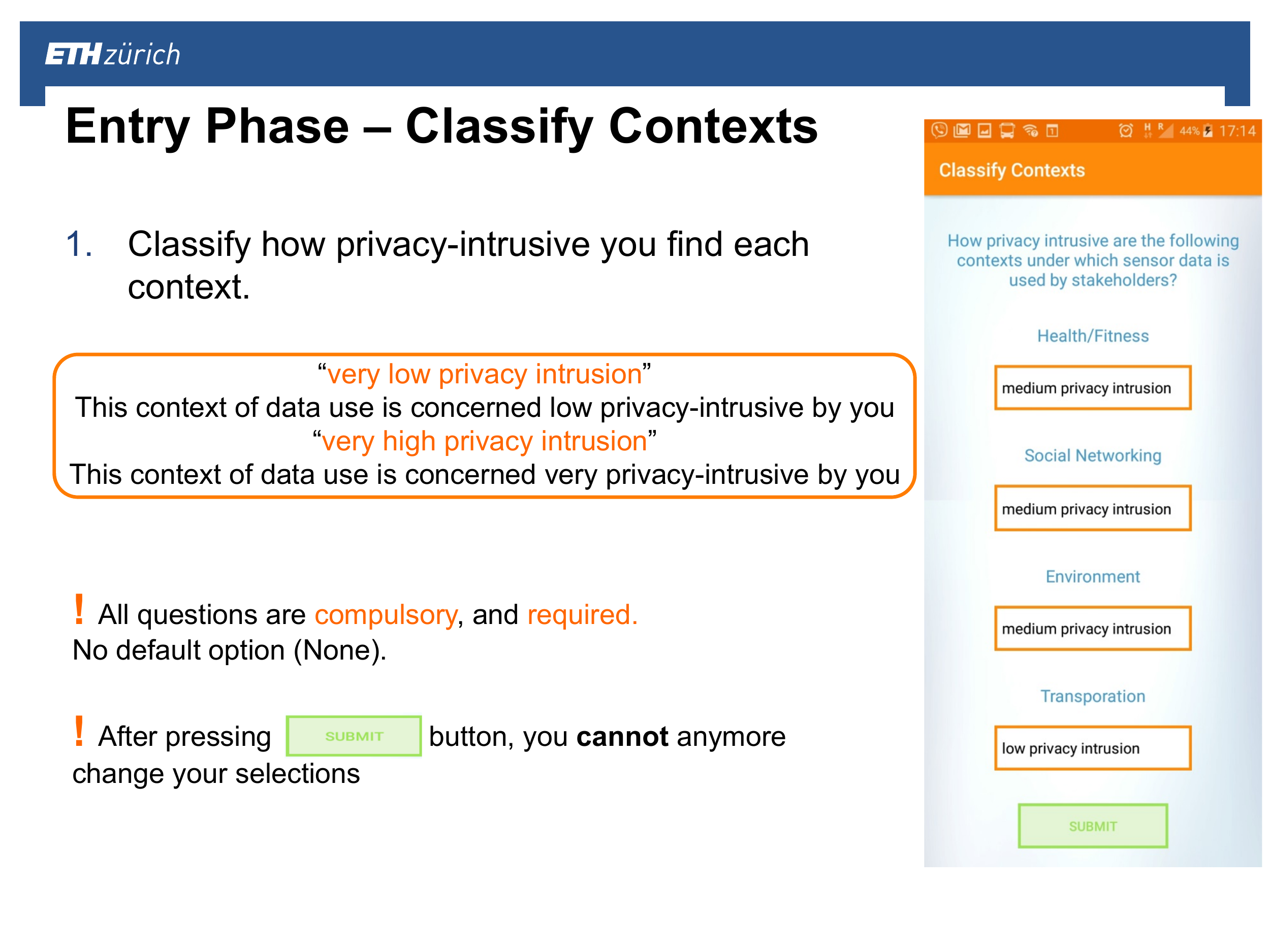}
	\includegraphics[width=0.31\textwidth]{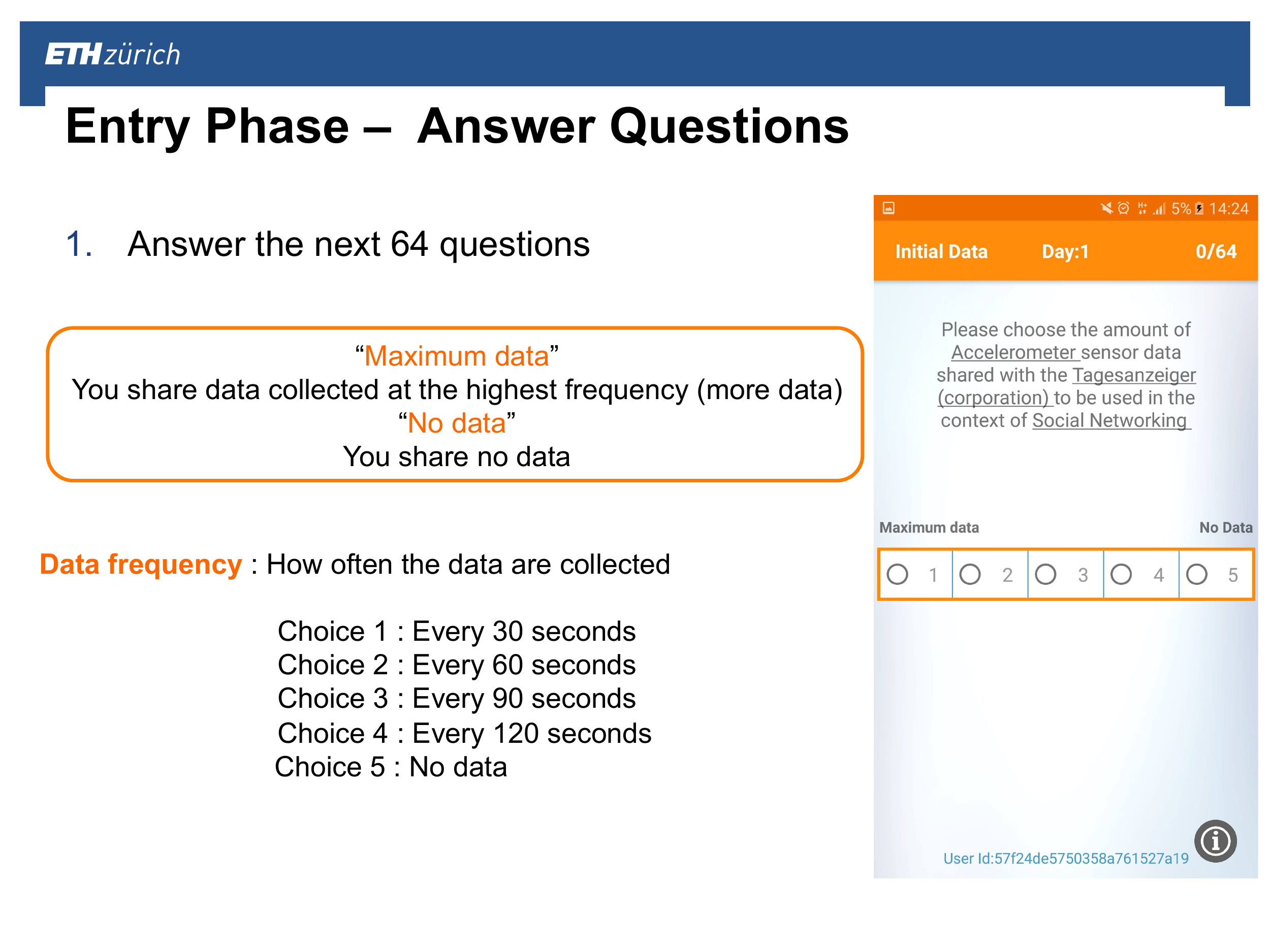}
	\includegraphics[width=0.31\textwidth]{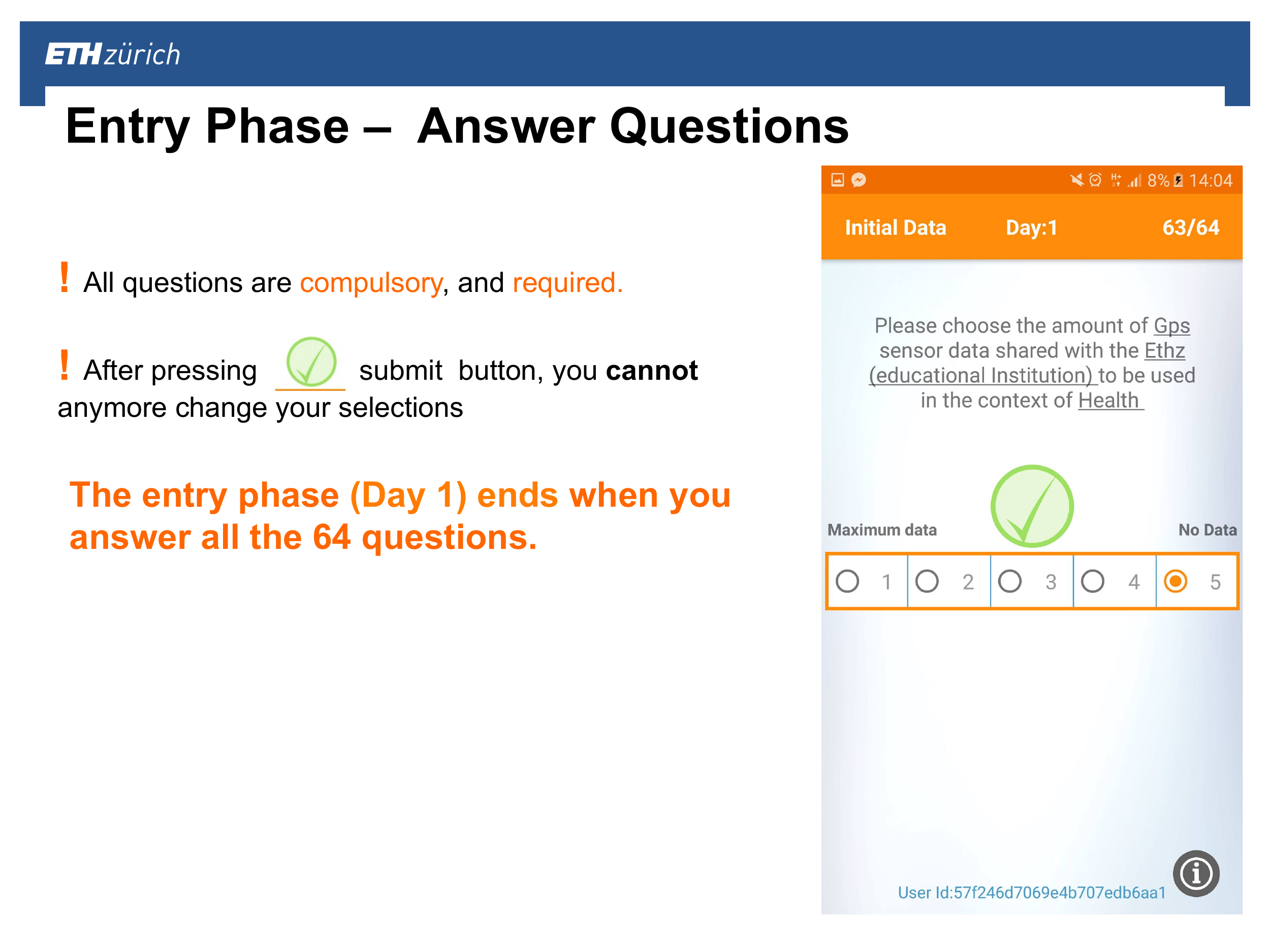}
	\includegraphics[width=0.31\textwidth]{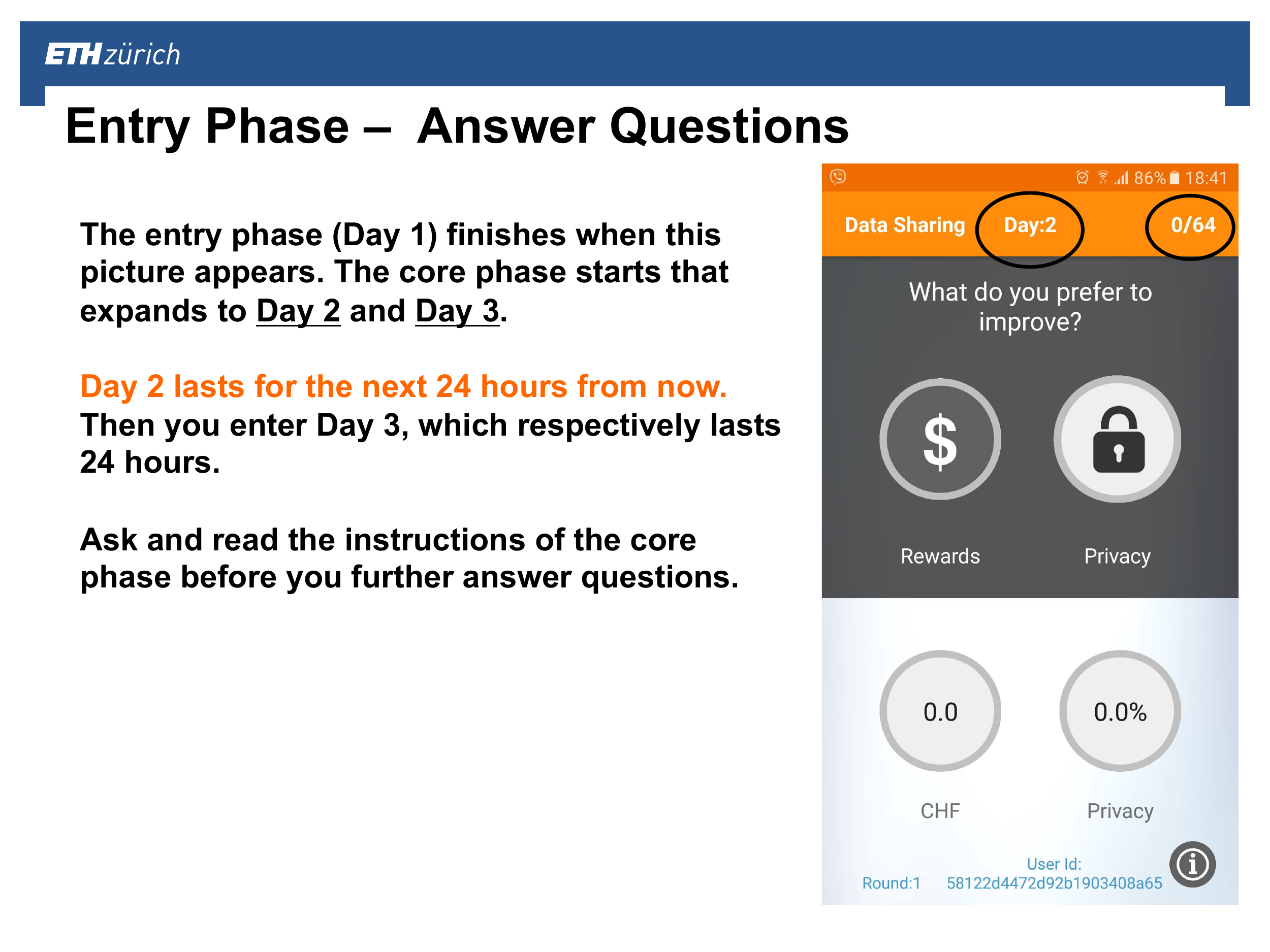}
	\caption{Instructions presented to the participants starting with the entry phase.}\label{fig:instructions-entry-phase}
\end{figure}

\begin{figure}[!htb]
	\centering
	\includegraphics[width=0.41\textwidth]{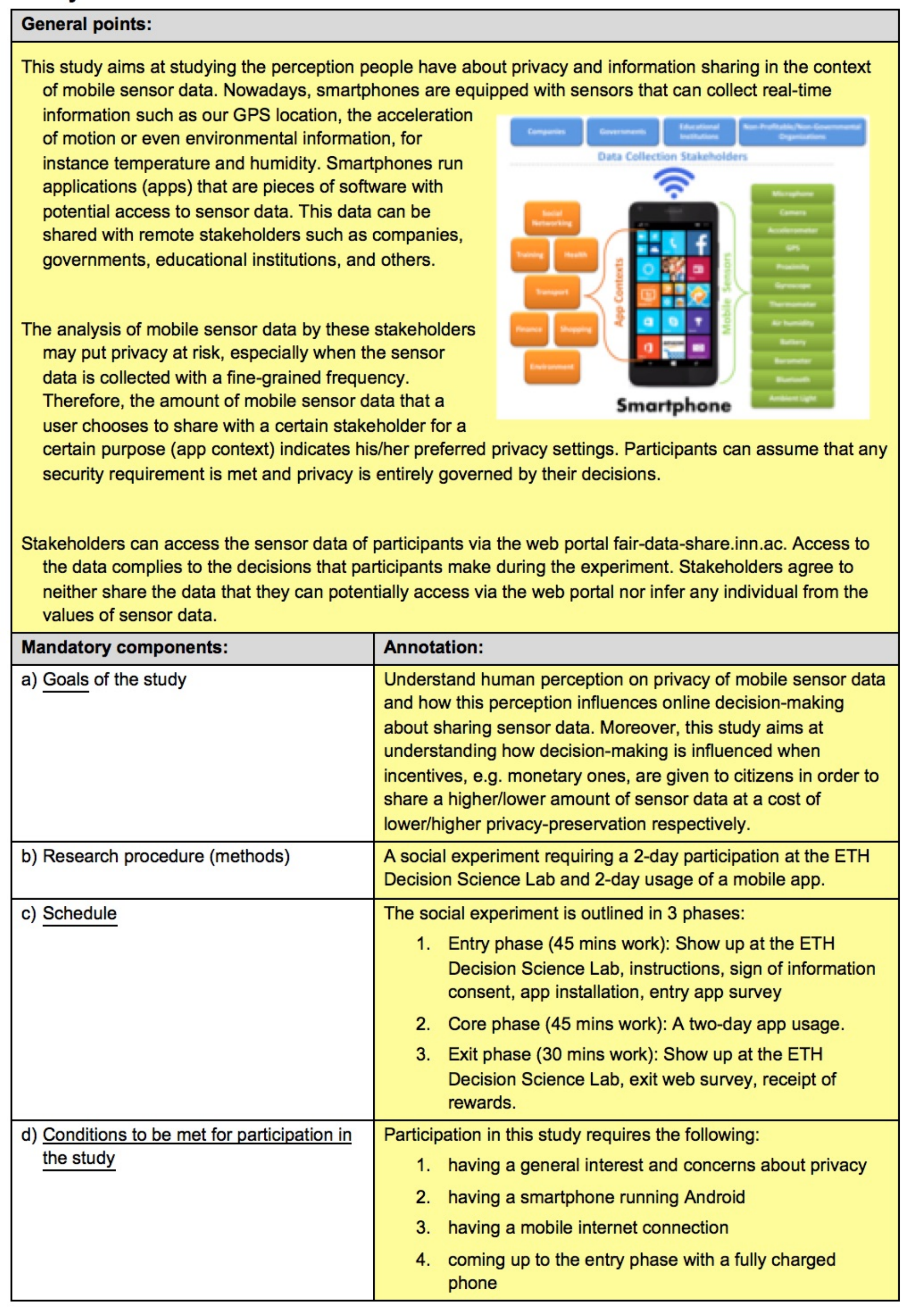}
	\includegraphics[width=0.48\textwidth]{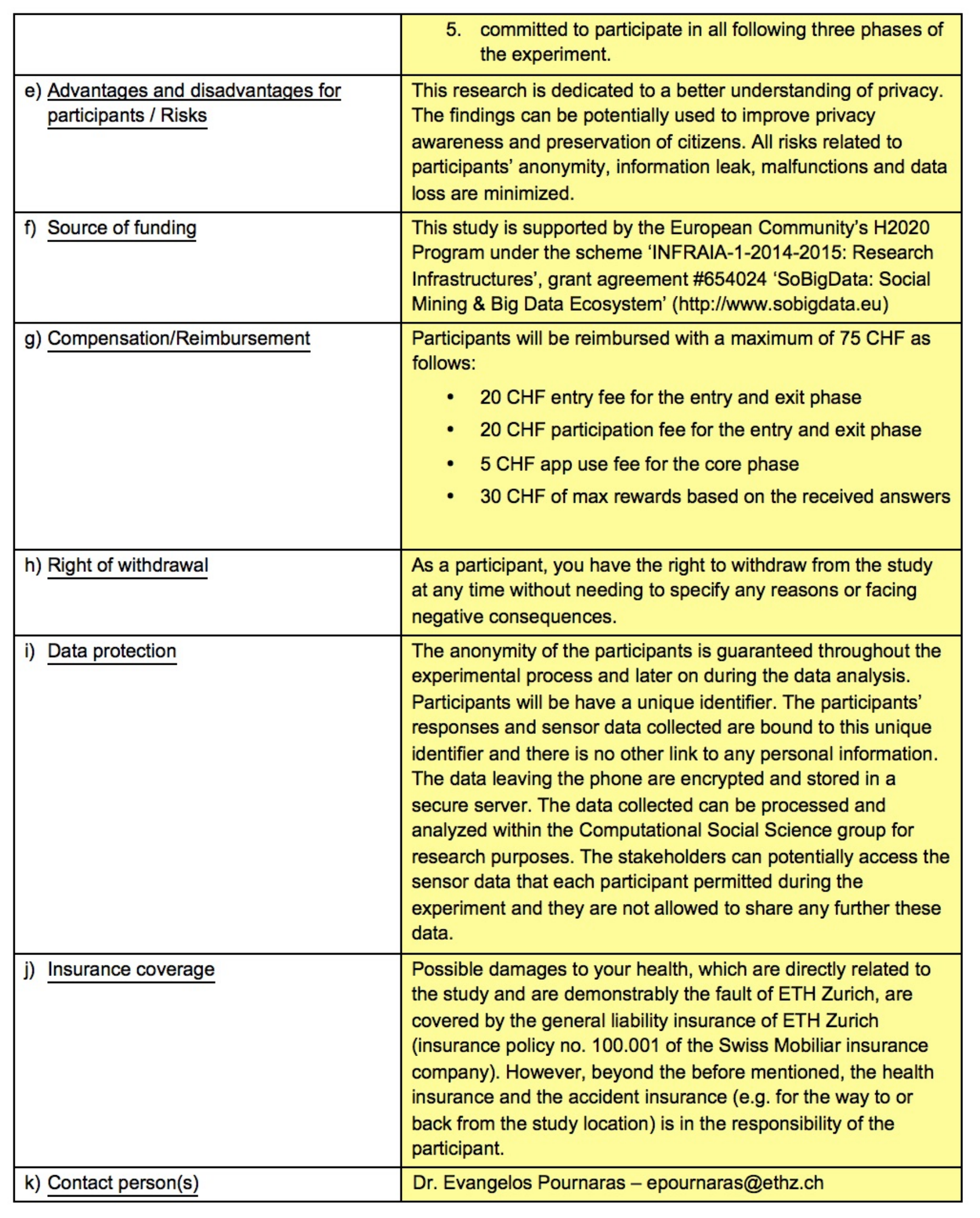}
	\includegraphics[width=0.43\textwidth]{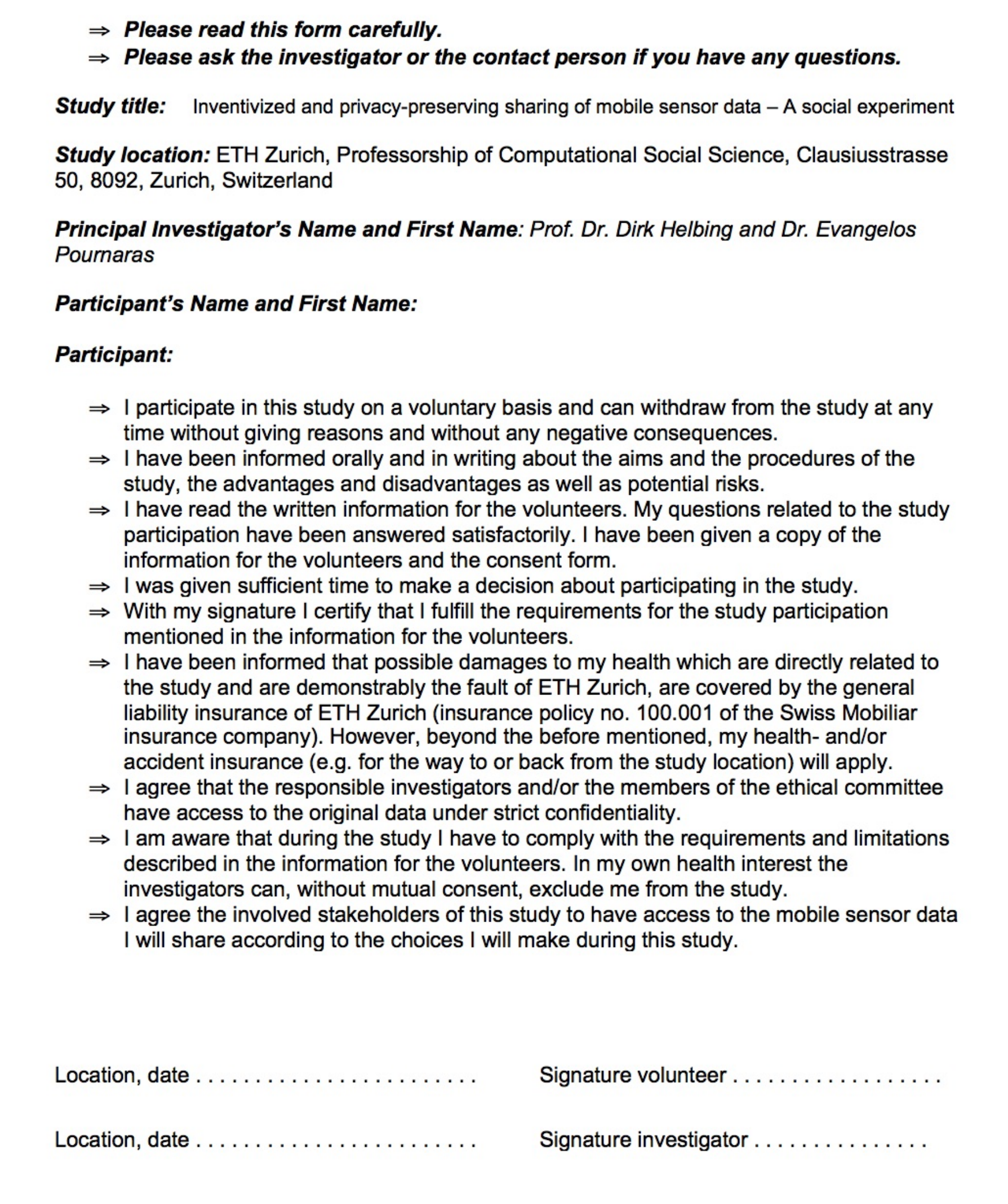}
	\caption{Information consent for participation in the designed experiment.}\label{fig:information-consent}
\end{figure}

The participants of each experimental session are verified by the DeSciL staff members by presenting a personal identification document, i.e. a passport or student card, nevertheless, the actual identity of the participants remains anonymous to the researchers using the lab. Participants are not allowed to interact with each other during the experiment and any questions need to be addressed in private directly to the experiment moderators by moving to a next room. In this way, biases about how each participant perceives and understands the experimental process are eliminated. This process is communicated to the participants before the beginning of the experiment. Next, participants are seated in a room with instructions about the experiment (Figure~\ref{fig:instructions-entry-phase}) and the information consent (Figure~\ref{fig:information-consent}) placed in front of them. 

\begin{figure}[!htb]
	\centering
	\subfigure[The Android app at Google Play]{\includegraphics[width=0.35\textwidth]{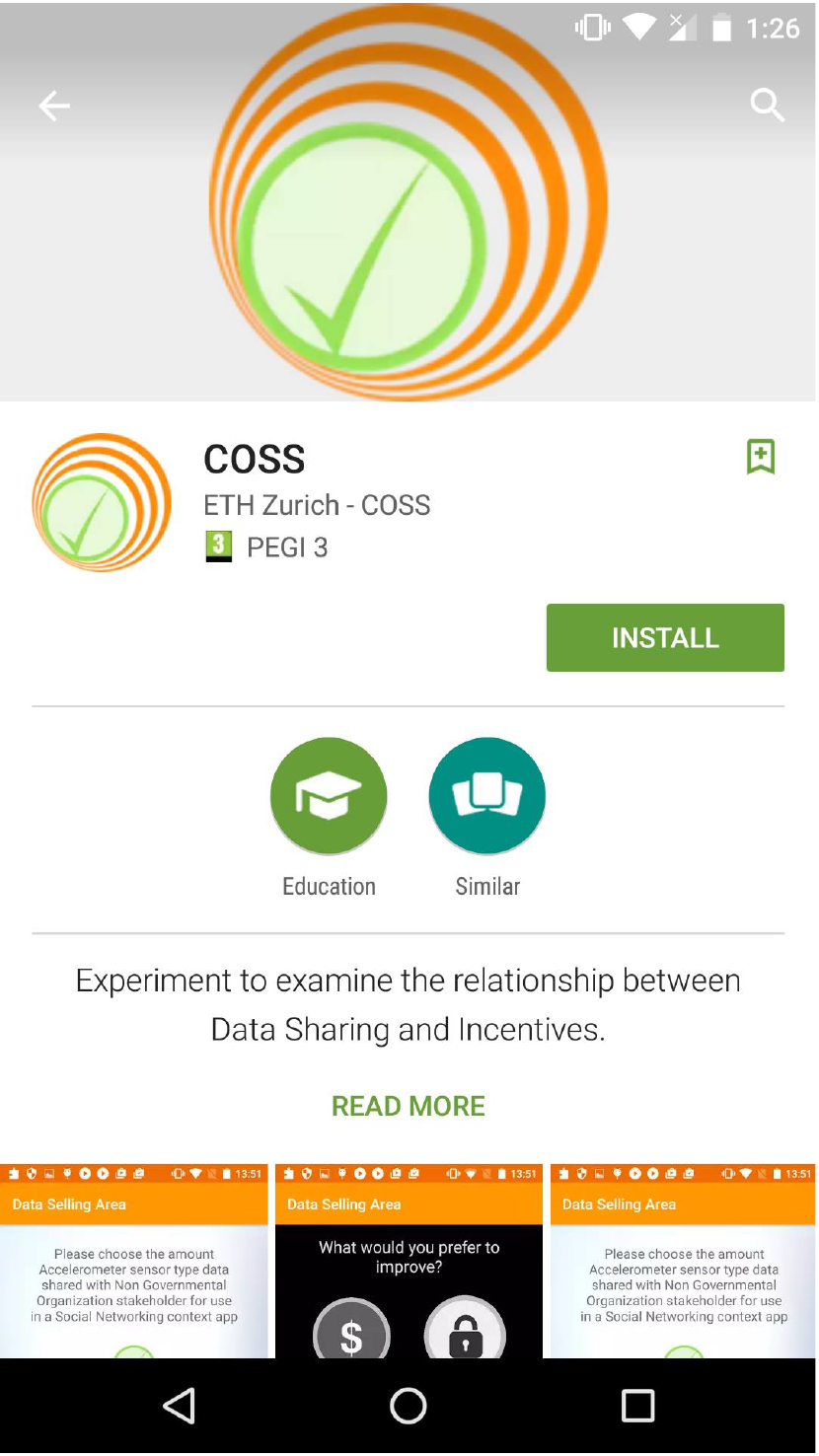}}
	\subfigure[An instantiation of the factorial question]{\includegraphics[width=0.35\textwidth]{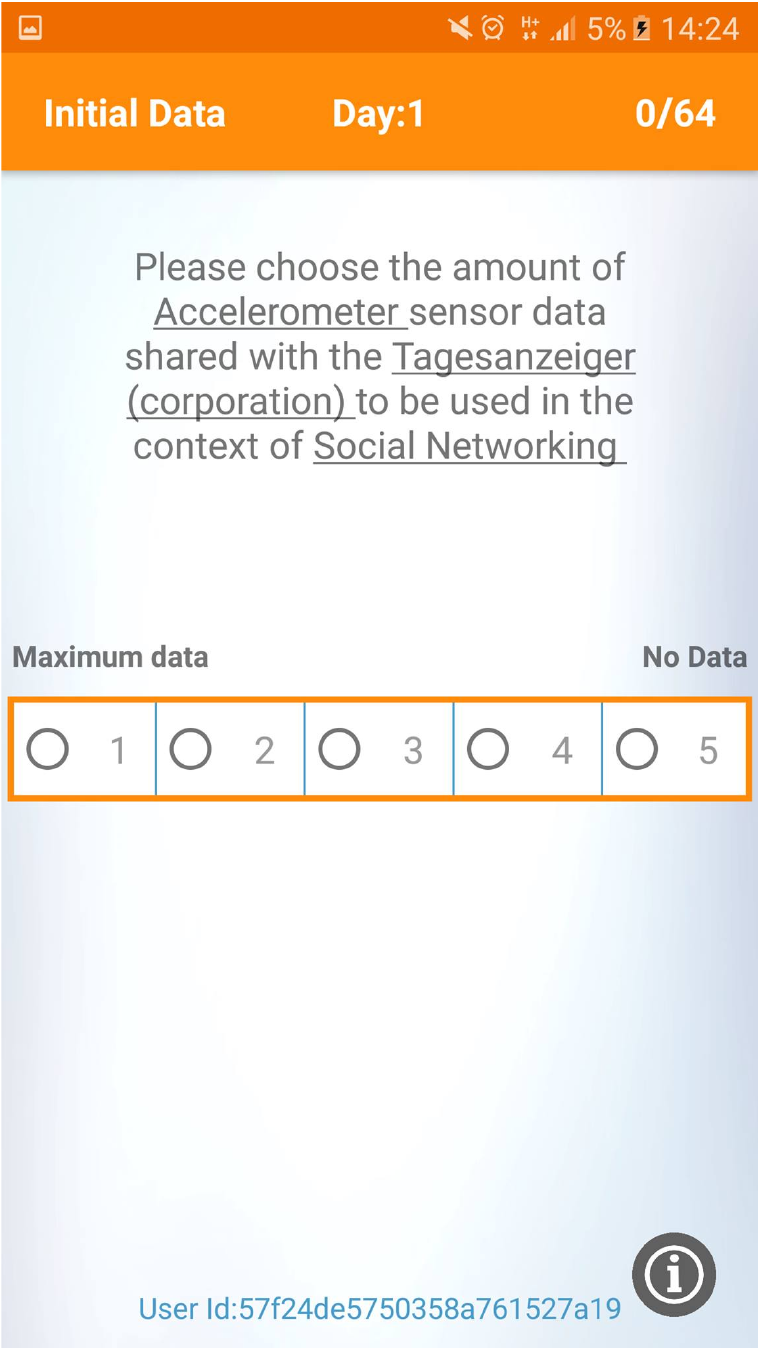}}
	\caption{Screens of the Android app during the entry phase.}\label{fig:app-entry}
\end{figure}

The Android app was made available in Google Play online store for the participants to download, see Figure~\ref{fig:app-entry}a. The app generates locally in the background a unique ID used as identifier of the participants in the experiment as well as in the data collected in the database. This ID can be viewed in the app by participants. The first screens of the app present the survey questions B.1-B.8 of Table~\ref{table:entry-phase}. 


\begin{table}[!htb]
	\caption{Survey questions for the entry phase.}\label{table:entry-phase}
	\centering
	\resizebox{\textwidth}{!}{%
		\begin{tabular}{p{0.15\textwidth}p{0.65\textwidth}p{0.19\textwidth}p{0.39\textwidth}}
			\toprule
			\textbf{ID} & \textbf{Question} & \textbf{Type}& \textbf{Options} \\
			\midrule
			B.1 & Gender & multiple choice, one selection & male, female  \\\midrule
			B.2 & Year of birth & multiple choice, one selection & 81 [1920,2000]  \\\midrule
			B.3 & Education level & multiple choice, one selection & less than high school, high school, some college, bachelors degree, masters degree, PhD degree  \\\midrule
		    B.4 & How concerned are you about the privacy of your mobile sensor data? & ratio scale & 5 [not at all,highly]\\\midrule
			B.5 & Employment status & multiple choice, one selection & full time, part time, not looking for work, looking for work, retired, student, disabled  \\\midrule
			B.6 & In which country did you spend most of your life? & multiple choice, one selection & all countries  \\\midrule
			B.7 & How often do you check your mobile phone a day? & multiple choice, one selection & <35, 36-70, 71-100, 101-135, >135\\\midrule
			B.8 & Which types of apps do you usually have on your smartphone? & multiple choice, multiple selections & education, entertainment, finance, game, health \& fitness, transportation \& navigation, music \& audio, news, productivity, shopping, social networking, medical, traveling, utilities, weather\\\midrule
			B.9 & How intrusive are the following features of information sharing? & group of questions & 3 questions\\
			\qquad B.9.1 & \qquad Sensors & multiple choice, single selection & very low, low, medium, high, very high\\
			\qquad B.9.2 & \qquad Data collectors & multiple choice, single selection & very low, low, medium, high, very high\\
			\qquad B.9.3 & \qquad Context/Purpose & multiple choice, single selection & very low, low, medium, high, very high\\\midrule
			B.10 & How privacy intrusive is the data sharing of the following sensors? & group of questions & 4 questions\\
			\qquad B.10.1 & \qquad Accelerometer & multiple choice, single selection & very low, low, medium, high, very high\\
			\qquad B.10.2 & \qquad Location & multiple choice, single selection & very low, low, medium, high, very high\\
			\qquad B.10.3 & \qquad Light & multiple choice, single selection & very low, low, medium, high, very high\\
			\qquad B.10.4 & \qquad Noise & multiple choice, single selection & very low, low, medium, high, very high\\\midrule
			B.11 & How privacy intrusive are the following data collectors of your mobile sensor data? & group of questions & 4 questions\\
			\qquad B.11.1 & \qquad Corporations & multiple choice, single selection & very low, low, medium, high, very high\\
			\qquad B.11.2 & \qquad Non-governmental Organizations & multiple choice, single selection & very low, low, medium, high, very high\\
			 \qquad B.11.3 & \qquad Governments & multiple choice, single selection & very low, low, medium, high, very high\\
			\qquad B.11.4 & \qquad Educational Institutes & multiple choice, single selection & very low, low, medium, high, very high\\\midrule
			B.12 & How privacy intrusive are the following contexts under which sensor data is used by stakeholders? & group of questions & 4 questions\\
			\qquad B.12.1 & \qquad Health/Fitness & multiple choice, single selection & very low, low, medium, high, very high\\
			\qquad B.12.2 & \qquad Social Networking & multiple choice, single selection & very low, low, medium, high, very high\\
			\qquad B.12.3 & \qquad Environment & multiple choice, single selection & very low, low, medium, high, very high\\
			\qquad B.12.4 & \qquad Transportation & multiple choice, single selection & very low, low, medium, high, very high\\
			\bottomrule
		\end{tabular}
	}
\end{table}

The next screens personalize the sharing of sensor data. Initially, the three criteria of (i) sensor type, (ii) data collector and (iii) context receive their weights according to the perception of each participant on how privacy intrusive they are. The answers of the group Question B.9 in the range `very low' to `very high' are mapped to the weights $\criterionWeight{\participant}{\criterion}$ of the three criteria as illustrated in Section~\ref{subsec:valuation}. Moving to the next screens, the same personalization process is repeated within each criterion: for the different sensor types (group Question~B.10), data collectors (group Question~B.11) and contexts (group Question~B.12).

Table~\ref{table:factorial-design} illustrates the three criteria and its elements during the experiment. For each feature, four possible elements are selected in the factorial experiment to keep a manageable number of $4*4*4=64$ total combinations. 

\begin{table}[!htb]
	\caption{The selected elements in the criteria for sharing mobile sensor data.}\label{table:factorial-design}
	\centering
\resizebox{\textwidth}{!}{%
		\begin{tabular}{lll}
			\toprule
			 \textbf{Sensor Type} & \textbf{Data Collector} & \textbf{Context}\\
			\midrule
			GPS & Corporation (Tagesanzeiger) & Social networking  \\
		    Microphone & Non-profit, non-governmental organizations (Swiss Made Software) & Environment  \\
		   Accelerometer & Educational institutes (ETH Zurich) & Transportation/traveling  \\
		   Light & Governmental organizations (The State Secretariat for Education, Research and Innovation) & Health/fitness \\
			\bottomrule
		\end{tabular}
	}
\end{table}

\begin{figure}[!htb]
	\centering
	\subfigure[Sensor types]{\includegraphics[width=0.3\textwidth]{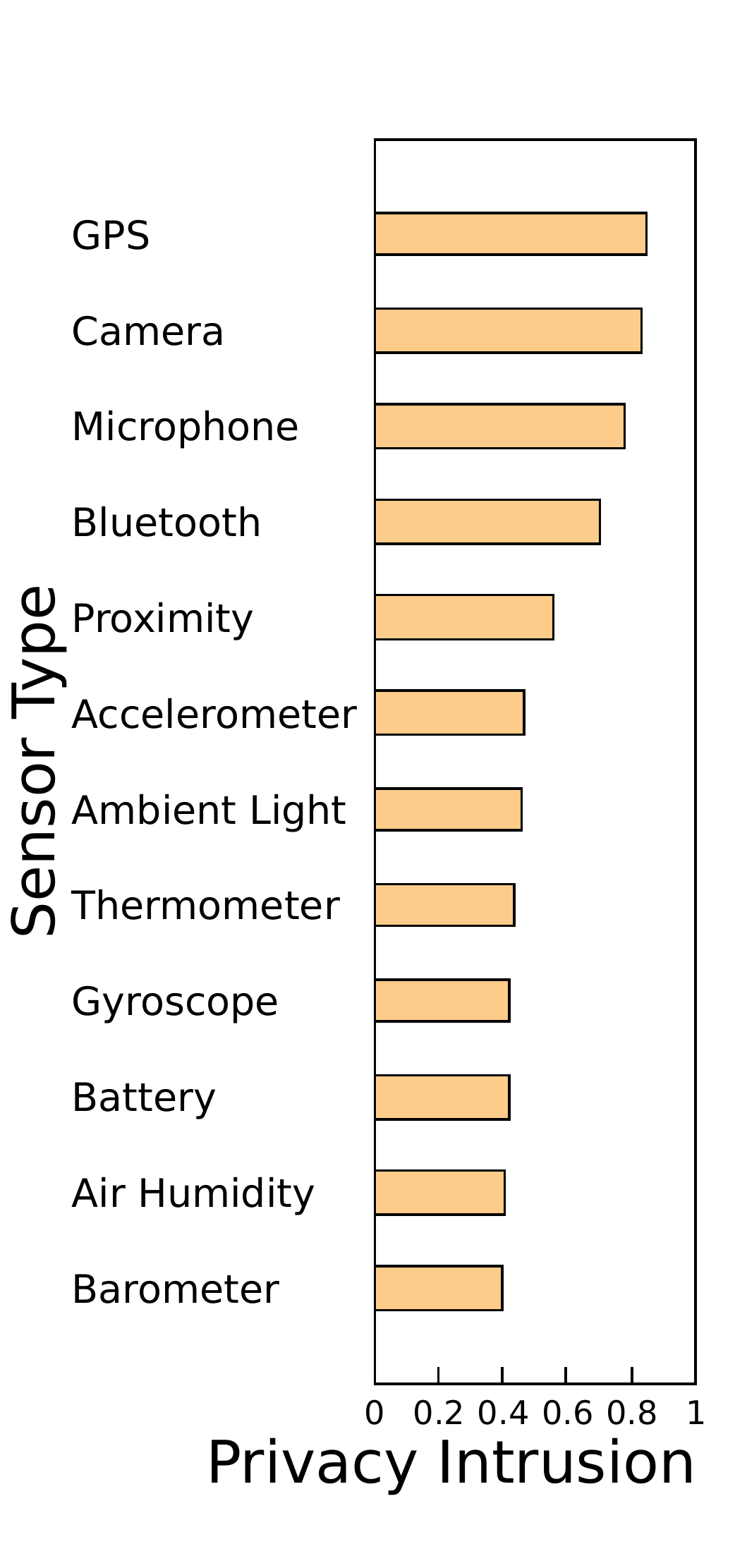}}
	\subfigure[Contexts]{\includegraphics[width=0.3\textwidth]{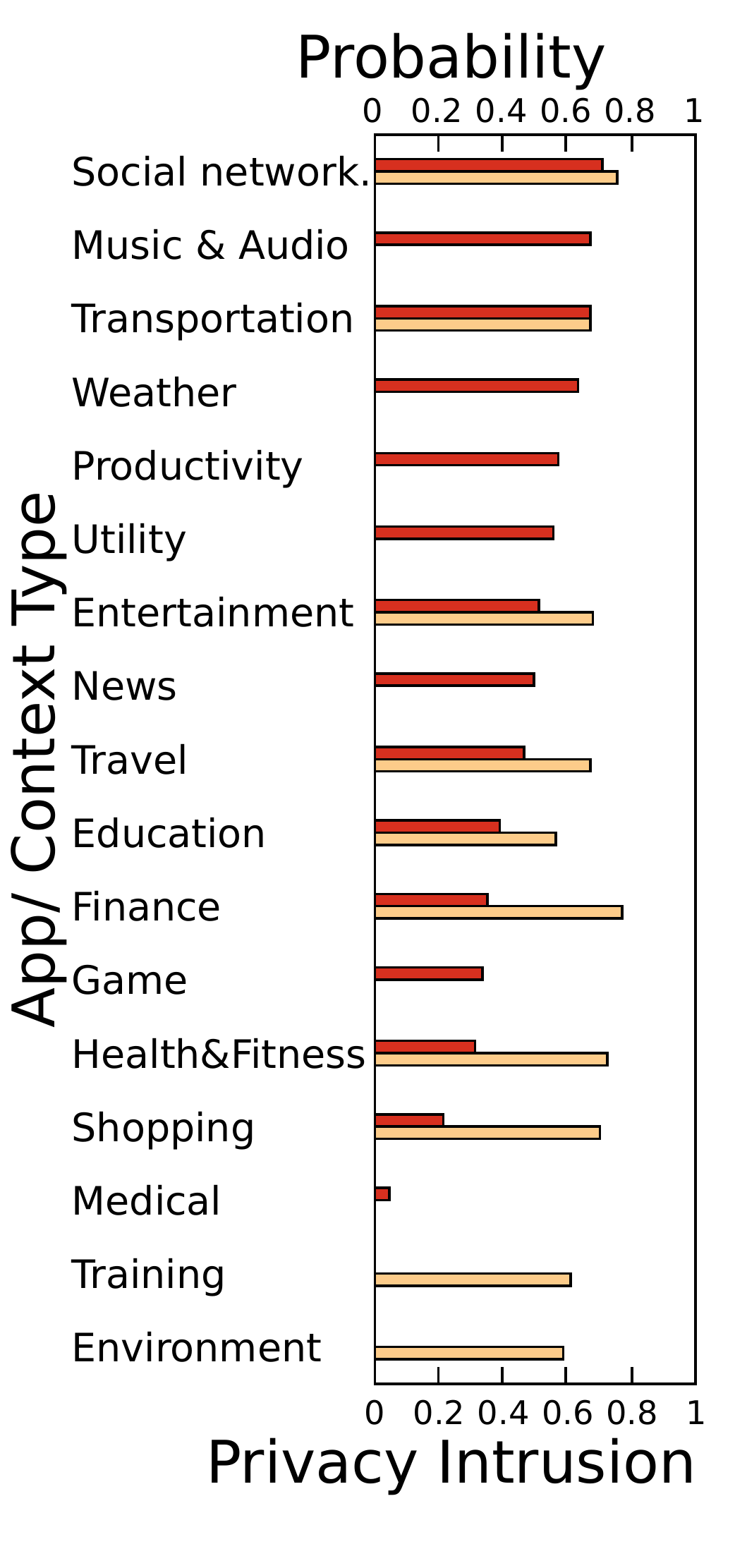}}
	\caption{Data used from Question~A.9,~A.13 (yellow bars, intrusion) and~A.6 (red bars, probability) of the preparatory phase for choosing the elements of the data-sharing criteria for the factorial experiment.}\label{fig:variables-selection}
\end{figure}

The elements of the data-sharing criteria are chosen after scrutinizing the responses received by the participants of the preparatory phase in Question~A.9 for the sensor types (Figure~\ref{fig:variables-selection}a) as well as Question~A.13 and~A.6 for the contexts (Figure~\ref{fig:variables-selection}b and~\ref{fig:variables-selection}c respectively). Two out of the top-3 highly privacy-intrusive sensors are selected. These are the GPS (privacy intrusion of 0.85) and microphone (privacy intrusion of 0.78). The camera sensor is ranked 2nd with privacy intrusion of 0.83. It is not selected as it requires the collection of more complex data and higher storage space in the smartphones. The accelerometer (ranked 6th with privacy intrusion of 0.47) and light (ranked 7th with privacy intrusion of 0.46) sensors are the other two ones selected that belong into the middle ranking range of privacy intrusion. 

Figure~\ref{fig:app-entry}b illustrates an instantiation example of the factorial question. After answering all questions, participants complete their participation in the entry phase and the smartphone app initializes the core phase. They receive the instructions of the core phase and they depart from DeSciL. Note that the answers to the instantiations of the factorial question during the entry phase are not monetary rewarded. The answers to these questions are the baseline with which the rewarded sharing of mobile sensor data during the core phase is compared. 

\subsection{Core phase}\label{subsubsec:core-phase}

\begin{figure}[!htb]
	\centering
	\includegraphics[width=0.31\textwidth]{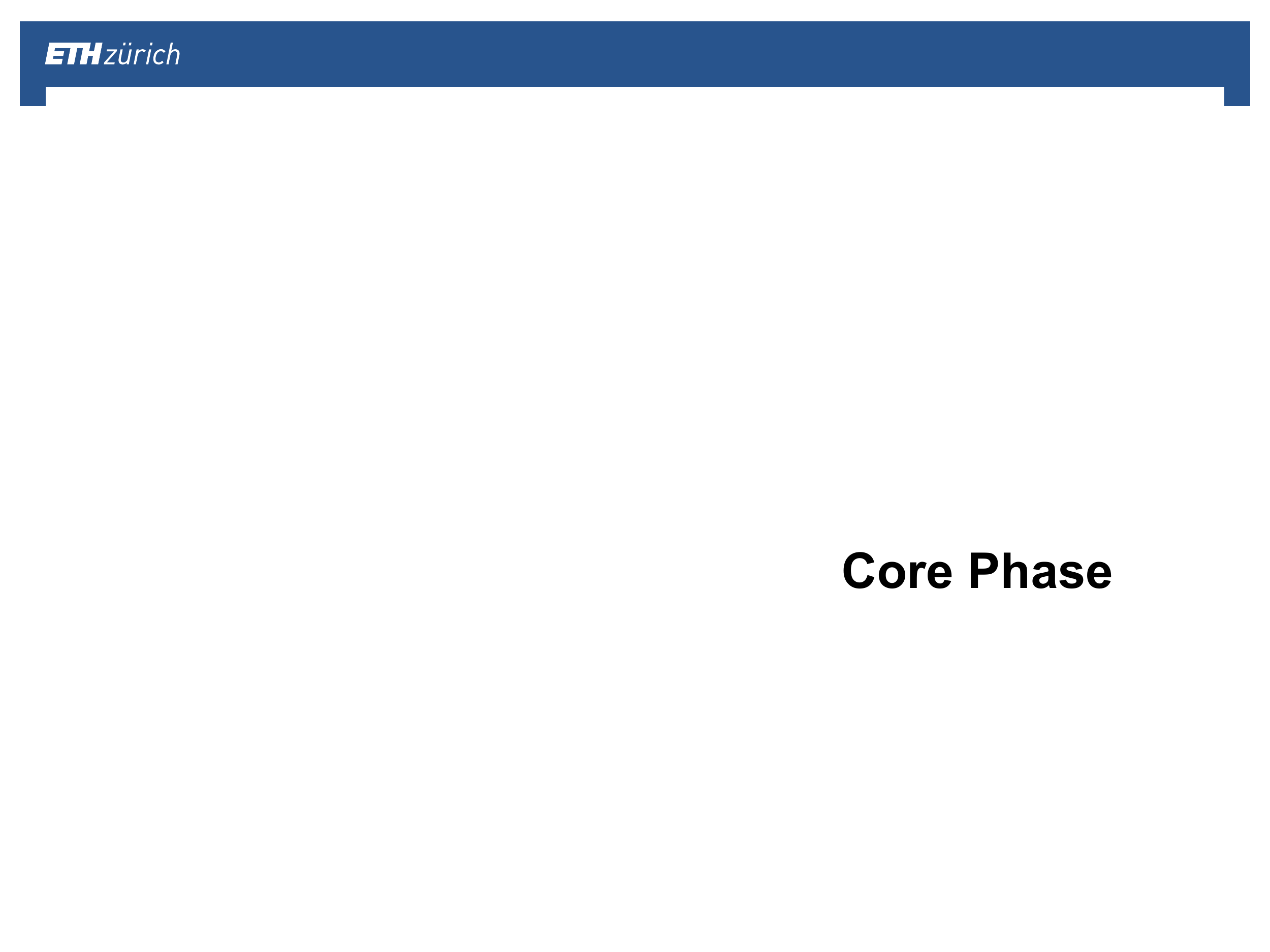}
	\includegraphics[width=0.31\textwidth]{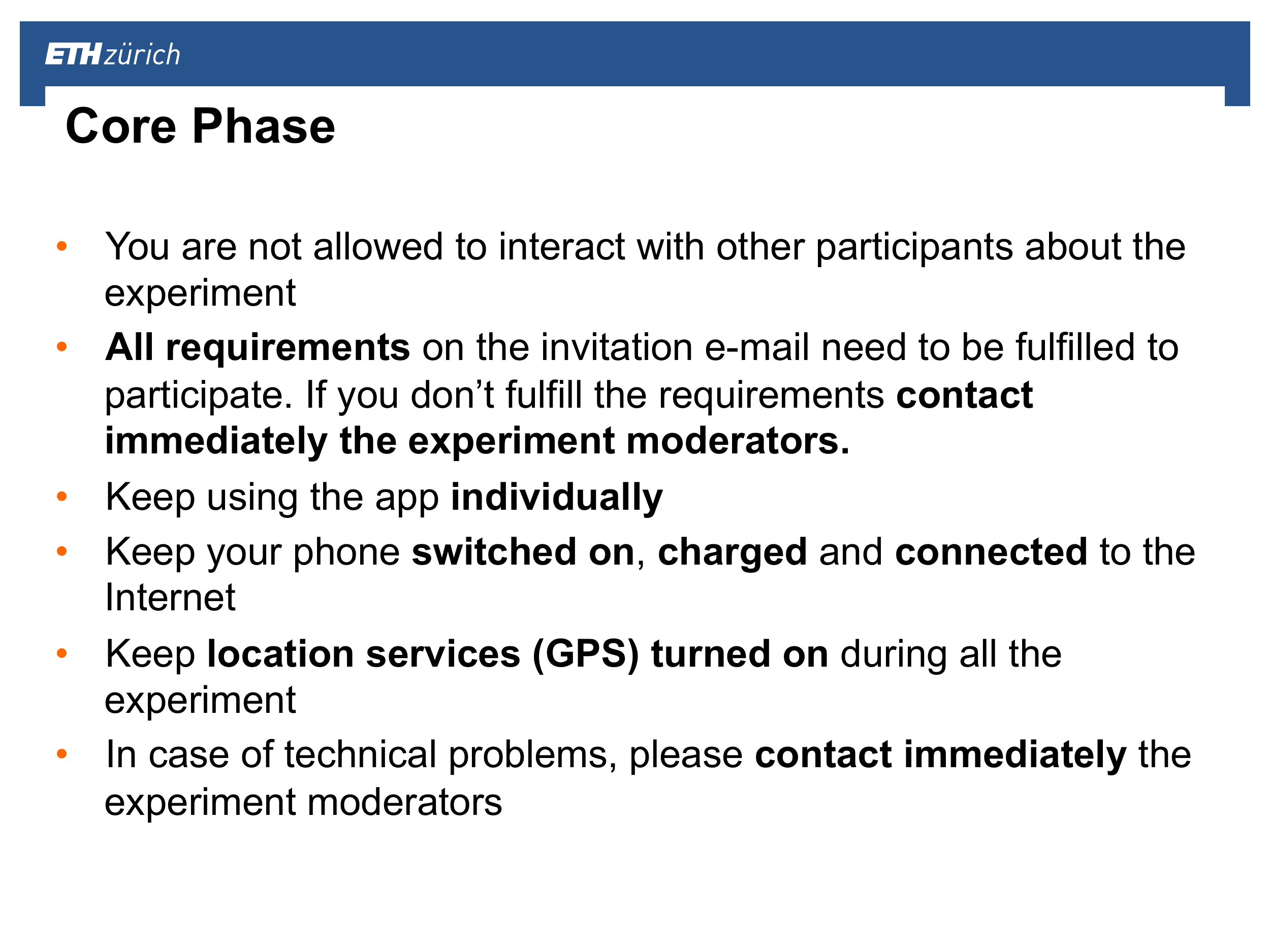}
	\includegraphics[width=0.31\textwidth]{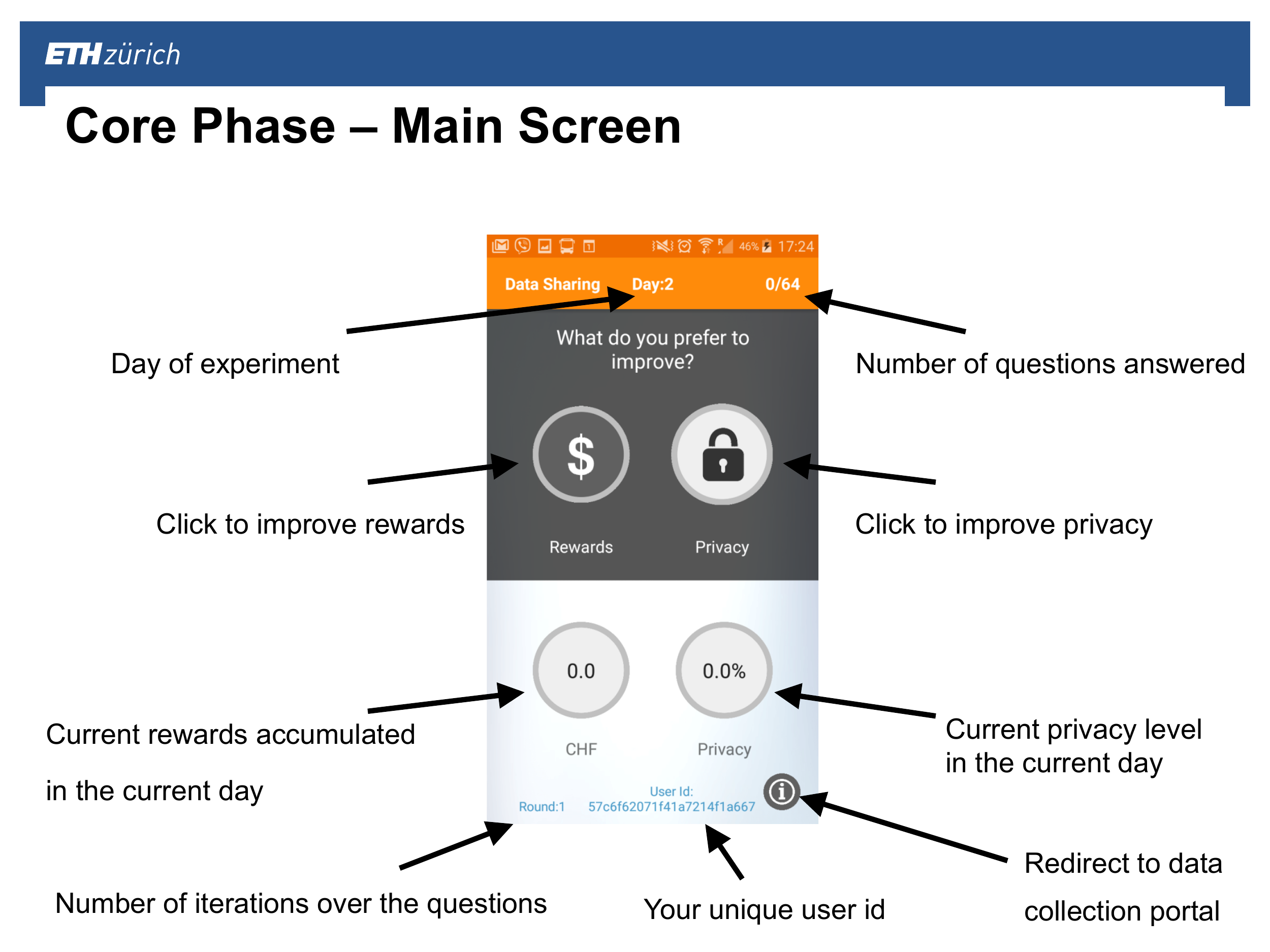}
	\includegraphics[width=0.31\textwidth]{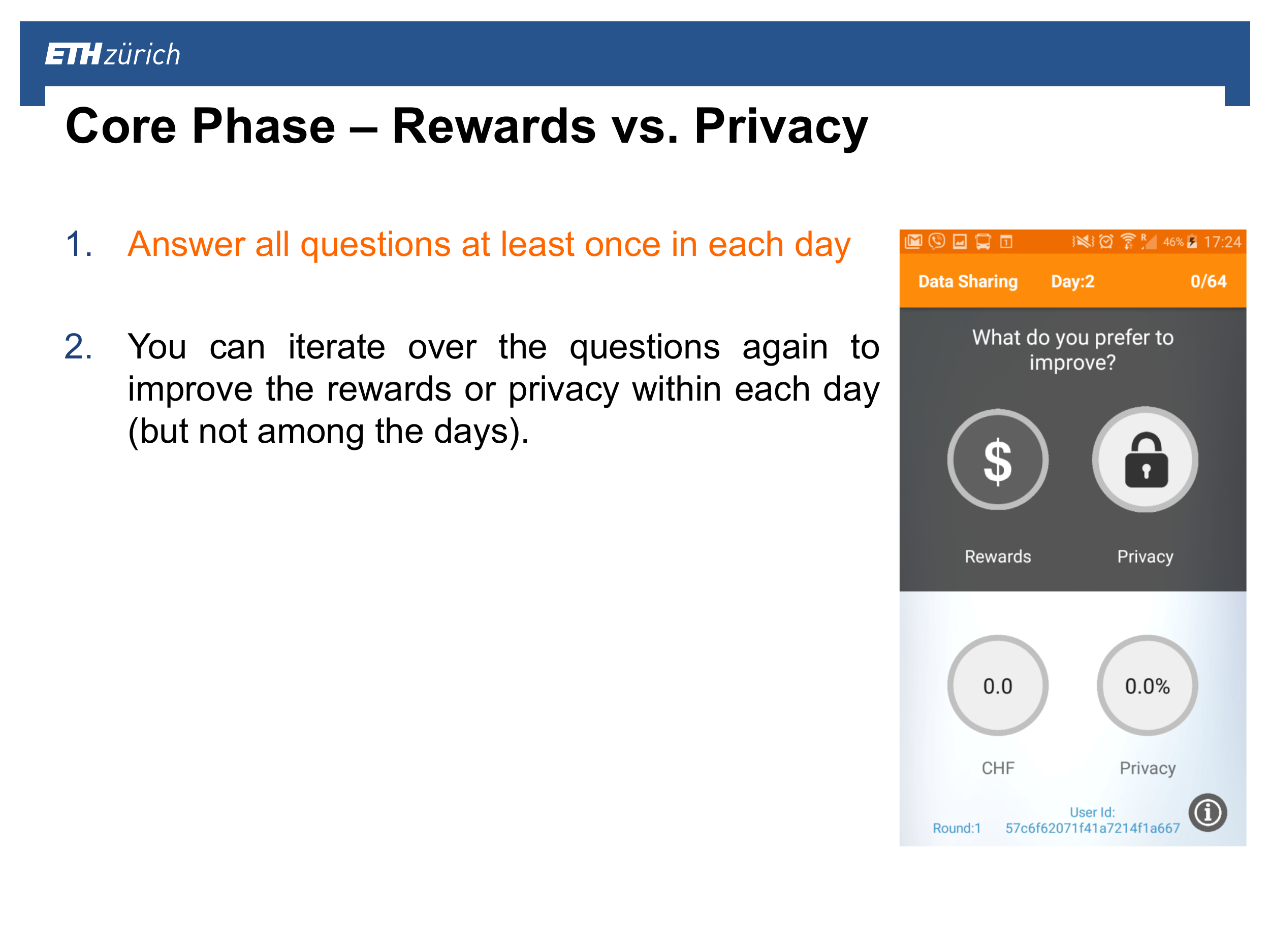}
	\includegraphics[width=0.31\textwidth]{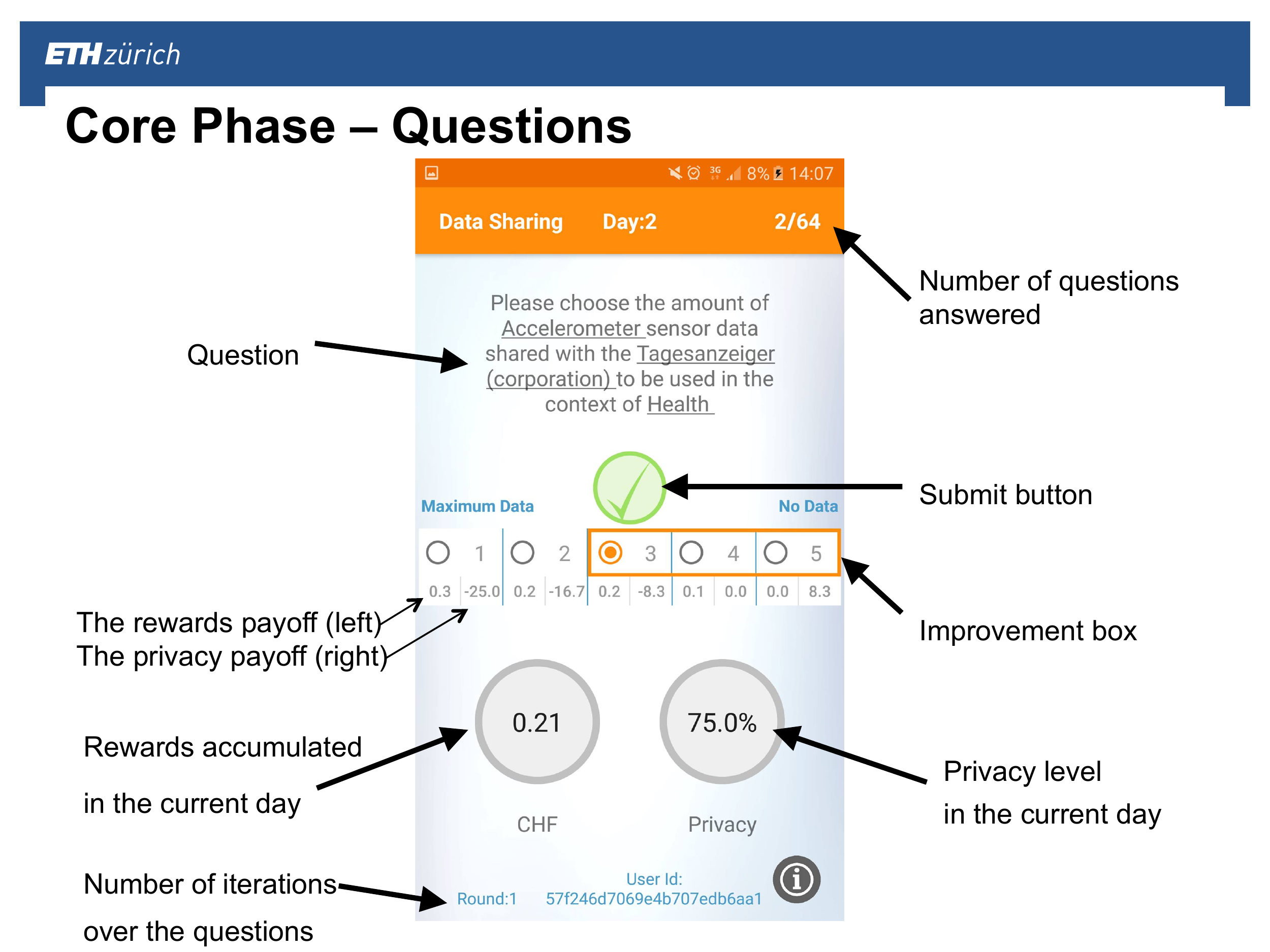}
	\includegraphics[width=0.31\textwidth]{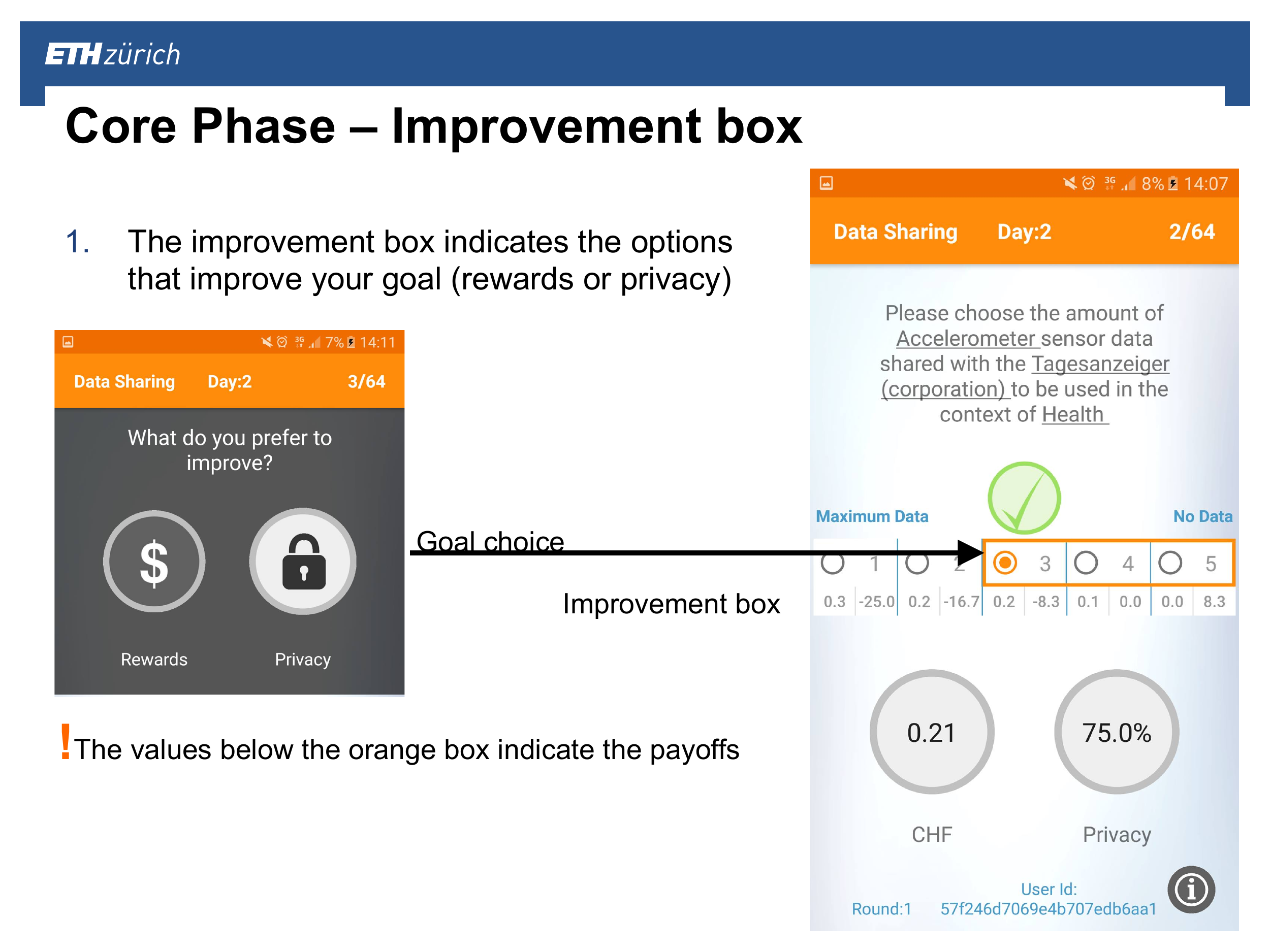}
	\caption{Instructions presented to the participants starting the core phase and after finishing the entry phase.}\label{fig:instructions-core-phase}
\end{figure}

\begin{figure}[!htb]
	\centering
	\includegraphics[width=0.31\textwidth]{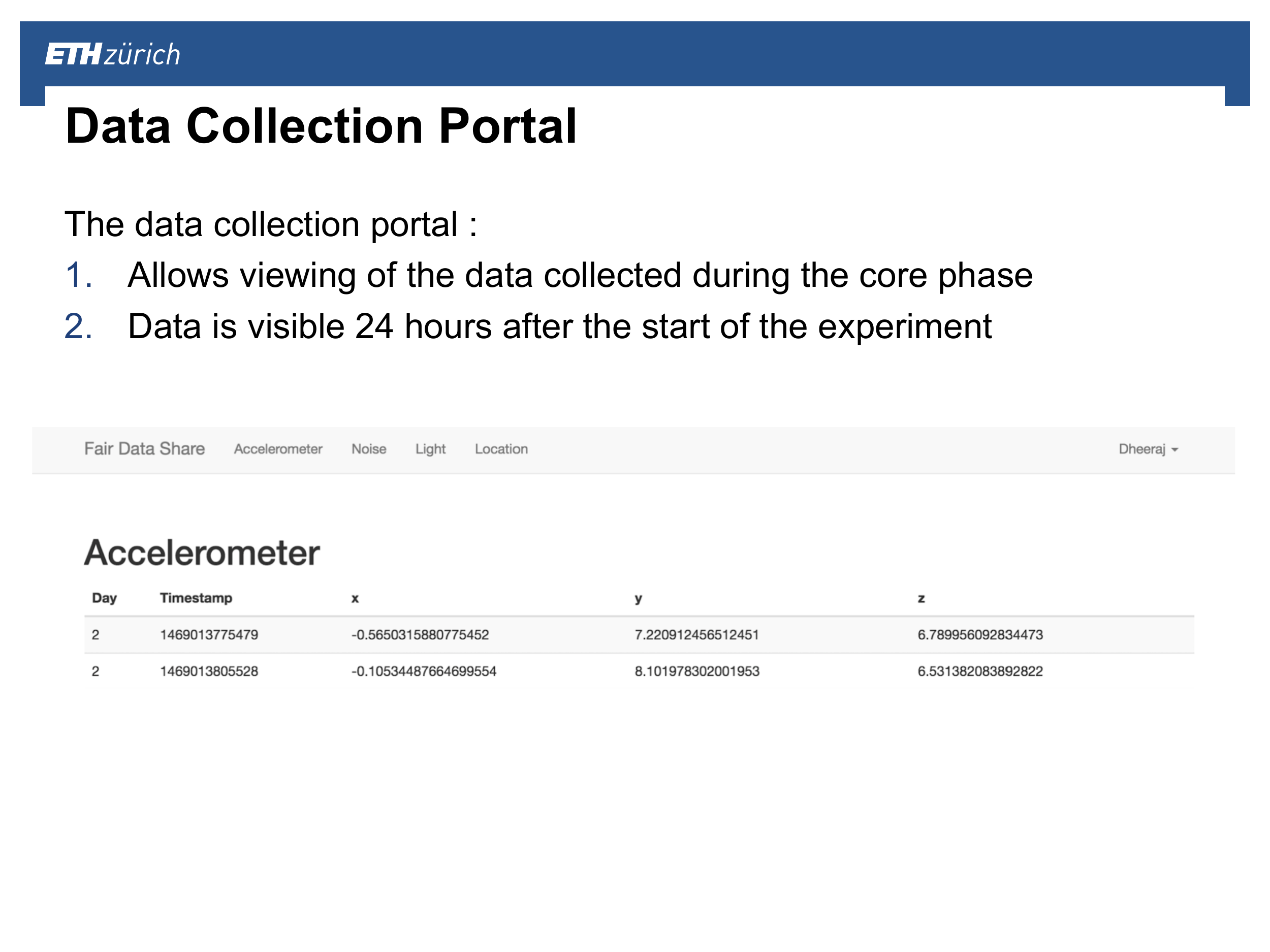}
	\includegraphics[width=0.31\textwidth]{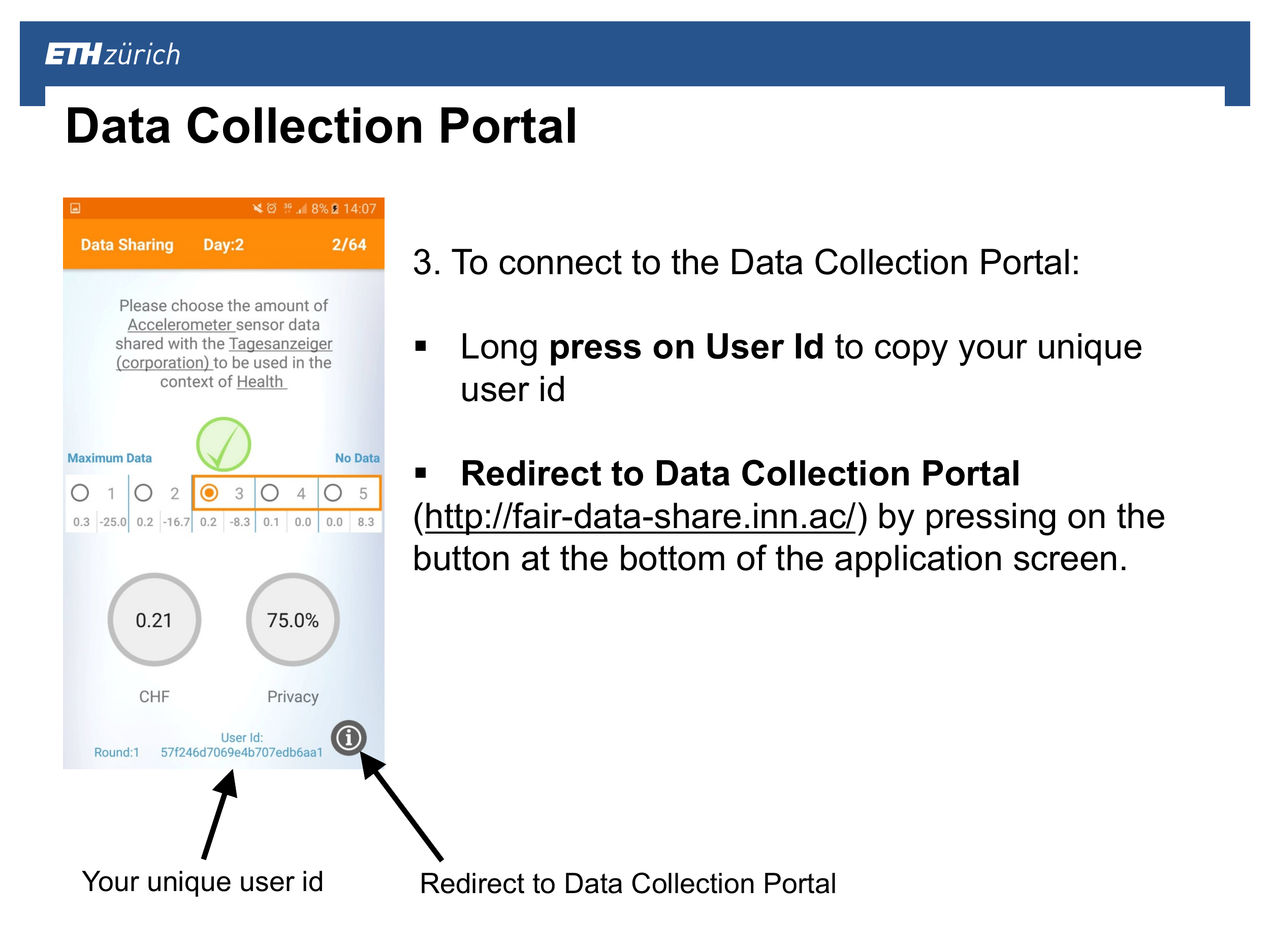}
	\includegraphics[width=0.31\textwidth]{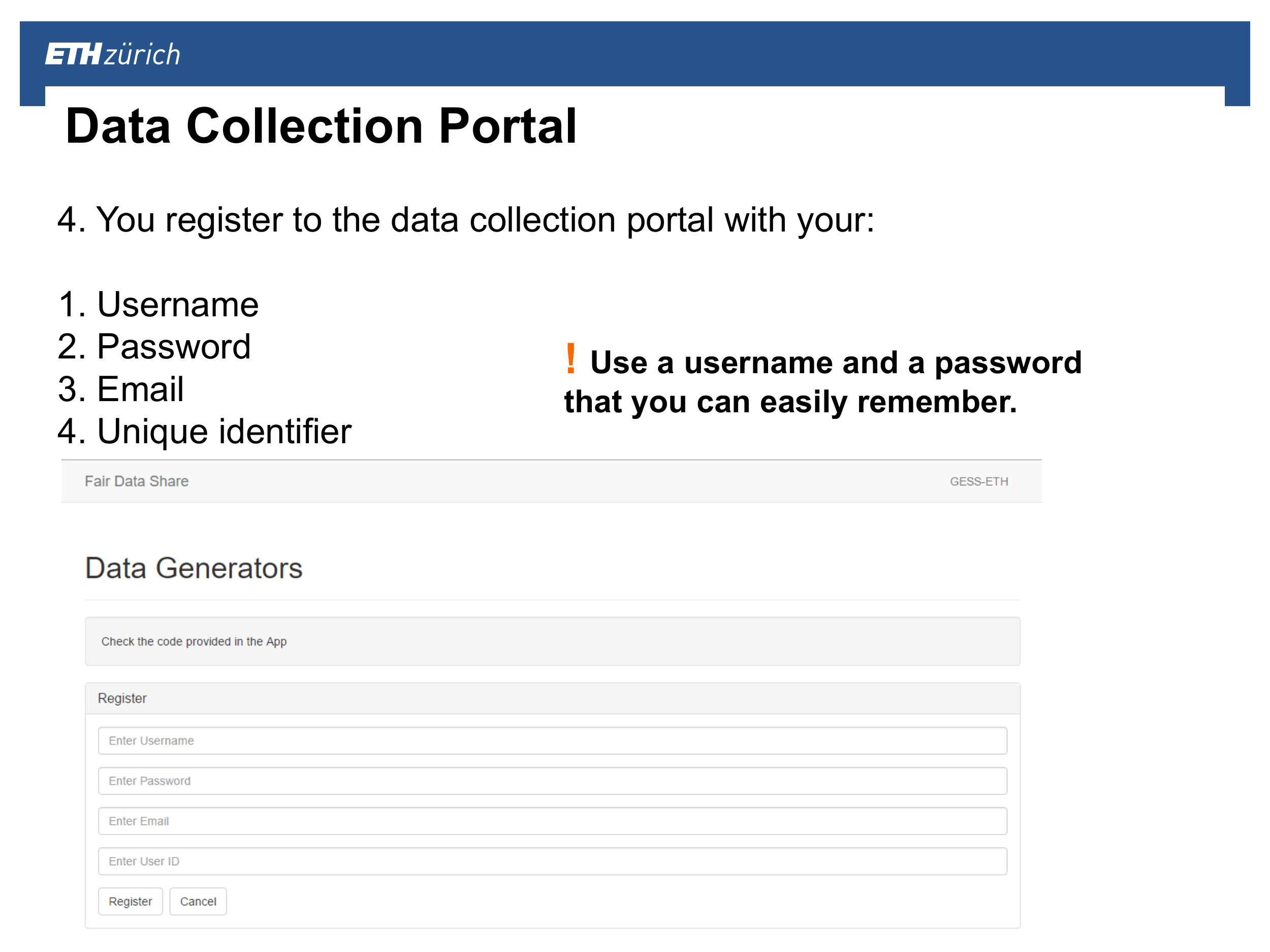}
	\includegraphics[width=0.31\textwidth]{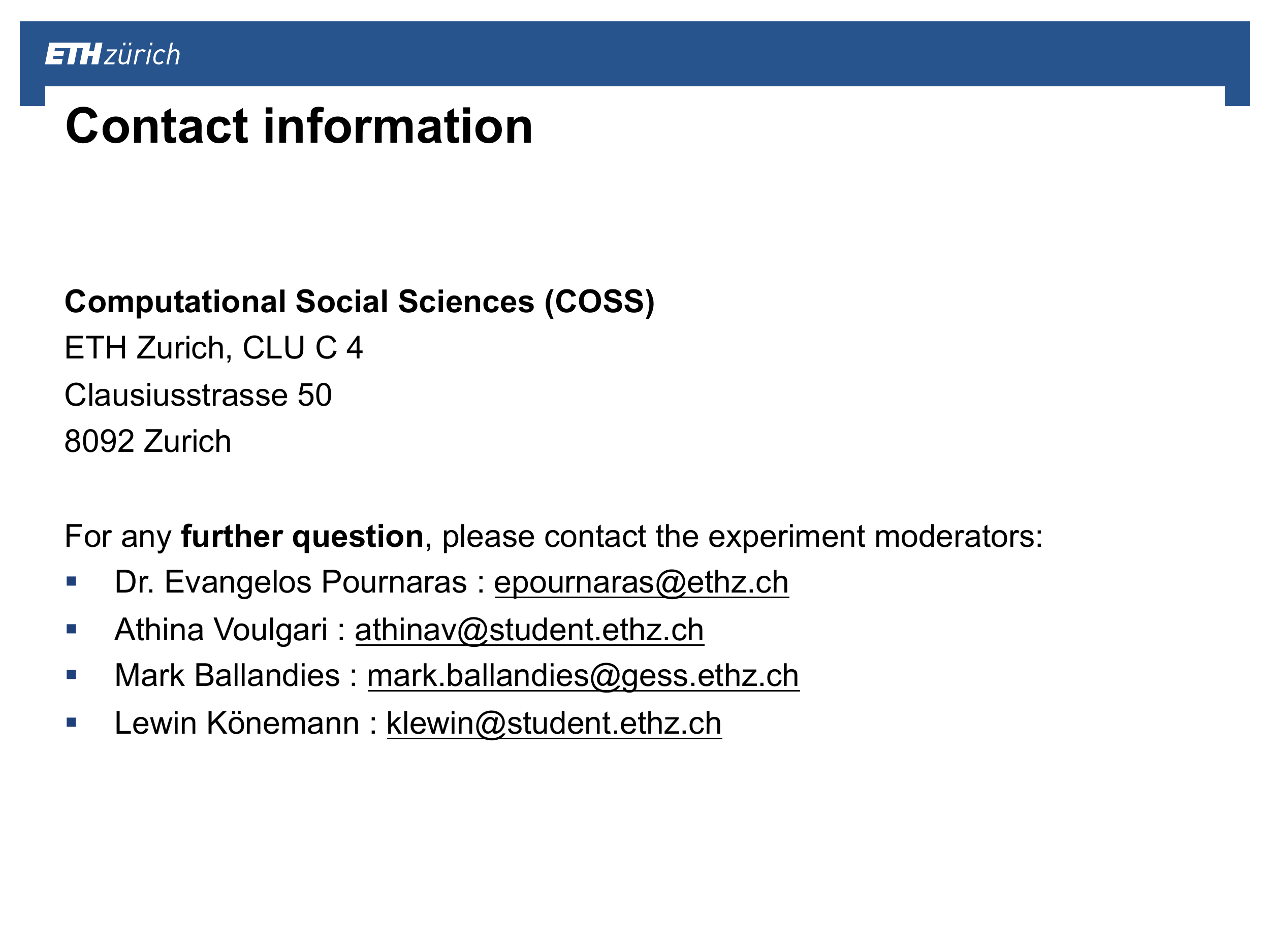}
	\caption{Instructions on the data-access portal presented to the participants starting the core phase and after finishing the entry phase.}\label{fig:instructions-portal}
\end{figure}

The core phase is initialized right after the completion of the entry phase when participants also receive the instructions shown in Figure~\ref{fig:instructions-core-phase}. They also receive at this phase the instructions about the data-access portal, see Figure~\ref{fig:instructions-portal}. The core phase lasts for two full days (48 hours, Mondays to Tuesdays and Tuesdays to Wednesdays as shown in Table~\ref{tab:sessions}.). It takes place out of DeSciL lab and integrates to the daily life of participants. At the beginning of each day in the core phase, the rewards are zero as no data sharing is performed unless the participants consent to this via their responses to the data-sharing scenarios. 

\subsection{Exit phase}\label{subsubsec:exit-phase}

The exit phase is performed on Thursdays, the 4th day of each experimental session (see Table~\ref{tab:sessions}), and involves the return of the participants to DeSciL. The staff members of the lab verify the identify of the participants and they are then seated to lab computers to fill in an online survey created in Qualtrics. The questions of the exit survey are outlined in Tables~\ref{table:exit-phase-general} to~\ref{table:exit-phase-experiment}. The matching of the data collected in this phase with the data of the previous phases is performed with the user ID inserted in Question~D.1. 

\begin{table}[!htb]
	\caption{General survey questions for the exit phase.}\label{table:exit-phase-general}
	\centering
	\resizebox{\textwidth}{!}{%
		\begin{tabular}{p{0.1\textwidth}p{0.9\textwidth}p{0.19\textwidth}p{0.1\textwidth}}
			\toprule
			\textbf{ID} & \textbf{Question} & \textbf{Type}& \textbf{Options} \\
			\midrule
			D.1 & Please enter your unique ID number & open-ended & -  \\\midrule
			D.2 & Which operating system and which version does your phone have (e.g. Android 6.0.1)? & open-ended & -  \\\midrule
			D.3 & Which mobile phone model have you used for this experiment? & open-ended & -  \\
			\bottomrule
		\end{tabular}
	}
\end{table}

\begin{table}[!htb]
	\caption{Survey questions for the exit phase--user interface and mobile app functionality.}\label{table:exit-phase-user-interface}
	\centering
	\resizebox{\textwidth}{!}{%
		\begin{tabular}{p{0.15\textwidth}p{1.17\textwidth}p{0.15\textwidth}p{0.25\textwidth}}
			\toprule
			\textbf{ID} & \textbf{Question} & \textbf{Type}& \textbf{Options} \\
			\midrule
			D.4 & How easy was it to use the mobile app of the experiment? & ratio scale & 5 [extremely difficult,extremely easy]\\\midrule
			D.5 & How would you rate the quality of the app? & ratio scale & 5 [extremely bad,extremely good]\\\midrule
			D.6 & How satisfied are you with each of the following features of the mobile app? & group of questions & 7 questions\\
			\qquad D.6.1 & \qquad Battery consumption & ratio scale & 5 [very low,very high]\\
			\qquad D.6.2 & \qquad Performance speed & ratio scale & 5 [very low,very high]\\
			\qquad D.6.3 & \qquad Colors & ratio scale & 5 [very low,very high]\\
			\qquad D.6.4 & \qquad Formulation of questions & ratio scale & 5 [very low,very high]\\
			\qquad D.6.5 & \qquad Content of questions & ratio scale & 5 [very low,very high]\\
			\qquad D.6.6 & \qquad Number of different questions & ratio scale & 5 [very low,very high]\\
			\qquad D.6.7 & \qquad Frequency of the notifications & ratio scale & 5 [very low,very high]\\\midrule
			 D.7 & Please evaluate the following features of the mobile app: & group of questions & 12 questions\\
			\qquad D.7.1 & \qquad How comprehensible was the indicator of the rewards accumulated? (Arrow 1) & ratio scale & 5 [very little,very much]\\
			\qquad D.7.2 & \qquad How useful was the indicator of the rewards accumulated to make a choice? (Arrow 1) & ratio scale & 5 [very little,very much]\\
			\qquad D.7.3 & \qquad How comprehensible was the indicator of the total privacy level? (Arrow 2) & ratio scale & 5 [very little,very much]\\
			\qquad D.7.4 & \qquad How useful was the indicator of the total privacy level to make a choice? (Arrow 2) & ratio scale & 5 [very little,very much]\\
			\qquad D.7.5 & \qquad How comprehensible was the indicator of rewards payoff for each data-sharing level? (Arrow 3) & ratio scale & 5 [very little,very much]\\
			\qquad D.7.6 & \qquad How useful was the indicator of rewards payoff for each data-sharing level to make a choice? (Arrow 3) & ratio scale & 5 [very little,very much]\\
			\qquad D.7.7 & \qquad How comprehensible was the indicator of privacy payoff for each data-sharing level to make a choice? (Arrow 4) & ratio scale & 5 [very little,very much]\\
			\qquad D.7.8 & \qquad How useful was the indicator of privacy payoff for each data-sharing level to make a choice? (Arrow 4) & ratio scale & 5 [very little,very much]\\
			\qquad D.7.9 & \qquad How comprehensible were the five options of data sharing? (Arrow 5) & ratio scale & 5 [very little,very much]\\
			\qquad D.7.10 & \qquad How useful were the five options of data sharing? (Arrow 5) & ratio scale & 5 [very little,very much]\\
			\qquad D.7.11 & \qquad How comprehensible was the improvement box? (Arrow 6) & ratio scale & 5 [very little,very much]\\
			\qquad D.7.12 & \qquad How useful was the improvement box to make a choice? (Arrow 6) & ratio scale & 5 [very little,very much]\\\midrule
			D.8 & Do you have any other comments regarding the indicators? & open-ended & -  \\
			\bottomrule
		\end{tabular}
	}
\end{table}

\begin{table}[!htb]
	\caption{Survey questions for the exit phase--privacy and rewards.}\label{table:exit-phase-privacy-rewards}
	\centering
	\resizebox{\textwidth}{!}{%
		\begin{tabular}{p{0.15\textwidth}p{1.1\textwidth}p{0.19\textwidth}p{0.45\textwidth}}
			\toprule
			\textbf{ID} & \textbf{Question} & \textbf{Type}& \textbf{Options} \\
			\midrule
			D.9 & Please evaluate the following questions about privacy: & group of questions & 4 questions\\
			\qquad D.9.1 & \qquad Did the experiment make you feel more aware of the privacy of mobile sensor data? & ratio scale & 5 [definitely not,definitely yes]\\
			\qquad D.9.2 & \qquad Did the values of privacy represent well your choices of privacy-preservation? & ratio scale & 5 [definitely not,definitely yes]\\
			 \qquad D.9.3 & \qquad Could you easily adjust your total privacy when it was not satisfactory? & ratio scale & 5 [definitely not,definitely yes]\\
			 \qquad D.9.4 & \qquad Did your privacy-preservation choices deserved the sacrifice of rewards? & ratio scale & 5 [definitely not,definitely yes]\\\midrule
			 D.10 & How satisfied are you with the following? & group of questions & 2 questions\\
			 \qquad D.10.1 & \qquad The total available amount of rewards (30 CHF) & ratio scale & 5 [extremely dissatisfied,extremely satisfied]\\
			 \qquad D.10.2 & \qquad The amount of rewards you gained during experiment out of the total available amount of rewards & ratio scale & 5 [extremely dissatisfied,extremely satisfied]\\\midrule
			 D.11 & Please evaluate the following statements about rewards: & group of questions & 5 questions\\
			\qquad D.11.1 & \qquad Did rewards convince you to share mobile sensor data? & ratio scale & 5 [definitely not,definitely yes]\\
			\qquad D.11.2 & \qquad Did rewards convince you to share more mobile sensor data than without rewards? & ratio scale & 5 [definitely not,definitely yes]\\
			\qquad D.11.3 & \qquad Did rewards make you more aware about the privacy of mobile sensor data? & ratio scale & 5 [definitely not,definitely yes]\\
		    \qquad D.11.4 & \qquad Did rewards make you more aware about the value of mobile sensor data? & ratio scale & 5 [definitely not,definitely yes]\\
			\qquad D.11.5 & \qquad Did rewards choices deserved the sacrifice of privacy? & ratio scale & 5 [definitely not,definitely yes]\\\midrule
			D.12 & Evaluate the change in rewards payoff (Arrow 3) among the different data-sharing options & ratio scale & 5 [very low,very high]\\	\midrule
			D.13 & Evaluate the change of privacy level payoff (Arrow 4) among the different data-sharing options. & ratio scale & 5 [very low,very high]\\
			\bottomrule
		\end{tabular}
	}
\end{table}

\begin{table}[!htb]
	\caption{Survey questions for the exit phase--experiment}\label{table:exit-phase-experiment}
	\centering
	\resizebox{\textwidth}{!}{%
		\begin{tabular}{p{0.15\textwidth}p{0.8\textwidth}p{0.19\textwidth}p{0.45\textwidth}}
			\toprule
			\textbf{ID} & \textbf{Question} & \textbf{Type}& \textbf{Options} \\
			\midrule
			D.14 & Have you participated before in the following: & group of questions & 3 questions\\
			\qquad D.14.1 & \qquad An experiment at ETH Decision Science Lab? & multiple choice, one selection &  yes, no\\
			\qquad D.14.2 & \qquad A social experiment elsewhere? & multiple choice, one selection &  yes, no\\
			\qquad D.14.3 & \qquad An experiment that requires the use of a mobile app? & multiple choice, one selection &  yes, no\\\midrule
			D.15 & How interesting was the experiment? & ratio scale & 5 [not interesting at all,extremely interesting]\\\midrule
			D.16 & Would you participate in a similar experiment again? & ratio scale & 5 [definitely not,definitely yes]\\\midrule
			D.17 & How satisfied are you with the following: & group of questions & 6 questions\\
			\qquad D.17.1 & \qquad The written instructions given during the experimental process & ratio scale & 5 [extremely dissatisfied,extremely satisfied]\\
			\qquad D.17.2 & \qquad Your participation in the entry phase & ratio scale & 5 [extremely dissatisfied,extremely satisfied]\\
			\qquad D.17.3 & \qquad Your participation in the core phase & ratio scale & 5 [extremely dissatisfied,extremely satisfied]\\
			\qquad D.17.4 & \qquad Your participation in the exit phase & ratio scale & 5 [extremely dissatisfied,extremely satisfied]\\
			\qquad D.17.5 & \qquad The technical support of the staff members moderating the experiment & ratio scale & 5 [extremely dissatisfied,extremely satisfied]\\
			\qquad D.17.6 & \qquad Your participation in the overall experiment & ratio scale & 5 [extremely dissatisfied,extremely satisfied]\\\midrule
			D.18 & Has your mobile phone been turned off during the experiment? & multiple choice, one selection & yes, no\\\midrule
			D.19 & Have you run out of battery during the experiment? & multiple choice, one selection & yes, no\\\midrule
			D.20 & If yes, please provide some more information (e.g. how long, how many times, at what time of the day) & open-ended & -  \\\midrule
			D.21 & Which of the following reasons prevented you from answering more questions? & multiple choice, multiple selection & I was not interested anymore, I was not enough motivated, I faced technical problems, I ran out of battery, I was busy, I was not satisfied by the experiment, I was concerned about my privacy, other\\\midrule
			D.22 & Did you think at any time to drop out of the experiment? & multiple choice, one selection & yes, no\\\midrule
			D.23 & If yes, what was the reason? & open-ended & -  \\\midrule
			D.24 & Did you experience any of the following technical problems? & multiple choice, multiple selection & application crashed, application froze, application was too slow, network connection problems, battery drain, other\\\midrule
			D.25 & Have you been aware of the Data Collection Portal? & multiple choice, one selection & yes, no\\\midrule
			D.26 & Have you ever visited the Data Collection Portal? & multiple choice, one selection & yes, no\\ \midrule
			D.27 & How many times did you visit the Data Collection Portal? & multiple choice, one selection & never, less than three times, more than 3 times\\\midrule
			D.28 & Did you know about this experiment before participating? & multiple choice, one selection & yes, no\\
			\bottomrule
		\end{tabular}
	}
\end{table}

The exit survey begins by acquiring general information about the mobile phone used during the experiment as shown in Table~\ref{table:exit-phase-general}. Questions about the user interface and functionality of the mobile app are posed (Table~\ref{table:exit-phase-user-interface}). The ease of use and the quality of the app are evaluated in Questions~D.4 and~D.5 respectively, with the group Question~D.6 evaluating the satisfaction level of several features such as colors, formulation of questions, number of questions and others. The group Question~D.7 evaluates how comprehensible and useful the user interface features are (Figure~\ref{fig:app-core} in the main paper). These questions are used to detect possible biases that may affect data-sharing choices. The questions of Table~\ref{table:exit-phase-privacy-rewards} follow that concern the rewards and privacy. A few factors evaluated are the awareness about privacy (Question~D.9.1), ease of privacy adjustments (Question~D.9.3), satisfaction level on rewards (Question~D.10), data-sharing incentivization by rewards (Question~D.11) and other. These questions further explain the data-sharing choices made during the entry and core phase. Table~\ref{table:exit-phase-experiment} includes the following questions about the experimental process. They evaluate the satisfaction level to several experimental aspects (Question~D.17), the participation level and technical problems (Questions~D.18-D.24) as well as the user experience of the data-access portal (Questions~D.25-D.27).

After the exit survey, participants have an interview with the moderators of the experimental session. The goal of the interview is to scrutinize in a more qualitative way how participants perceive the overall experimental process as well as to discuss some behavioral artifacts observed in the data collected by the Kinvey backend during the previous phases. Moreover, when data are not successfully transferred to Kinvey, the data are manually transferred from the participants' phones to the moderators' computers after participants' consent. At the end of the interview, the moderators compute and validate the final total compensation of each participant who receives the compensation by the lab moderators before departing from DeSciL.

\subsection{Compensation and monetary incentives}\label{subsec:compensation}

The computed rewards are personalized according to the model of Section~\ref{sec:model}. The entry phase receives higher compensation as it requires the initial engagement and the execution of more complex tasks with the smartphone compared to the exit phase.

The distribution of the rewards for the app use follows a geometric progression and is implemented by transforming  Equation~\ref{eq:scenario-rewards} as follows: 

\begin{equation}\label{eq:scenario-rewards-participation}
	\rewardsSharingScenario{\participant}{\sharingScenarioIndex}=\maxRewardsSharingScenario{\participant}{\sharingScenarioIndex}\cdot \sqrt[\sharingOptions-1]{\frac{\participationBudget}{\budget}}^{\selectedSharingOption{\participant}{\sharingScenarioIndex}-1}.
\end{equation}

\noindent where \maxRewardsSharingScenario{\participant}{\sharingScenarioIndex} is the maximum rewards that can be gained in sharing scenario \sharingScenarioIndex computed by Equation~\ref{eq:scenario-max-rewards}, $\sharingOptions=5$ is the number of sharing options, \selectedSharingOption{\participant}{\sharingScenarioIndex} is the participant's selection, \participationBudget is the participation budget and \budget is the total available budget.

The allocated amounts for the compensation of participants are decided empirically after consultation with the DeSciL staff members. Factors that influence the decisions are the following: the available budget, the target of employing around 100 participants, the complexity of the designed experimental process, Swiss economy and the student profile of the participants in the DeSciL pool. The amounts reflect a trade-off: high enough to inventivize and engage participants with this novel experimental process while not too high to study data-sharing dilemmas between privacy and monetary rewards. The effectiveness of the selected amounts is evaluated using Questions D.9-D.13 of Table~\ref{table:exit-phase-privacy-rewards}. These results show that the designed rewards were effective for their purpose. A 57.7\% of the participants were too busy to answer more questions, while 33.6\% needed further motivation (Question D.21). 

\begin{figure}[!htb]
	\centering
	\includegraphics[width=0.49\textwidth]{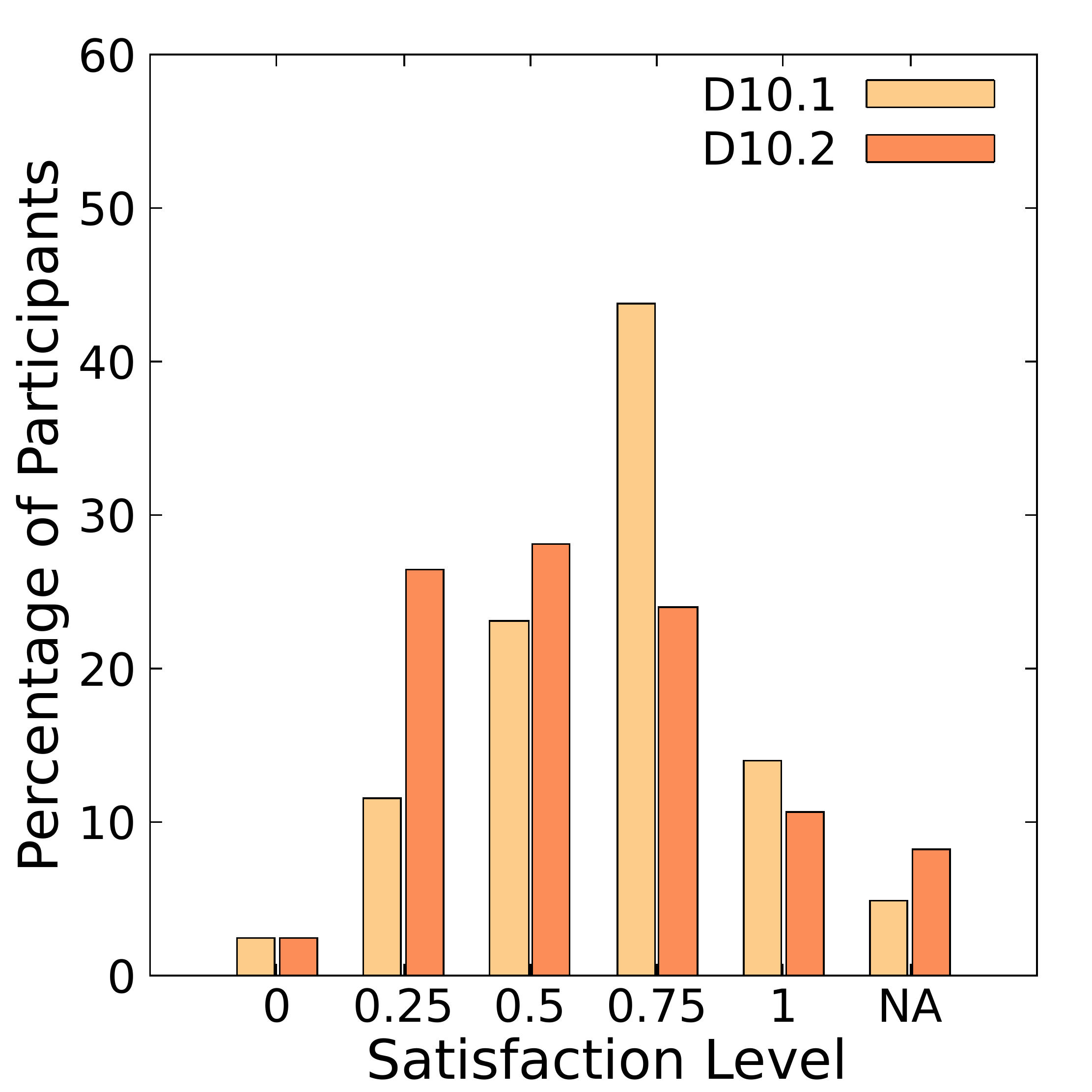}
	\includegraphics[width=0.49\textwidth]{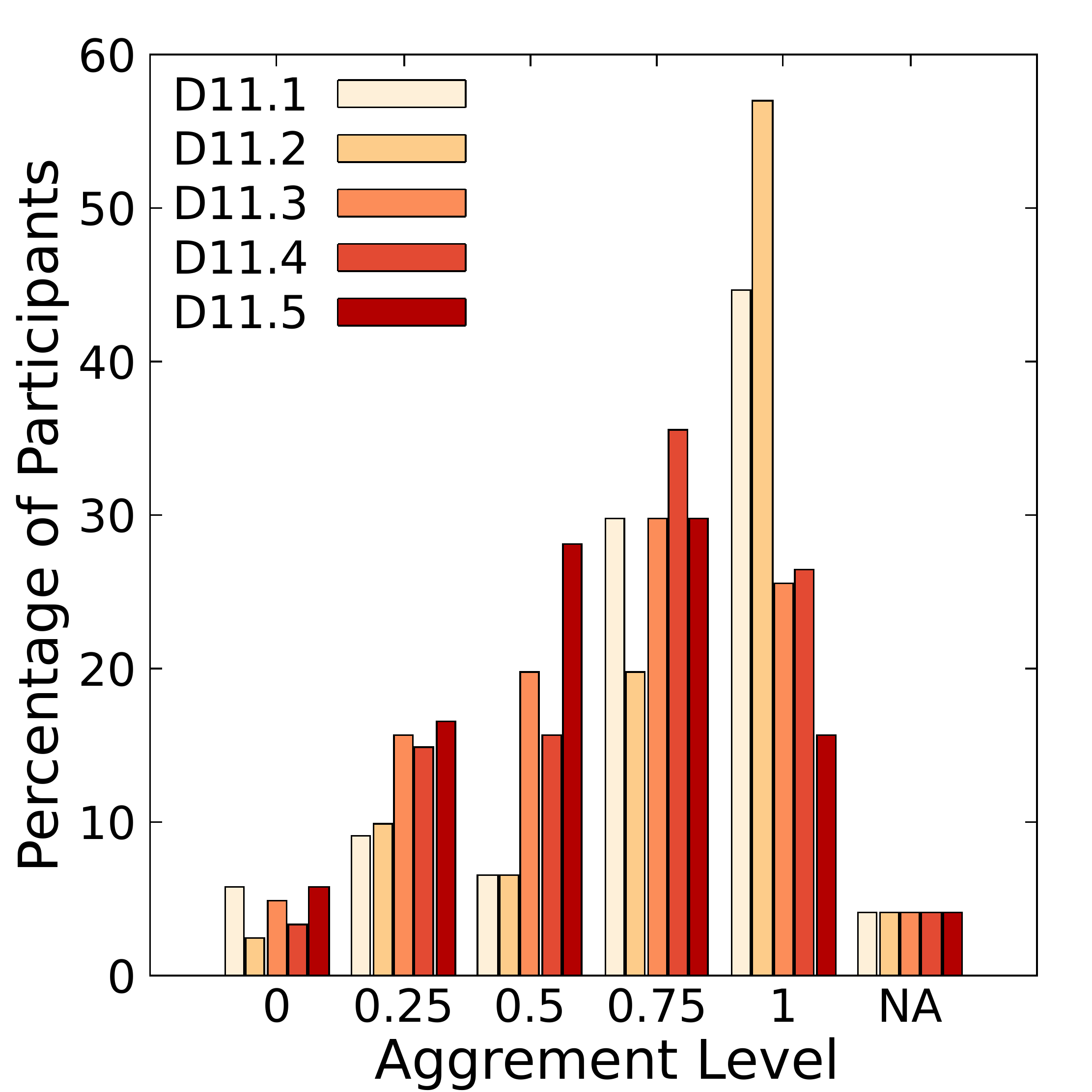}
	\caption{The assessment of the received rewards by the participants of the experiment using group Questions D.10 and D.11.}\label{fig:rewards-evaluation}
\end{figure}

\section{Implementation of the Technical Infrastructure}\label{sec:infrastructure}

The data collected by participants' smartphone app are stored and managed locally by an implementation of the nervousnet framework~\cite{Pournaras2015} that provides high-level application programming interfaces (APIs) to store, query and analyze data on smartphones. Remotely on the server, the data are stored and managed by Kinvey~\cite{Kinvey2016} that provides secure communication by using TLS/SSL encryption between smartphones and the Kinvey backend. The data-access web portal relies on Node.js and a MongoDB database. 

The quality of the app (Question D.5) is evaluated 61\% positively. The mobile phone remained turned on during the experiment in 82.7\% of the participants (Question D.18), while only 13.8\% of the participants ran out of battery (Question D.19) and a 25.9\% reported battery drain problems (Question D.24).

\section{The Privacy and Rewards Gain of Data-sharing Scenarios}\label{subsec:gain-scenarios}

\begin{figure}[!htb]
	\centering
	\subfigure[Privacy gain of data-sharing scenarios]{\includegraphics[width=1.0\textwidth]{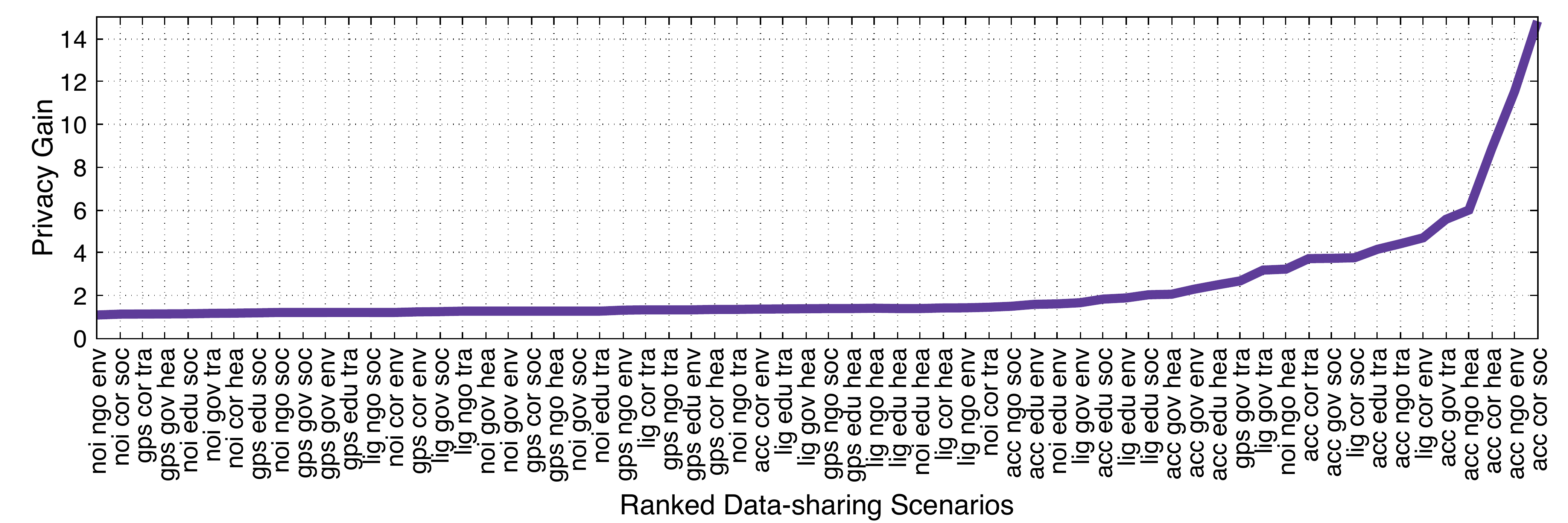}}
	\subfigure[Reward gain of data-sharing scenarios]{\includegraphics[width=1.0\textwidth]{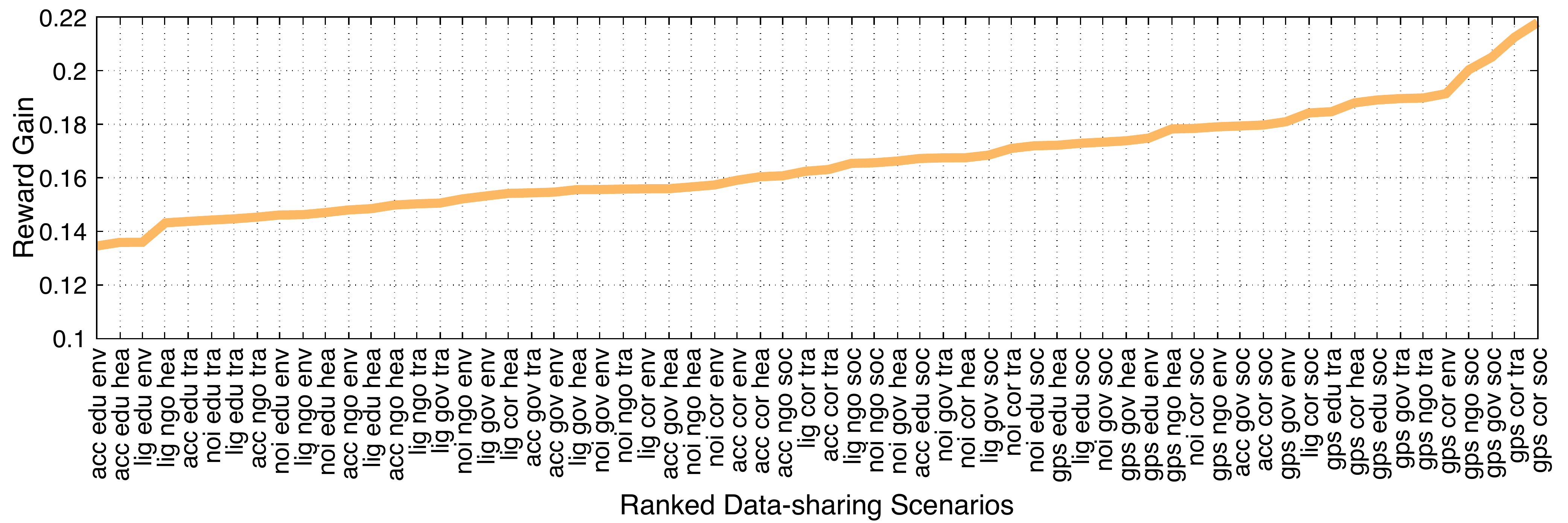}}
	\caption{Mean privacy and reward gain for the 64 data-sharing scenarios under rewarded data sharing. The scenarios are sorted from lowest to highest gain.}\label{fig:privacy-reward-gain-scenarios}
\end{figure}

Figure~\ref{fig:privacy-reward-gain-scenarios} illustrates the mean privacy and reward gain of the data-sharing scenarios retrieved as a response of choosing to improve privacy and rewards respectively (see Figure~\ref{fig:app-core} in the main paper). 

Table~\ref{tab:privacy-reward-gain-elements} outlines the mean privacy and reward gain of the different data-sharing elements that consist the 64 data-sharing scenarios. 

\begin{table}[!htb]
	\caption{Mean privacy and reward gain of the different data-sharing elements involved in the data-sharing scenarios.}\label{tab:privacy-reward-gain-elements}
	\centering
		\begin{tabular}{lllllllllllll}\toprule
			Mean Gain & acc & lig & noi & gps & cor & ngo & gov & edu & soc & env & tra & hea \\\midrule
			Privacy & 4.75	& 1.92 & 1.4 & 1.36 & 3.13 & 2.56 & 2.02 & 1.71 & 2.5 & 2.3 & 2.29 & 2.33 \\\midrule
			Reward & 0.16 & 0.16 & 0.16 & 0.19 & 0.18 & 0.16 & 0.17 & 0.16 & 0.18 & 0.16 & 0.16 & 0.16 \\\bottomrule
		\end{tabular}
\end{table}

\section{Privacy Loss and Rewarded Data-sharing Choices of Groups}\label{sec:privacy-loss-groups}

\begin{figure}[!htb]
	\centering
	\subfigure[Probability density function of privacy for intrinsic and rewarded data sharing.]{\includegraphics[width=0.32\textwidth]{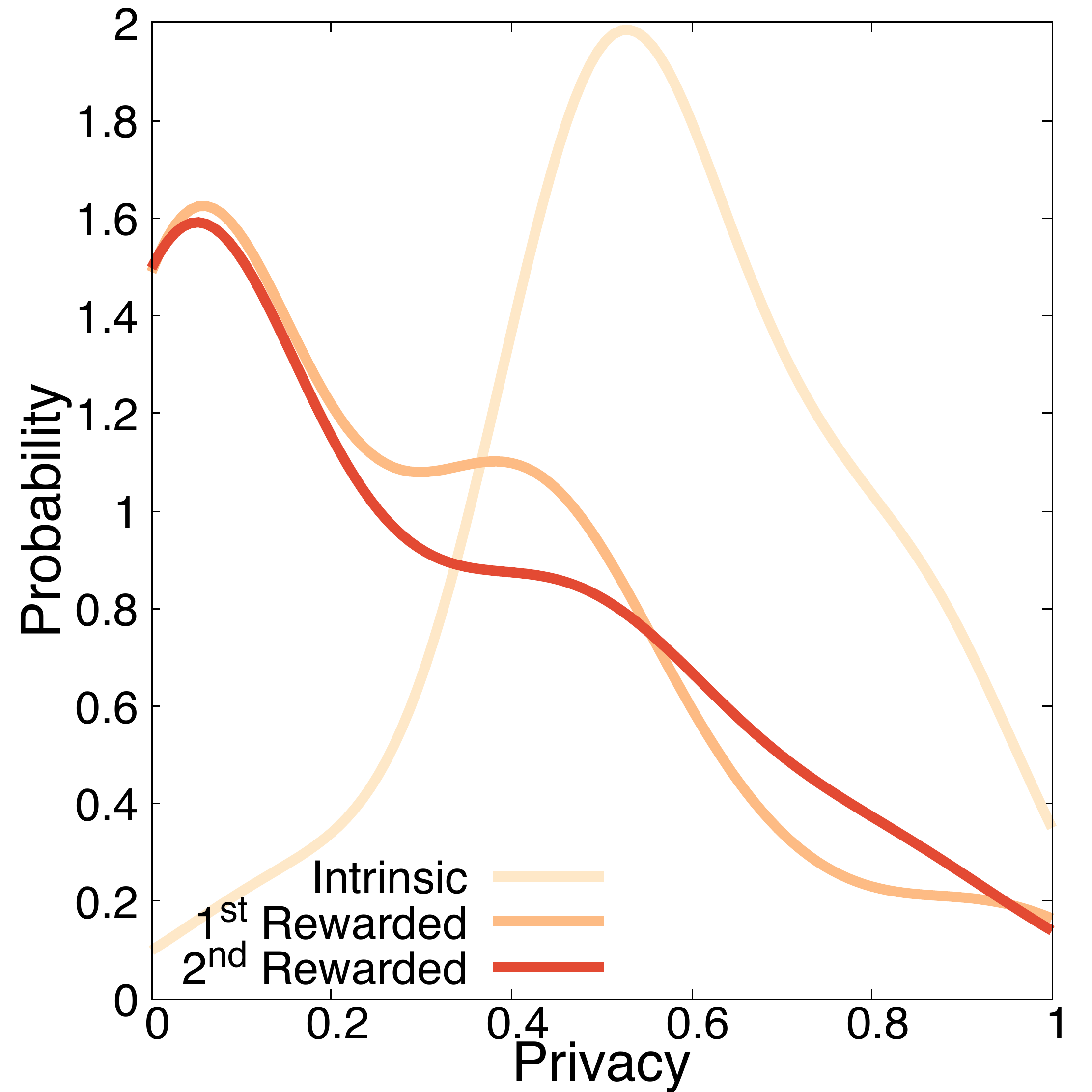}}
	\subfigure[Cumulative density function of privacy for intrinsic and rewarded data sharing.]{\includegraphics[width=0.32\textwidth]{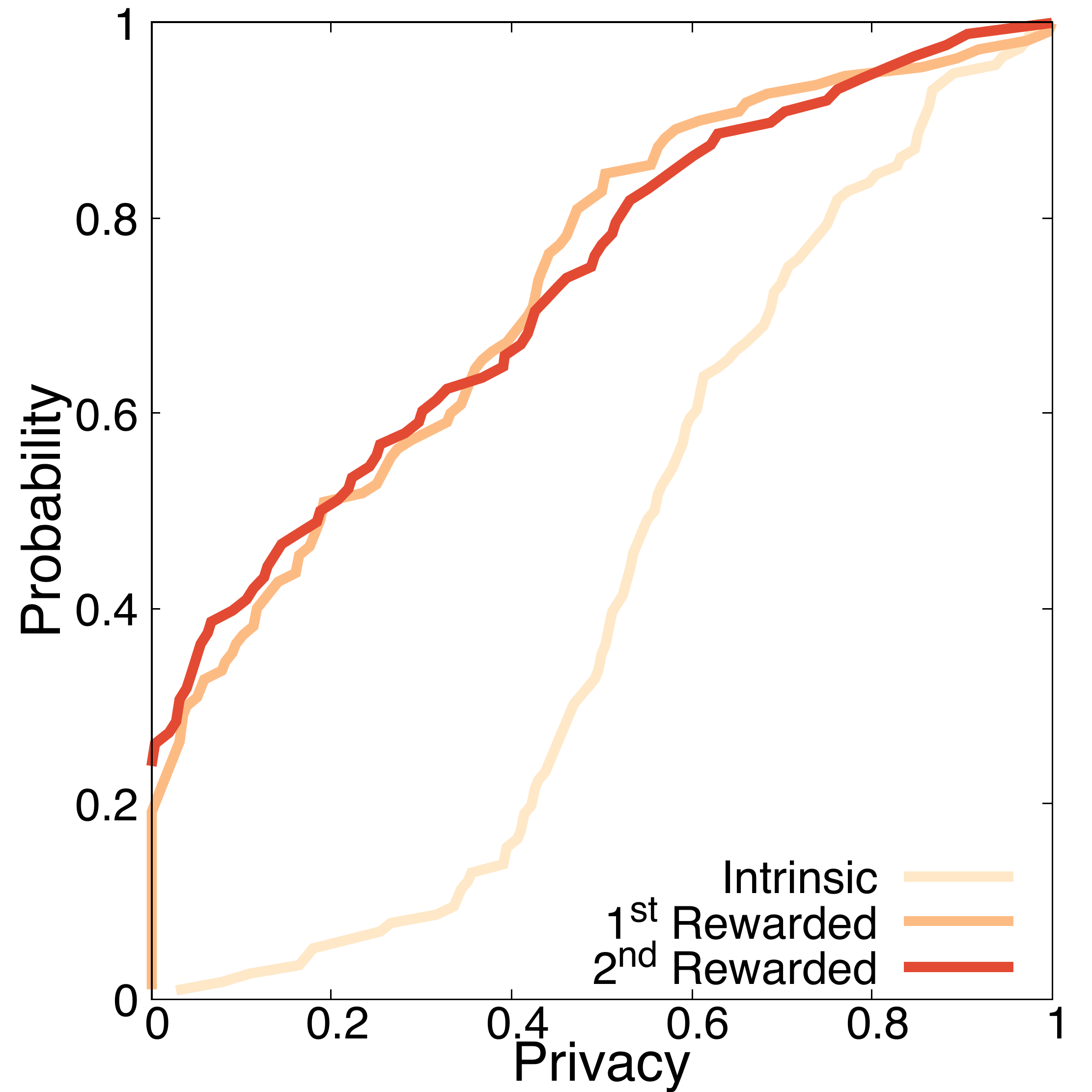}}
	\subfigure[Mean privacy of groups over consecutive rewarded data-sharing choices. ]{\includegraphics[width=0.32\textwidth]{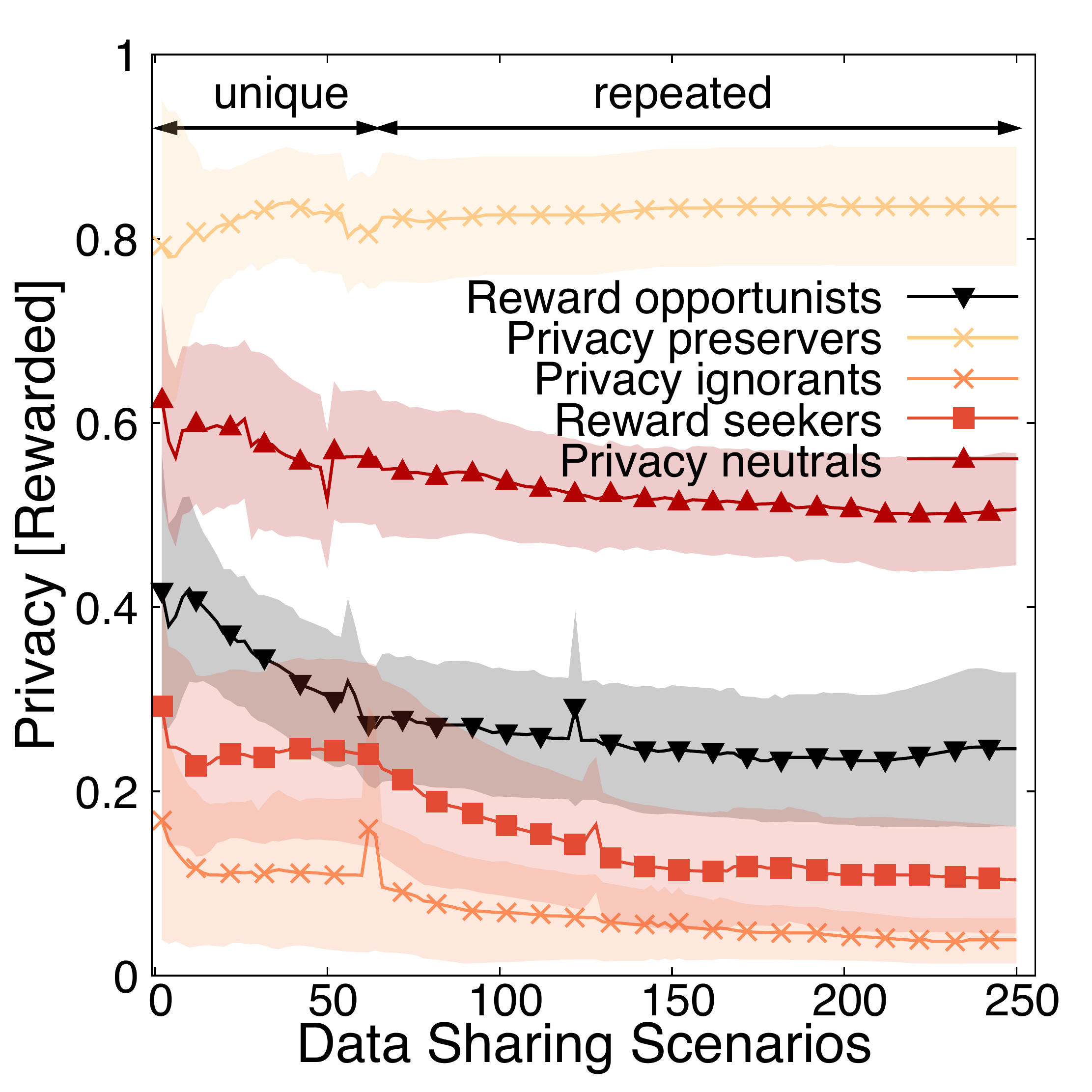}}
	\caption{Privacy loss under rewarded data sharing and the behavior of the groups over repeated data-sharing choices under the \nth{2} rewarded data sharing.}\label{fig:privacy-loss-groups}
\end{figure}

Figure~\ref{fig:privacy-loss-groups}a and~\ref{fig:privacy-loss-groups}b illustrate the probability and cumulative density functions for the intrinsic, \nth{1} and \nth{2} rewarded data sharing. The two experimental conditions for rewarded data sharing show very similar densities, while intrinsic data sharing comes with a single peak around the privacy level of 0.55. 

Figure~\ref{fig:privacy-loss-groups}c shows the privacy level over consecutive data-sharing choices under the \nth{2} rewarded data sharing. Compared to Figure~\ref{fig:behavioral-patterns}b in the main paper showing the \nth{1} rewarded data sharing, the group behaviors are similar. Reward opportunists show a further decline of their privacy level.

\section{Goal Signals for Coordinated Data Sharing}\label{sec:goal-signals}

\begin{figure}[!htb]
	\centering
	\includegraphics[width=0.8\textwidth]{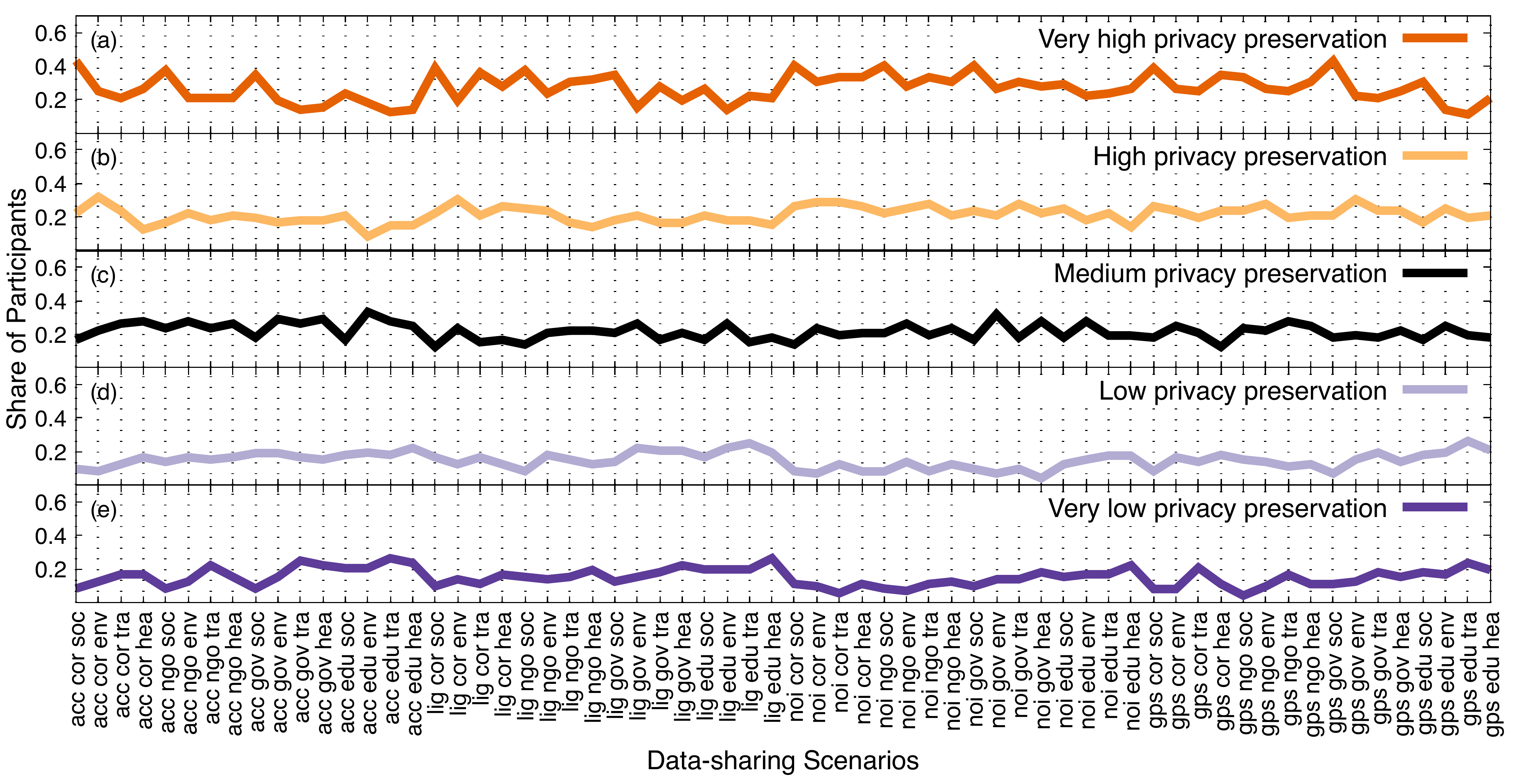}
	\caption{The five goal signals of privacy preservation: from very high to very low. For each signal, the values of a data-sharing scenario measure the share of participants that choose a certain level of privacy preservation.}\label{fig:coordinated-data-sharing}
\end{figure}

Figure~\ref{fig:coordinated-data-sharing} illustrates the five goal signals of privacy preservation. They represent a distribution of required amount of data over the 64 data-sharing scenarios. They are referred within the range of very high to very low privacy preservation. This is because each signal measures the ratio of participants that choose a certain data-sharing level for each data-sharing scenario under intrinsic data sharing. Note that for each data-sharing scenario in Figure~\ref{fig:coordinated-data-sharing}, the shares of participants sum up to 1. 

%

\section{Data-sharing Mismatch}\label{sec:mismatch}

\begin{figure}[!htb]
	\centering
	\includegraphics[width=1.0\textwidth]{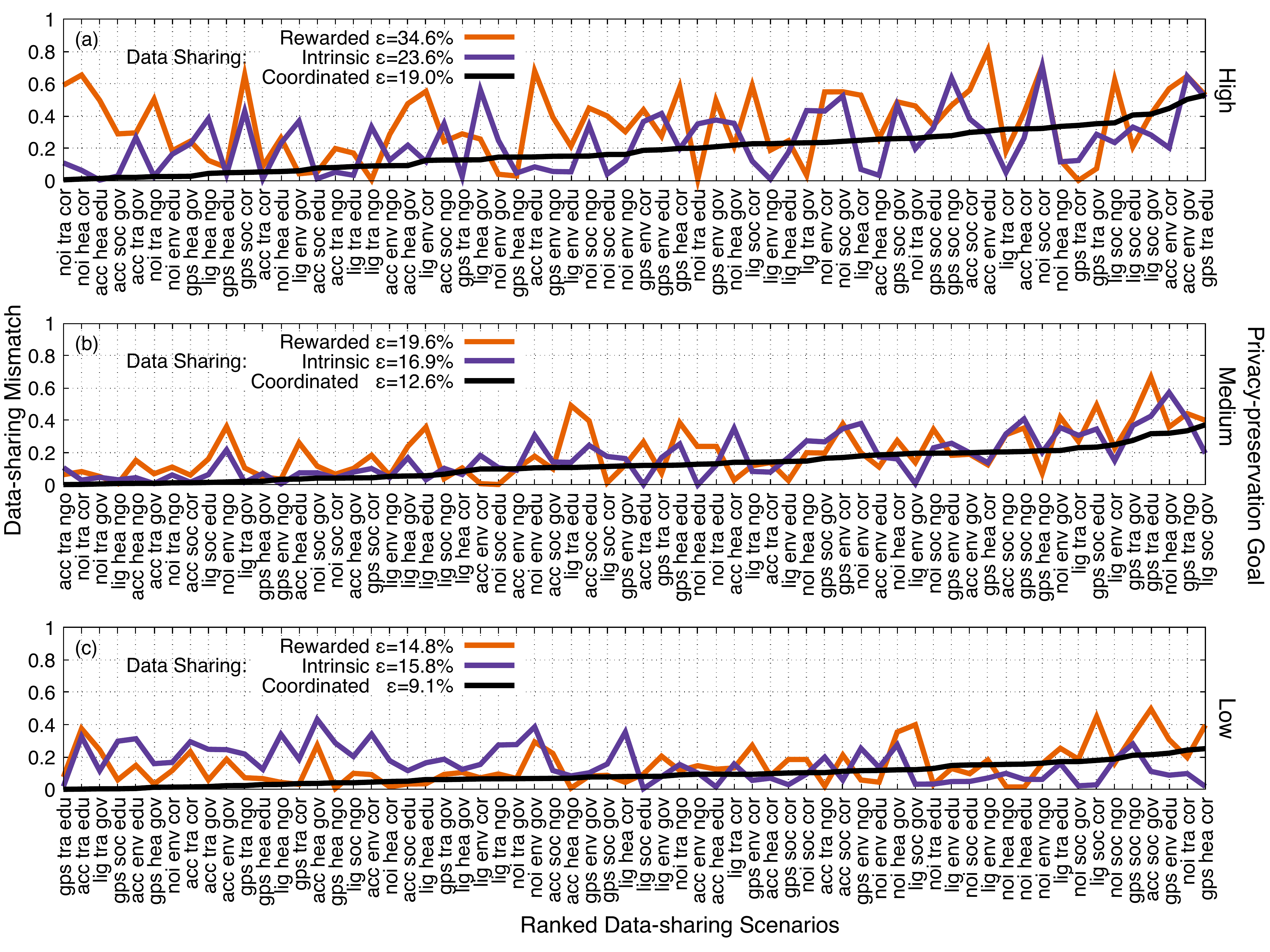}
	\caption{Data-sharing mismatch (root mean square error $\varepsilon$) for the 64 data-sharing scenarios and for the three goal signals of high, medium and low privacy preservation. Values are sorted from lowest to highest mismatch according to the the coordinated data sharing. Coordinated data sharing shows higher efficiency than intrinsic and rewarded data sharing.}\label{fig:privacy-mismatch-scenarios-L-M-H}
\end{figure}

Figure~\ref{fig:privacy-mismatch-scenarios-L-M-H} shows the data-sharing mismatch for the rest of the three goal signals of privacy preservation: low, medium and high. The results here confirm the findings illustrated in Figure~\ref{fig:privacy-mismatch-scenarios}b of the main paper: mismatch increases for higher privacy-preservation goals as agents mainly have one privacy preserving option (intrinsic) to choose from. 

\section{Valuations of Collective Privacy Recovery}\label{sec:privacy-valuation schemes}

Four different valuations of privacy are compared in Table~\ref{tab:valuations}. All valuations are a function of \rewardsSharingScenarios{\participant} that is the mean privacy level over all data-sharing scenarios, measured by the gained rewards as outlined in Equation~\ref{eq:scenario-rewards}:

\begin{table}[!htb]
	\caption{Four valuation schemes and their range of values.}\label{tab:valuations}
	\centering
	\resizebox{\columnwidth}{!}{%
		\begin{tabular}{lllllll}\toprule
			\multirow{2}{*}{Valuation} & \multirow{2}{*}{Relation} & \multirow{2}{*}{min/max \rewardsSharingScenariosIntrinsic{\participant}} & \multicolumn{2}{c}{Without Rewards ($\rewardsSharingScenarios{\participant}\equiv \rewardsSharingScenariosIntrinsic{\participant}$)} & \multicolumn{2}{c}{With Rewards}
			\\\cmidrule(lr){4-5}\cmidrule(lr){6-7}
			& &   & min \privacyCost{\participant}{0} & max \privacyCost{\participant}{17.5} & min \privacyCost{\participant}{0} &  max \privacyCost{\participant}{17.5} \\\midrule	
			Absolute shared data & $\privacyCost{\participant}{\rewardsSharingScenarios{\participant}}=\rewardsSharingScenarios{\participant}$ &  & 0 & 17.5 & 0 & 17.5  \\
			Absolute sacrificed rewards & $\privacyCost{\participant}{\rewardsSharingScenarios{\participant}}=\rewardsSharingScenarios{\participant} -  \sharingBudget$ & & -17.5 & 0 & -17.5 &  0 \\
			\multirow{2}{*}{Relative shared data} & \multirow{2}{*}{$\privacyCost{\participant}{\rewardsSharingScenarios{\participant}}=\rewardsSharingScenarios{\participant} - \rewardsSharingScenariosIntrinsic{\participant}$} &  $\rewardsSharingScenariosIntrinsic{\participant}=0$ & 0 & 0 & 0 &  17.5  \\
			& & $\rewardsSharingScenariosIntrinsic{\participant}=17.5$ & 0 & 0 & -17.5 &  0 \\
			\multirow{2}{*}{Relative sacrificed rewards} & \multirow{2}{*}{$\privacyCost{\participant}{\rewardsSharingScenarios{\participant}}=\rewardsSharingScenarios{\participant} - ( \sharingBudget-\rewardsSharingScenariosIntrinsic{\participant})$} & $\rewardsSharingScenariosIntrinsic{\participant}=0$ & -17.5 & -17.5& -17.5 &  0 \\
			& & $\rewardsSharingScenariosIntrinsic{\participant}=17.5$ & -17.5 & 17.5 & 0 &  17.5 \\\bottomrule			
		\end{tabular}
	}
\end{table}

\begin{itemize}
	\item \textbf{Absolute shared data}: The privacy cost \privacyCost{\participant}{\rewardsSharingScenarios{\participant}} equals the gained rewards \rewardsSharingScenarios{\participant}. This is the default valuation used throughout the main paper. The minimum privacy cost is $0$, while the maximum is $17.5$ that is the maximum rewards that an individual could gain in the lab experiment.
	\item \textbf{Absolute sacrificed rewards}: The privacy cost \privacyCost{\participant}{\rewardsSharingScenarios{\participant}} equals the gained rewards \rewardsSharingScenarios{\participant} minus the fixed data-sharing rewards \sharingBudget. This scheme is equivalent to the absolute shared data as \sharingBudget is constant. This valuation measures more directly the loss of rewards in exchange of privacy preservation. The minimum privacy cost is $-17.5$, while the maximum one is $0$. 
	\item \textbf{Relative shared data}: The privacy cost \privacyCost{\participant}{\rewardsSharingScenarios{\participant}} equals the gained rewards \rewardsSharingScenarios{\participant} minus the privacy level under intrinsic data sharing, measured as well in terms of (hypothetical) gained rewards (\rewardsSharingScenariosIntrinsic{\participant}). The privacy cost of this scheme measures the additional privacy loss under rewarded data sharing over the intrinsic one, assuming that the intrinsic data sharing comes with no privacy cost. Depending on the level of intrinsic data sharing, the minimum privacy cost is $-17.5$, while the maximum is $17.5$ (the behavior of reward opposer and reward opportunist respectively as shown in Table~\ref{tab:groups} of the main paper). 
	\item \textbf{Relative sacrificed rewards}: The privacy cost \privacyCost{\participant}{\rewardsSharingScenarios{\participant}} equals the gained rewards \rewardsSharingScenarios{\participant} minus the privacy preservation under intrinsic data sharing measured by $\sharingBudget-\rewardsSharingScenariosIntrinsic{\participant}$. This scheme is equivalent to the one of absolute sacrificed rewards with the addition of the privacy cost \rewardsSharingScenariosIntrinsic{\participant} under intrinsic data sharing. Depending on the level of intrinsic data sharing, the minimum privacy cost is $-17.5$, while the maximum one is $17.5$.
\end{itemize}

\begin{figure}[!htb]
	\centering
	\includegraphics[width=1.0\textwidth]{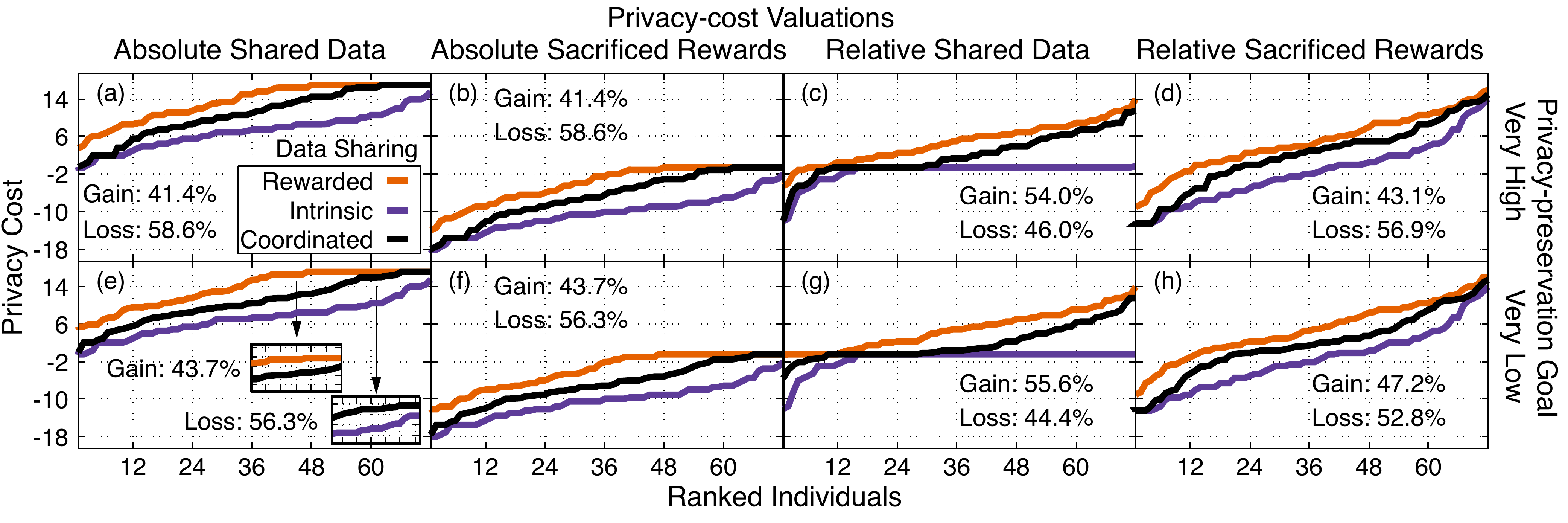}
	\caption{The four privacy valuations illustrated in Table~\ref{tab:valuations}. The privacy cost is measured for the intrinsic, rewarded and coordinated data sharing under the very high and very low privacy preservation goal. The highest privacy gain is observed for the relative shared data and the relative sacrificed rewards. }\label{fig:privacy-cost-valuations}
\end{figure}

The collective privacy recovery under intrinsic, rewarded and coordinated data sharing is assessed using the four different valuations schemes under the very high and very low privacy preservation goal. Figure~\ref{fig:privacy-cost-valuations} shows the privacy cost per individual for each of these cases. All lines are sorted from lowest to highest privacy cost. Each plot in Figure~\ref{fig:privacy-cost-valuations} comes with the mean relative privacy gain and loss of the coordinated data sharing compared to rewarded and intrinsic data sharing respectively.  The privacy cost of intrinsic data sharing corresponds to the data-sharing plan with the minimum privacy cost and it is calculated using EPOS with $\alpha=0, \beta =1$. In contrast, the privacy cost of rewarded data sharing corresponds to the data-sharing plan with the maximum privacy cost and it is calculated using EPOS with $\alpha=0, \beta =1$ and data-sharing plans with reversed sign. 

The highest privacy gains are observed under the valuation scheme of relative shared data: 54\% and 55.6\% for the very high and very low privacy preservation goal. This means that coordinated data sharing shows a further privacy recovery when evaluating the data-sharing choices based on the additional privacy cost that individuals pay over the intrinsic data sharing. The relative sacrificed rewards follow with 43.1\% and 47.2\% respectively. The default valuation scheme of absolute shared data has the lowest privacy gain of 41.4\% and 43.7\% respectively, which equals the absolute sacrificed rewards as \sharingBudget is constant (lines shifted to negative values). The mean privacy gain for the very low privacy preservation goal is 2.7\% higher than the very high one. Similarly with the observation in Figure~\ref{fig:privacy-mismatch-scenarios}a of the main paper, two rewarded options of individuals with low privacy on average provide higher flexibility than a single one with high privacy preservation. 

With their higher privacy gains, the alternative valuation schemes find applicability in the further adoption of the data-sharing plans recommended to users. They can also be used to provide augmented explanations of what these recommended plans mean for the data collective, while raising awareness of the different privacy manifestations and collective privacy gains.

\section{Privacy Reinforcement}\label{sec:privacy-reinforcement}

\begin{figure}[!htb]
	\centering
	\includegraphics[width=0.49\textwidth]{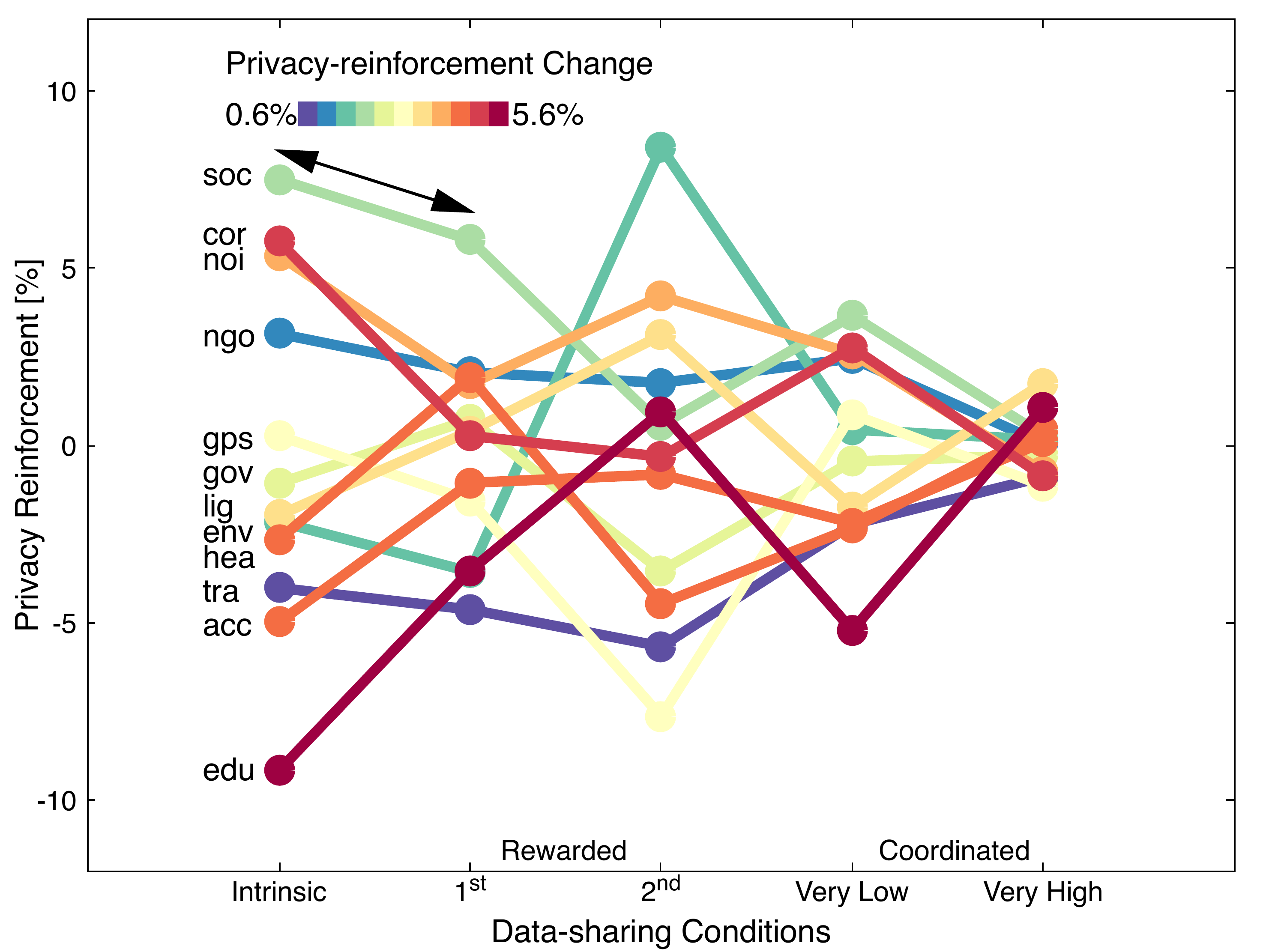}
	\caption{Privacy reinforcement of the different sensors, collectors and contexts under intrinsic, rewarded and coordinated data sharing. The 12 colored lines are ranked according to the reinforcement change (intrinsic - \nth{1} rewarded data sharing).}\label{fig:privacy-reinforcement}
\end{figure}

Figure~\ref{fig:privacy-reinforcement} illustrates the privacy reinforcement of the different data-sharing elements. The key finding is that the perceived privacy sensitivity of the data-sharing elements is likely to reinforce privacy under intrinsic and coordinated data sharing rather than the rewarded ones.

The mean absolute privacy reinforcement under intrinsic data sharing is higher than \nth{1} rewarded and the two coordinated data-sharing conditions: 4\%>2.27\%> 2.24\%>0.65\% respectively. Under intrinsic data sharing, social networking, corporation, noise sensor and NGO reinforce privacy gain, while education, accelerometer and transportation a privacy loss. Privacy reinforcement under intrinsic data sharing is correlated with the attitudinal privacy sensitivity ($R=0.63, t(10)=2.57, p=0.028$). This means that privacy risk awareness is likely to reinforce privacy protection. There is a correlation in the privacy reinforcement under intrinsic and the \nth{1} rewarded data sharing ($R=0.73, t(10)=3.4, p=0.0067$). In the \nth{2} rewarded data sharing, GPS shifts to a 3.5\% reinforcement of privacy loss, while environment shifts to a 8.4\% reinforcement of privacy gain. Coordinated data sharing with the very low privacy-preservation goal is positively correlated to attitudinal ($R=0.65, t(10)=3.68, p=0.023$), intrinsic ($R=0.96, t(10)=11.58, p=\num{4.07e-7}$) and the \nth{1} rewarded ($R=0.62, t(10)=2.48, p=0.032$) data sharing. With the very high privacy-preservation goal, the correlation to attitudinal data sharing is negative: $R=-0.61, t(10)=-2.43, p=0.035$.

\section{Conjoint Analysis}\label{sec:conjoint-analysis}

The assumptions of conjoint analysis are discussed and assessed in the context of the conducted experiment~\cite{Hainmueller2014}. No direct carryover effects are involved under instinct data sharing as participants are exposed to each data-sharing scenario once. Under rewarded data sharing, the privacy-rewards balance introduces a carryover effect that is subject of study in this paper. Because rewards are personalized (i.e. each data-sharing scenario is retrieved to satisfy the intended action of improving rewards or privacy) and because responses to repeated data-sharing scenarios are made on-demand, carryover effects mainly originate from tuning the privacy-rewards balance. No influential order effects are anticipated within the designed rating-based conjoint experiment. 
In regards to the order of the data-sharing elements, each data-sharing scenario is presented in natural language as determined by the Factorial Question in Section~\ref{subsec:experiment} of the main paper. Decision-making quality is not expected to decrease for $\numOfCriteria=3<10$ data-sharing criteria as shown in earlier experimental tests in literature~\cite{Malhotra1982,Hainmueller2014}. 
As this is not a choice-based conjoint experiment, the order of the data-sharing levels (Figure~\ref{fig:app-core}b of the main paper) simply adheres to design principles of likert scales and graphical user interfaces. 
As the experiment relies on a full-factorial design without rendering any data-sharing scenario as infeasible, order effects among the scenarios are unlikely. It is though personalization under rewarded data sharing that can yield, in theory, atypical data-sharing choices, i.e. one that can increase the accumulated rewards when a participant chooses to improve privacy, and vice versa. Excluding these or reducing the likelihood of their occurrence is expected to improve external validity~\cite{Hainmueller2014}, i.e. participants do not lose interest or react contrary to their privacy-reward goal improvement. 

The performed conjoint analysis relies on the following multiple linear regression model:

\begin{equation}\label{eq:regression-model}
	\coefficient{0}{0}+\coefficient{1}{1}\cdot \dummyVariable{1}{1}+...+ \coefficient{1}{\numOfCriterionDimensions{1}-1}\cdot \dummyVariable{1}{\numOfCriterionDimensions{1}-1}+...+\coefficient{\numOfCriteria}{\numOfCriterionDimensions{\numOfCriteria}-1} \cdot \dummyVariable{\numOfCriteria}{\numOfCriterionDimensions{\numOfCriteria}-1}+\regressionError
\end{equation}

\noindent where $\coefficient{\criterion}{\criterionDimensionIndex}$ for each of the criteria $\criterion \in \{1,...,\numOfCriteria\}$ and elements $\criterionDimensionIndex \in \{1,...,\numOfCriterionDimensions{\numOfCriteria}-1\}$ are the estimated coefficients of the regression model, with $\coefficient{0}{0}$ representing the intercept and $\regressionError$ the regression error. The $\dummyVariable{\criterion}{\criterionDimensionIndex}$ for each of the criteria $\criterion \in \{1,...,\numOfCriteria\}$ and elements $\criterionDimensionIndex \in \{1,...,\numOfCriterionDimensions{\numOfCriteria}-1\}$ are the independent dummy variables that represent the absence or presence of a data-sharing element within a data-sharing scenario. Note that one data-sharing element for each criterion is removed from the model (accelerometer, corporation, social networking) to resolve the linear dependency with which the effect of the confounded variables cannot be separated by the regression. 

Using the estimated coefficients, the partworth utilities can be estimated for each data-sharing criterion $\criterion$ as follows: 

\begin{equation}\label{eq:criterion-utility}
	\criterionUtility{\criterion}=\frac{\max_{\criterionDimensionIndex=1}^{\numOfCriterionDimensions{\criterion}}\coefficient{\criterion}{\criterionDimensionIndex}-\min_{\criterionDimensionIndex=1}^{\numOfCriterionDimensions{\criterion}}\coefficient{\criterion}{\criterionDimensionIndex}}{\sum_{\criterion=1}^{\numOfCriteria}(\max_{\criterionDimensionIndex=1}^{\numOfCriterionDimensions{\criterion}}\coefficient{\criterion}{\criterionDimensionIndex}-\min_{\criterionDimensionIndex=1}^{\numOfCriterionDimensions{\criterion}}\coefficient{\criterion}{\criterionDimensionIndex})}. 
\end{equation}

\noindent The partworth utilities measure the relative importance of the criteria within a regression model: which of the data type, collector or context is the most important when individuals make data-sharing decisions. Similarly, the relative importance of each data-sharing element for each criterion is calculated as follows:

\begin{equation}\label{eq:element-utility-per-criterion}
	\elementUtilityPerCriterion{\criterion}{\criterionDimensionIndex}=\frac{\coefficient{\criterion}{\criterionDimensionIndex}-\frac{1}{\numOfCriterionDimensions{\criterion}}\cdot\sum_{\criterionDimensionIndex=1}^{\numOfCriterionDimensions{\criterion}}\coefficient{\criterion}{\criterionDimensionIndex}}{\sum_{\criterion=1}^{\numOfCriteria}(\max_{\criterionDimensionIndex=1}^{\numOfCriterionDimensions{\criterion}}\coefficient{\criterion}{\criterionDimensionIndex}-\min_{\criterionDimensionIndex=1}^{\numOfCriterionDimensions{\criterion}}\coefficient{\criterion}{\criterionDimensionIndex})}.
\end{equation}

\noindent The relative importance calculation can be adjusted for each data-sharing element among all criteria as follows: 

\begin{equation}\label{eq:element-utility}
	\elementUtility{\criterion}{\criterionDimensionIndex}=\frac{\coefficient{\criterion}{\criterionDimensionIndex}-\frac{1}{\numOfCriteria\cdot \numOfCriterionDimensions{\criterion}}\cdot\sum_{\criterion=1}^{\numOfCriteria}\sum_{\criterionDimensionIndex=1}^{\numOfCriterionDimensions{\criterion}}\coefficient{\criterion}{\criterionDimensionIndex} }{\max_{\criterion=1}^{\numOfCriteria} \max_{\criterionDimensionIndex=1}^{\numOfCriterionDimensions{\criterion}}\coefficient{\criterion}{\criterionDimensionIndex}-\min_{\criterion=1}^{\numOfCriteria} \min_{\criterionDimensionIndex=1}^{\numOfCriterionDimensions{\criterion}}\coefficient{\criterion}{\criterionDimensionIndex}}.
\end{equation}

\begin{figure}[!htb]
	\centering
	\includegraphics[width=1.0\textwidth]{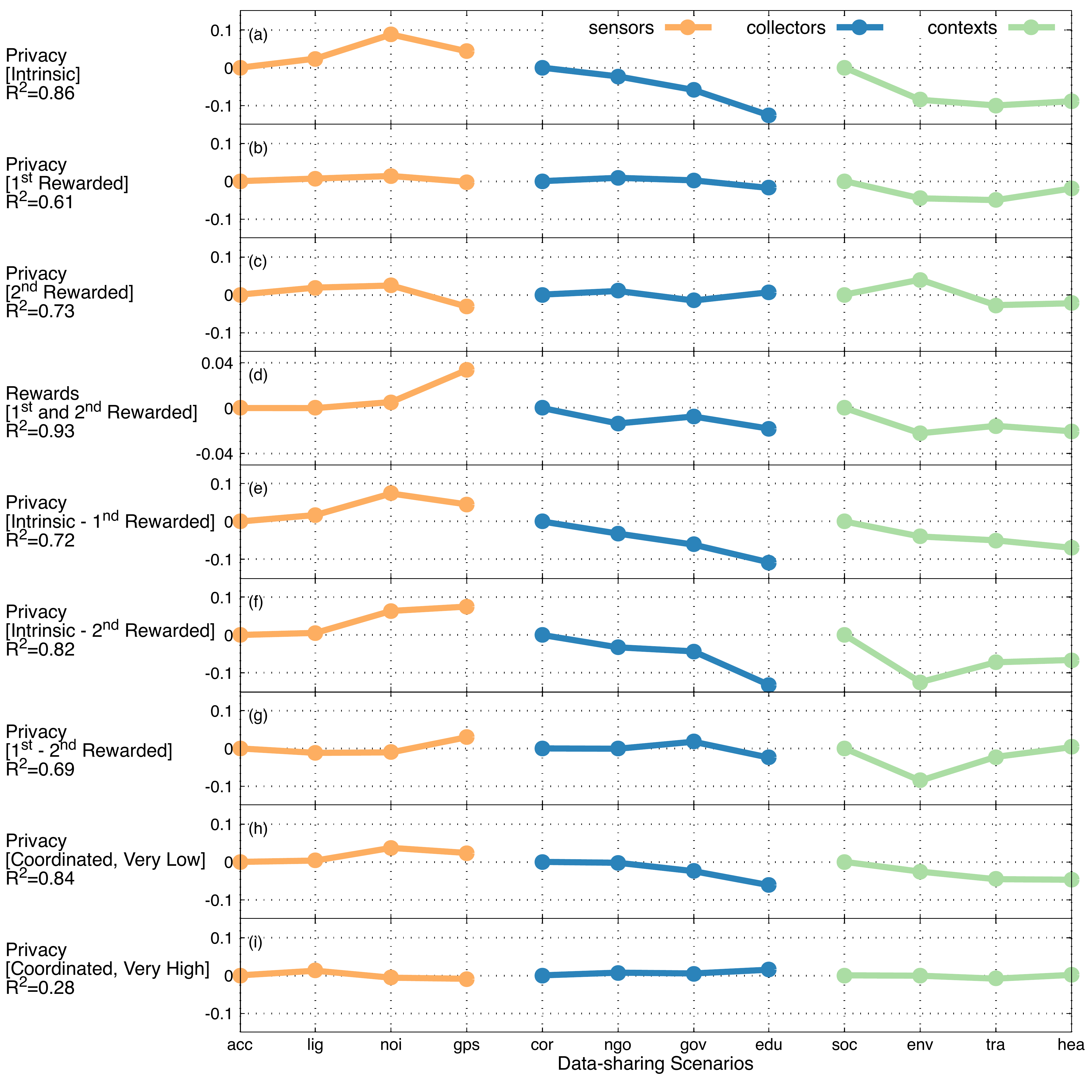}
	\caption{Coefficients of multiple linear regression used in conjoint analysis. Nine models with different dependent variables for privacy and rewards are compared. Four of these models with $R^2>0.8$ are shown in the main paper, Figure~\ref{fig:conjoint-analysis-per-criterion}a.}\label{fig:conjoint-coefficients}
\end{figure}

The model of Equation~\ref{eq:regression-model} is evaluated at the population level for different dependent variables of privacy $\privacyScenario{\sharingScenarioIndex}$ and rewards $\rewardScenario{\sharingScenarioIndex}$ with values over the 64 data-sharing scenarios. These variables are selected among the different experimental conditions and they determine the compared conjoint analysis models. The regression coefficients are illustrated in Table~\ref{tab:coefficients} and Figure~\ref{fig:conjoint-coefficients}. The rest of the conjoint analysis and metrics are shown in Table~\ref{tab:conjoint-results}. 

\begin{table}[!htb]
	\caption{The coefficients $\coefficient{\criterion}{\criterionDimensionIndex}$ of nine multiple regression models, each with a different dependent variable of privacy or rewards. The four statistically more powerful models ($R^{2}$>0.8) are illustrated in the main paper. These values are used to analyze the relative importance of data-sharing criteria and elements in Table~\ref{tab:conjoint-results}.}\label{tab:coefficients}
	\centering
	\resizebox{\columnwidth}{!}{%
		\begin{tabular}{llllllllllllll}\toprule
			Models & acc & lig & noi & gps & cor & ngo & gov & edu & soc & env & tra & hea & Intercept \\\midrule
			Privacy &0	& 0.023972603	& 0.087756849	& 0.043450342 & 0 & -0.023116438 & -0.058861301	& -0.125856164 & 0 & -0.084974315	& -0.099957192	& -0.088827055 & 0.654430651\\
			\text{[Intrinsic]} &   &	& & & & & & & & & & &\\\midrule
			Privacy &0	& 0.006753024	& 0.013699034	& -0.001945327 & 0 & 0.008754161 & 0.002140409	& -0.018011013 & 0 & -0.045397867	& -0.050165302	& -0.019421827 & 0.343937021\\
			\text{[\nth{1} Rewarded]} &   &	& & & & & & & & & & &\\\midrule
			Privacy &0	& 0.018454293	& 0.024093745	& -0.031555723 & 0 & 0.009563823 & -0.015594024	& 0.006242213 & 0 & 0.039665329	& -0.02786132	& -0.022817284 & 0.310195822\\
			\text{[\nth{2} Rewarded]} &   &	& & & & & & & & & & &\\\midrule
			Rewards &0	& -0.0000117	& 0.005297903	& 0.033532185 & 0 & -0.013561209 & -0.0074504	& -0.018137996 & 0 & -0.022210396	& -0.015665036	& -0.020453195 & 0.180147342\\
			\text{[\nth{1} \& \nth{2} Rewarded]} &   &	& & & & & & & & & & &\\\midrule
			Privacy &0	& 0.017219578	& 0.074057816	& 0.045395669 & 0 & -0.031870599 & -0.061001711	& -0.107845152 & 0 & -0.039576448	& -0.04979189	& -0.069405228 & 0.31049363\\
			\text{[Intrinsic $-$ \nth{1} Rewarded]} &   &	& & & & & & & & & & &\\\midrule
			Privacy &0	& 0.00551831	& 0.063663104	& 0.075006065 & 0 & -0.032680261 & -0.043267278	& -0.132098377 & 0 & -0.124639644	& -0.072095871	& -0.066009771 & 0.344234829\\
			\text{[Intrinsic $-$ \nth{2} Rewarded]} &   &	& & & & & & & & & & &\\\midrule
			Privacy &0	& -0.011701269	& -0.010394712	& 0.029610396& 0 & -0.000809662 & 0.017734433	& -0.024253226 & 0 & -0.085063196	& -0.022303981	& 0.003395458 & 0.033741198\\
			\text{[\nth{1} $-$ \nth{2} Rewarded]} &   &	& & & & & & & & & & &\\\midrule
			Privacy &0	& 0.003576389	& 0.037118056	& 0.023663194 & 0 & -0.002256944 & -0.024444444 & -0.061510417 & 0 & -0.025625	& -0.045746528	& -0.047048611 & 0.555911458\\
			\text{[Coordinated, very low]} &   &	& & & & & & & & & & &\\\midrule
			Privacy &0	& 0.012326389	& -0.006076389	& -0.009027778 & 0 & 0.007118056 & 0.005034722 & 0.014930556 & 0 & -0.000868056	& -0.008680556	& 0.001215278 & 0.489479167\\
			\text{[Coordinated, very high]} &   &	& & & & & & & & & & &\\\bottomrule
		\end{tabular}
	}
\end{table}

\begin{table}[!htb]
	\caption{Conjoint analysis based on nine multiple linear regression models, each with a different dependent variable of privacy or rewards. The four statistically more powerful models ($R^{2}$>0.8) are illustrated in the main paper. The table also shows all calculated partworth utilities (relative importance [\%]) and their significance.}\label{tab:conjoint-results}
	\centering
	\resizebox{\columnwidth}{!}{%
		\begin{tabular}{lllllllllllllllllll}\toprule
			Models & Regression statistics &  & Conjoint statistics & Sensor & Collector & Context & acc & lig & noi & gps & cor & ngo & gov & edu & soc & env & tra & hea \\\midrule
			& Multiple $R$: & 0.93 & $\criterionUtility{\criterion}$& 27.99	& 40.14	& 31.88 & &&&&&&&&&&& \\
			Privacy & $R^{2}$: & 0.86 & $\elementUtilityPerCriterion{\criterion}{\criterionDimensionIndex}$ & & &&-12.37	& -4.73	& 15.61	& 1.48 & 16.57 & 9.20 & -2.20	& -23.57 & 21.83 & -5.27	& -10.05	& -6.50 \\
			\text{[Intrinsic]} & Adjusted $R^{2}$: & 0.84 & $\elementUtility{\criterion}{\criterionDimensionIndex}$& & &&8.67  &	16.32	& 36.66	& 22.53	& 8.67	& 1.30	& -10.10	& -31.46	& 8.67	& -18.42	& -23.20	& -19.65 \\
			& ANOVA $p$-value: & $\num{5.43e-20}$ &  $p$-value:& & &&0	& 0.031	& \num{6.67e-11}	& \num{1.86e-04}	& 0	& 0.037	& \num{1.34e-06}	& \num{2.55e-16}	& 0	& \num{1.73e-10}	& \num{1.07e-12}	& \num{4.62e-11} \\\midrule
			& Multiple $R$: & 0.78 & $\criterionUtility{\criterion}$& 16.90	& 28.91	& 54.19 &&&&&&&&&&&& \\
			Privacy& $R^{2}$: & 0.61 & $\elementUtilityPerCriterion{\criterion}{\criterionDimensionIndex}$& & &&-5.00	& 2.30	& 9.80	& -7.10	& 1.92	& 11.38	& 4.23	& -17.53	& 31.05	& -17.99	& -23.14	& 10.07 \\
			\text{[\nth{1} Rewarded]} & Adjusted $R^{2}$: & 0.55 & $\elementUtility{\criterion}{\criterionDimensionIndex}$& & &&9.33	& 16.62	& 24.12	& 7.22	& 9.33	& 18.78	& 11.64	& -10.13	& 9.33	& -39.71	& -44.86	& -11.65 \\
			& ANOVA $p$-value: & $\num{1.47e-08}$ &  $p$-value:& & &&0	& 0.348	& 0.060	& 0.786	& 0	& 0.225	& 0.765	& 0.015	& 0	& \num{4.40e-08}	& \num{3.63e-09}	& 0.009 \\\midrule
			& Multiple $R$: & 0.85 & $\criterionUtility{\criterion}$& 37.52	& 16.96	& 45.52&&&&&&&&&&&& \\
			Privacy& $R^{2}$: & 0.73 & $\elementUtilityPerCriterion{\criterion}{\criterionDimensionIndex}$& & &&-1.85	& 10.59	& 14.39	& -23.13	& -0.04	& 6.41	& -10.55	& 4.17	& 1.86	& 28.60	& -16.93	& -13.53 \\
			\text{[\nth{2} Rewarded]}& Adjusted $R^{2}$: & 0.68 & $\elementUtility{\criterion}{\criterionDimensionIndex}$& & &&-0.01	& 12.43	& 16.23	& -21.28	& -0.01	& 6.44	& -10.52	& 4.20	& -0.01	& 26.73	& -18.79	& -15.39 \\
			& ANOVA $p$-value: & $\num{2.77e-12}$ &  $p$-value:& & &&0	& 0.033	& 0.006	& \num{4.59e-04}	& 0	& 0.263	& 0.071	& 0.464	& 0	& \num{1.90e-05}	& 0.002	& 0.009 \\\midrule		
			& Multiple $R$: & 0.97 & $\criterionUtility{\criterion}$& 45.40	& 24.55	& 30.06&&&&&&&&&&&&  \\											
			Rewards & $R^{2}$: & 0.93 & $\elementUtilityPerCriterion{\criterion}{\criterionDimensionIndex}$& & &&-13.13	& -13.15 & -5.96	& 32.25	& 13.25	& -5.11	& 3.16	& -11.30	& 19.73	& -10.32	& -1.47	& -7.95 \\											
			\text{[\nth{1} \& \nth{2} Rewarded]}& Adjusted $R^{2}$: & 0.92 & $\elementUtility{\criterion}{\criterionDimensionIndex}$& & &&6.62	& 6.60	& 13.79	& 52.00	& 6.62	& -11.74	& -3.47	& -17.93	& 6.62	& -23.44	& -14.58	& -21.06 \\								
			& ANOVA $p$-value: & $\num{3.47e-28}$ &  $p$-value:& & &&0	& 0.995	& \num{6.19e-03} & \num{1.62e-24}	& 0	& \num{1.37e-09}	& \num{1.90e-04}	& \num{1.63e-13}	& 0	& \num{8.77e-17}	& \num{2.05e-11}	& \num{2.10e-15}\\\midrule
			& Multiple $R$: & 0.85 & $\criterionUtility{\criterion}$& 29.47	& 42.91	& 27.62 &&&&&&&&&&&& \\
			Privacy & $R^{2}$: & 0.72 & $\elementUtilityPerCriterion{\criterion}{\criterionDimensionIndex}$& & &&-13.60	& -6.74	& 15.87	& 4.47	& 19.97	& 7.29	& -4.31	& -22.95	& 15.79	& 0.05	& -4.02	& -11.82 \\
			\text{[Intrinsic $-$ \nth{1} Rewarded]} & Adjusted $R^{2}$: & 0.68 & $\elementUtility{\criterion}{\criterionDimensionIndex}$& & &&7.39	& 14.24	& 36.86	& 25.45	& 7.39	& -5.29	& -16.89	& -35.52	& 7.39	& -8.36	& -12.42	& -20.23 \\
			& ANOVA $p$-value: & $\num{3.65e-12}$ &  $p$-value:& & &&0	& 0.194	& \num{5.92e-07}	& 0.001	& 0 & 0.018	& \num{2.09e-05}	& \num{3.99e-11}	& 0	& 0.004	& \num{3.62e-04}	& \num{2.15e-06} \\\midrule
			& Multiple $R$: & 0.91 & $\criterionUtility{\criterion}$& 22.61	& 39.82	& 37.57&&&&&&&&&&&& \\
			Privacy& $R^{2}$: & 0.82 & $\elementUtilityPerCriterion{\criterion}{\criterionDimensionIndex}$& & &&-10.87	& -9.20	& 8.32	& 11.74	& 15.68	& 5.83	& 2.64	& -24.14	& 19.80	& -17.77	& -1.93	& -0.10 \\
			\text{[Intrinsic $-$ \nth{2} Rewarded]}& Adjusted $R^{2}$: & 0.79 & $\elementUtility{\criterion}{\criterionDimensionIndex}$& & &&8.20	& 9.87	& 27.39	& 30.81	& 8.20	& -1.65	& -4.84	& -31.62	& 8.20	& -29.37	& -13.53	& -11.69 \\
			& ANOVA $p$-value: & $\num{4.95e-17}$ &  $p$-value:& & &&0	& 0.681	& \num{1.47e-05}	& \num{7.10e-07}	& 0	& 0.018	& 0.002	& \num{1.04e-13}	& 0	& \num{7.65e-13}	& \num{1.57e-06}	& \num{7.97e-06} \\\midrule
			& Multiple $R$: & 0.83 & $\criterionUtility{\criterion}$& 24.05	& 24.45	& 51.50&&&&&&&&&&&& \\
			Privacy & $R^{2}$: & 0.69 & $\elementUtilityPerCriterion{\criterion}{\criterionDimensionIndex}$& & &&-1.09	& -7.91	& -7.15	& 16.15	& 1.07	& 0.60	& 11.39	& -13.05	& 15.13	& -34.39	& 2.15	& 17.11 \\
			\text{[\nth{1} $-$ \nth{2} Rewarded]}& Adjusted $R^{2}$: & 0.63 & $\elementUtility{\criterion}{\criterionDimensionIndex}$& & &&5.04	& -1.78	& -1.02	& 22.28	& 5.04	& 4.56	& 15.36	& -9.09	& 5.04	& -44.49	& -7.95	& 7.01 \\
			& ANOVA $p$-value: & $\num{8.70e-11}$ &  $p$-value:& & &&0	& 0.289	& 0.346	& 0.009	& 0	& 0.941	& 0.110	& 0.031	& 0	& \num{2.21e-10}	& 0.046	& 0.757 \\\midrule
			& Multiple $R$: & 0.92 & $\criterionUtility{\criterion}$& 25.48	& 42.22	& 32.30&&&&&&&&&&&& \\
			Privacy& $R^{2}$: & 0.84 & $\elementUtilityPerCriterion{\criterion}{\criterionDimensionIndex}$& & &&-11.04	& -8.59	& 14.44	& 5.20	& 15.14	& 13.59	& -1.64	& -27.09	& 20.32	& 2.73	& -11.08	& -11.97 \\
			\text{[Coordinated, very low]}& Adjusted $R^{2}$: & 0.81 & $\elementUtility{\criterion}{\criterionDimensionIndex}$& & &&8.14	& 10.59	& 33.62	& 24.38	& 8.14	& 6.59	& -8.64	& -34.09	& 8.14	& -9.45	& -23.26	& -24.16 \\
			& ANOVA $p$-value: & $\num{2.52e-18}$ &  $p$-value:& & &&0	& 0.544	& \num{4.77e-08}	& \num{1.68e-04}	& 0	& 0.701	& \num{1.08e-04}	& \num{1.13e-14}	& 0	& \num{5.51e-05}	& \num{1.94e-10}	& \num{8.48e-11} \\\midrule
			& Multiple $R$: & 0.53 & $\criterionUtility{\criterion}$& 46.24	& 32.33	& 21.43 &&&&&&&&&&&& \\
			Privacy& $R^{2}$: & 0.28 & $\elementUtilityPerCriterion{\criterion}{\criterionDimensionIndex}$& & &&1.50	& 28.20	& -11.65	& -18.05	& -14.66	& 0.75	& -3.76	& 17.67	& 4.51	& 2.63	& -14.29	& 7.14 \\
			\text{[Coordinated, very high]}& Adjusted $R^{2}$: & 0.16 & $\elementUtility{\criterion}{\criterionDimensionIndex}$& & &&-2.88	& 23.81	& -16.04	& -22.43	& -2.88	& 12.53	& 8.02	& 29.45	& -2.88	& -4.76	& -21.68	& -0.25 \\
			& ANOVA $p$-value: & $\num{2.51e-02}$ &  $p$-value:& & &&0	& 0.062	& 0.352	& 0.169	& 0	& 0.276	& 0.440	& 0.025	& 0	& 0.894	& 0.185	& 0.852 \\\bottomrule
		\end{tabular}
	}
\end{table}

Eight models with privacy as the dependent variable as assessed: intrinsic, \nth{1}, \nth{2} rewarded, intrinsic$-$\nth{1} rewarded,  intrinsic$-$\nth{2} rewarded, \nth{1} rewarded$-$\nth{2} rewarded, coordinated for very low and very high privacy preservation. One model with rewards as the dependent variable is assessed:  \nth{1} and \nth{2} rewarded of those individuals who intent and do improve rewards as in Figure~\ref{fig:app-core} in the main paper. In addition, the following four models with the mismatch as dependent variable are assessed: intrinsic, rewarded, coordinated from very low to very high privacy preservation. As they statistically perform poorly, they are not shown in Table~\ref{tab:conjoint-results}. 

\begin{figure}[!htb]
	\centering
	\includegraphics[width=1.0\textwidth]{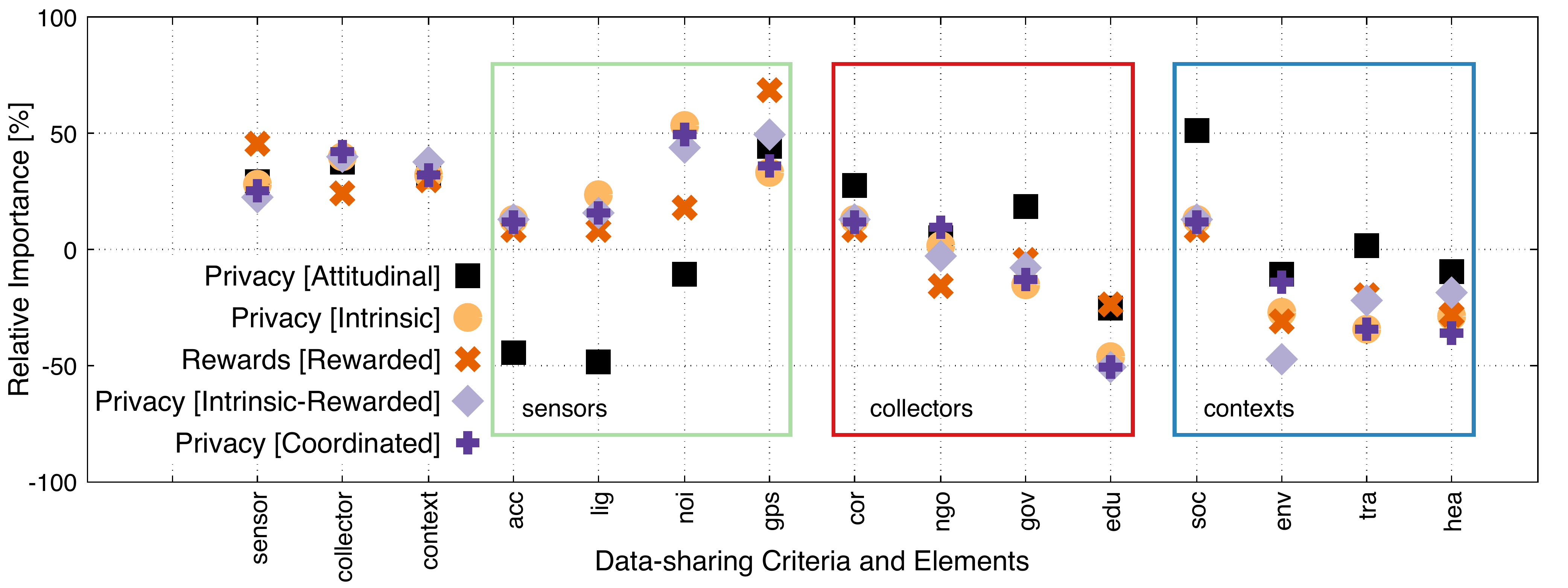}
	\caption{The relative importance (partworth utilities) of the data-sharing criteria and elements (relative among all criteria) derived from the different regression models of conjoint analysis and the perceived privacy sensitivity.}\label{fig:conjoint-analysis-per-element}
\end{figure}

Figure~\ref{fig:conjoint-analysis-per-element} illustrates the relative importance ($	\criterionUtility{\criterion}$, $
\elementUtility{\criterion}{\criterionDimensionIndex}$) of the data-sharing criteria and elements among all criteria, in contrast to Figure~\ref{fig:conjoint-analysis-per-criterion} in the main paper that shows the relative importance ($\criterionUtility{\criterion}$, $\elementUtilityPerCriterion{\criterion}{\criterionDimensionIndex}$) of the elements within each criterion. 

The relative importance ($\criterionUtility{\criterion}$) of the data-sharing criteria is the same as shown in Figure~\ref{fig:conjoint-analysis-per-criterion} of the main paper. For all models, sensor data such as GPS (46.82\%), noise (41.4\%) and light (16.04\%) show the highest mean positive relative importance among all elements of the three criteria, while education (-42.74\%) from collectors and environment (-29.78\%), health (-27.8\%) and transportation (-27.35\%) from context show the lowest one. In contrast to the regression models, the perceived privacy sensitivity of GPS (44.37\%), corporation (27.8\%) and social networking (51.44\%) are the highest positive ones, while the accelerometer (-44.28\%), light (-48.56\%) and education (-25.51\%) show the highest negative ones.

\section{Validation of Groups}\label{sec:grouping}

Table~\ref{tab:bootstrap} illustrates the results of the bootstrap evaluation method for the 5 different group behaviors extracted from the experimental data. 

\begin{table}[!htb]
	\caption{Results of the bootstrap evaluation method~\cite{Hennig2007} (\texttt{clusterboot} of R) for the stability of the clusters. Each entry of results is represented as `bootmean (bootbrd)', where bootmean is the clusterwise mean Jaccard similaritiy and bootbrd is the clusterwise number of times a cluster is disolved.}\label{tab:bootstrap}
	\centering
	\begin{tabular}{llll}\toprule
		\multicolumn{1}{r}{Clustering algorithms} & k-means  & hierachical & pamkCBI \\\midrule
		Privacy ignorants & 0.79 (8)  & 0.67 (41) & 0.58 (48) \\
		Privacy neutrals & 0.93 (0)  & 0.88 (1) & 0.7 (31) \\
		Privacy preservers & 0.89 (7)  & 0.76 (16) & 0.7 (31) \\
		Rewards seekers &  0.83 (1) & 0.75 (17) & 0.61 (37)  \\
		Rewards opportunists & 0.84 (6)  & 0.76 (14) & 0.56 (51) \\\bottomrule
	\end{tabular}
\end{table}


Furthermore, the split of the participants over the data-sharing groups is compared to privacy categories identified in the general population from studies such as the ones of Westin~\cite{Westin1991,Kumaraguru2005}. This comparison can only be indicative though: a random sample from a US population back in 1990 is compared to a non-random sample from a Swiss population in 2016. Moreover, the survey questions are not identical to the formulated data-sharing prompts. Nevertheless, this comparison has a value out of the the fact that there are groups that capture the intended privacy of a broader population vs. groups that capture the actual data-sharing decisions of typical smartphone users. 

Westin's studies classify individuals in three behavioral categories based on survey responses: \emph{privacy fundamentalists, pragmatists and unconcerned}. They cover the whole spectrum of data-sharing levels depicted in the exemplary of Table~\ref{tab:groups} in the main paper. Based on this, we match the data-sharing groups to Westin's categories under intrinsic data sharing, i.e. the data-sharing behavior of individuals is not considered under rewarded data sharing. The matching is illustrated in Table~\ref{tab:groups-population}. The observed groups sizes show a remarkable match to Westin's privacy categories.

\begin{table}[!htb]
	\caption{Matching the Westin's classification~\cite{Westin1991,Kumaraguru2005} to the data-sharing groups without rewards (Table~\ref{tab:groups} in this main paper).}\label{tab:groups-population}
	\centering
		\begin{tabular}{llll}\toprule
			\multicolumn{2}{l}{Westin's population categories~\cite{Westin1991,Kumaraguru2005}} & \multicolumn{2}{l}{Data-sharing Groups ($n=84$).}   \\\midrule
			\multirow{2}{*}{Privacy fundamentalists} & \multirow{2}{*}{25\%} & Privacy preservers & \multirow{2}{*}{26.2\%} \\
			 &  & Reward opportunists &  \\\midrule
			\multirow{2}{*}{Privacy pragmatists} & \multirow{2}{*}{57\%} & Privacy neutrals & \multirow{2}{*}{57.14\%} \\
			&  & Reward seekers &  \\\midrule 
			Privacy unconcerned & 18\% & Privacy ignorants & 16.7\% \\\bottomrule
		\end{tabular}
\end{table}

\section{Analysis of Variance for Data-sharing Criteria and Groups}\label{sec:ANOVA}

\begin{figure}[!htb]
	\centering
	\includegraphics[width=0.8\textwidth]{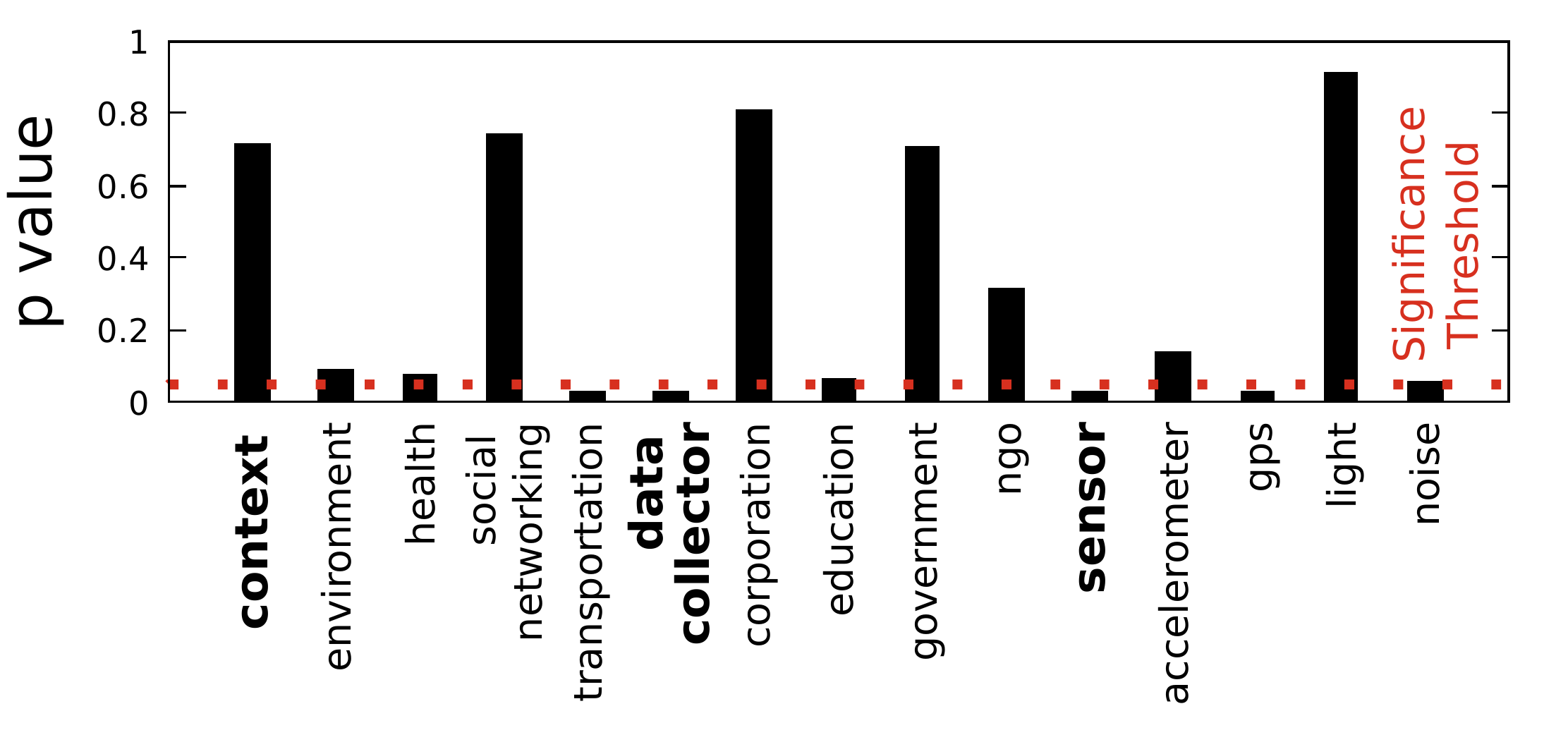}
	\caption{Statistical significance ($p$ values) of the different data-sharing criteria and their elements to explain the 5 group behaviors in data sharing. Data collectors and sensors are significant ($p<0.05$ values), in particular the GPS, as well as the transportation context. Moreover, the following data-sharing elements fall close to the significance threshold: environment, and education contexts, accelerometer and noise sensors, and the educational institutes as data collector.}\label{fig:criteria-significance}
\end{figure}

The Analysis of Variance (ANOVA) is made with IBM SPSS 24.0. Figure~\ref{fig:criteria-significance} summarizes the $p$ values obtained for each data-sharing criterion and its elements. Using the Levene's test, the homogeneity of variances is confirmed ($p>0.05$) for the majority of the data-sharing criteria and their elements: sensor ($p=0.169$), data collector ($p=0.328$), context ($p=0.956$), GPS ($p=0.156$), light ($p=0.896$), noise ($p=0.432$), corporation ($p=0.607$), educational institute ($p=0.35$), government ($p=0.074$), NGO ($p=0.993$), health ($p=0.314$), social networking ($p=0.486$). It is not confirmed for: accelerometer ($p=0.04$), transportation ($p=0.039$) and environment ($p=0.005$). The whole report analysis is illustrated in Table~\ref{tab:ANOVA}. 

\begin{table}[!htb]
	\caption{One-way ANOVA report analysis. Dependent variables: Questions B.9-B12 in Table~\ref{table:entry-phase}. Independent variable: The privacy change when groups' data-sharing choices are rewarded.}\label{tab:ANOVA}
	\centering
	\resizebox{\columnwidth}{!}{%
		\begin{tabular}{llllllll}\toprule
			\multicolumn{2}{l}{Data Sharing Criterion} & & Sum of Squares & Degrees of Freedom ($df$) & Mean Squares & F-test & Significance \\\midrule
			\multirow{3}{*}{Sensors} &  & Between  & 10.745& 4 & 2.686 & 2.756 & 0.031 \\
			&  & Within & 107.202 & 110 & 0.975 &  &  \\
			&  & Total & 117.948 & 114 &  &  &  \\\midrule
			&  \multirow{3}{*}{Accelerometer} & Between  & 6.083 & 4 & 1.521 & 1.774 & 0.139 \\
			&  & Within & 94.317 & 110 & 0.857 &  &  \\
			&  & Total & 100.4 & 114 &  &  &  \\\midrule
			&  \multirow{3}{*}{Location} & Between  & 8.805 & 4 & 2.201 & 2.719 & 0.033 \\
			&  & Within & 89.056 & 110 & 0.81 &  &  \\
			&  & Total & 97.861 & 114 &  &  &  \\\midrule
			&  \multirow{3}{*}{Light} & Between  & 0.838 & 4 & 0.21 & 0.241 & 0.914 \\
			&  & Within & 95.509 & 110 & 0.868 &  &  \\
			&  & Total & 96.348 & 114 &  &  &  \\\midrule
			&  \multirow{3}{*}{Noise} & Between  & 14.293 & 4 & 3.573 & 2.384 & 0.056 \\
			&  & Within & 164.873 & 110 & 1.499 &  &  \\
			&  & Total & 179.165 & 114 &  &  &  \\\midrule
			\multirow{3}{*}{Data collectors} &  & Between  & 9.85 & 4 & 2.463 & 2.862 & 0.027 \\
			&  & Within & 94.637 & 110 & 0.86 &  &  \\
			&  & Total & 104.487 & 114 &  &  &  \\\midrule
			&  \multirow{3}{*}{Corporations} & Between  & 2.019 & 4 & 0.505 & 0.399 & 0.809 \\
			&  & Within & 140.559 & 111 & 1.266 &  &  \\
			&  & Total & 142.578 & 115 &  &  &  \\\midrule
			&  \multirow{3}{*}{NGOs} & Between  & 5.426 & 4 & 1.356 & 1.199 & 0.315 \\
			&  & Within & 125.566 & 111 & 1.131 &  &  \\
			&  & Total & 130.991 & 115 &  &  &  \\\midrule
			&  \multirow{3}{*}{Goverments} & Between  & 3.322 & 4 & 0.831 & 0.534 & 0.711 \\
			&  & Within & 172.566 & 111 & 1.555 &  &  \\
			&  & Total & 175.888 & 115 &  &  &  \\\midrule
			&  \multirow{3}{*}{Educational institutes} & Between  & 7.881 & 4 & 1.97 & 2.27 & 0.066 \\
			&  & Within & 96.36 & 111 & 0.868 &  &  \\
			&  & Total & 104.241 & 115 &  &  &  \\\midrule
			\multirow{3}{*}{Context/purpose} &  & Between  & 2.056 & 4 & 0.514 & 0.532 & 0.712 \\
			&  & Within & 106.24 & 110 & 0.966 &  &  \\
			&  & Total & 108.296 & 114 &  &  &  \\\midrule
			&  \multirow{3}{*}{Health/fitness} & Between  & 12.013 & 4 & 3.003 & 2.158 & 0.078 \\
			&  & Within & 154.496 & 111 & 1.392 &  &  \\
			&  & Total & 166.509 & 115 &  &  &  \\\midrule
			&  \multirow{3}{*}{Social networking} & Between  & 1.468 & 4 & 0.367 & 0.495 & 0.74 \\
			&  & Within & 82.325 & 111 & 0.742 &  &  \\
			&  & Total & 83.793 & 115 &  &  &  \\\midrule
			&  \multirow{3}{*}{Environment} & Between  & 7.606 & 4 & 1.901 & 2.08 & 0.088 \\
			&  & Within & 101.455 & 111 & 0.914 &  &  \\
			&  & Total & 109.06 & 115 &  &  &  \\\midrule
			&  \multirow{3}{*}{Transport} & Between  & 12.589 & 4 & 3.147 & 2.779 & 0.03 \\
			&  & Within & 125.713 & 111 & 1.133 &  &  \\
			&  & Total & 138.302 & 115 &  &  &  \\
			\bottomrule
		\end{tabular}
	}
\end{table}

The report analysis of the post hoc Tukey's range test ($\alpha=0.05$) is illustrated in Table~\ref{tab:ANOVA-posthoc-sensors},~\ref{tab:ANOVA-posthoc-data collectors} and~\ref{tab:ANOVA-posthoc-context}. 

\begin{table}[!htb]
	\caption{Post hoc Tukey's range test ($\alpha=0.05$) on sensors explaining the privacy change when groups' data-sharing choices are rewarded.}\label{tab:ANOVA-posthoc-sensors}
	\centering
	\resizebox{\columnwidth}{!}{%
		\begin{tabular}{llllllll}\toprule
			& &  &   &  &  & \multicolumn{2}{c}{95\% Confidence Interval} \\
			\multicolumn{2}{l}{Data Sharing Criterion} & & Mean Group Differences &  Standard Deviation Error & Significance & Lower Bound & Upper Bound \\\midrule
			\multirow{10}{*}{Sensors} &  & reward seekers - privacy ignorants & 0.058& 0.301 & 1.0 & -0.78 & 0.89 \\
			&  & privacy neutrals - privacy ignorants  & 0.779 & 0.314 & 0.102 & -0.09 & 1.65 \\
			&  & privacy neutrals - reward seekers & 0.721 & 0.243 & 0.03 & 0.05 & 1.4 \\
			&  & privacy preservers - privacy ignorants  & 0.511 & 0.416 & 0.735 & -0.64 & 1.67 \\
			&  & privacy preservers - reward seekers & 0.453& 0.366 & 0.729 & -0.56 & 1.47 \\
			&  & privacy preservers - privacy neutrals  & -0.268 & 0.377 & 0.953 & -1.31 & 0.78 \\
			&  & reward opportunists - privacy ignorants & 0.233 & 0.325 & 0.952 & -0.67 & 1.13 \\
			&  & reward opportunists - reward seekers  & 0.175 & 0.257 & 0.96 & -0.54 & 0.89 \\
			&  & reward opportunists - privacy neutrals  & -0.546 & 0.272 & 0.271 & -1.3 & 0.21 \\
			&  & reward opportunists - privacy preservers & -0.278 & 0.386 & 0.952 & -1.35 & 0.79 \\\midrule
			&  \multirow{10}{*}{Accelerometer} & reward seekers - privacy ignorants  & 0.438 & 0.283 & 0.536 & -0.35 & 1.22 \\
			&  & privacy neutrals - privacy ignorants  & 0.703 & 0.294 & 0.126 & -0.11 & 1.52 \\
			&  & privacy neutrals - reward seekers & 0.266 & 0.23 & 0.776 & -0.37 & 0.9 \\
			&  & privacy preservers - privacy ignorants  & 0.1& 0.378 & 0.999 & -0.95 & 1.15 \\
			&  & privacy preservers - reward seekers & -0.338 & 0.33 & 0.844 & -1.25 & 0.58 \\
			&  & privacy preservers - privacy neutrals  & -0.603 & 0.34 & 0.392& -1.55 & 0.34 \\
			&  & reward opportunists - privacy ignorants & 0.35 & 0.305 & 0.78 & -0.5 & 1.2 \\
			&  & reward opportunists - reward seekers  & -0.088 & 0.243 & 0.996 & -0.76 & 0.59 \\
			&  & reward opportunists - privacy neutrals  & -0.353 & 0.256 & 0.64 & -1.06 & 0.36 \\
			&  & reward opportunists - privacy preservers & 0.25 & 0.349 & 0.952 & -0.72 & 1.22 \\\midrule
			&  \multirow{10}{*}{Location} & reward seekers - privacy ignorants  & 0.733 & 0.275 & 0.066 & -0.03 & 1.5 \\
			&  & privacy neutrals - privacy ignorants  & 0.837 & 0.286 & 0.033 & 0.04 & 1.63 \\
			&  & privacy neutrals - reward seekers & 0.103 & 0.223 & 0.99 & -0.52 & 0.72 \\
			&  & privacy preservers - privacy ignorants  & 0.933 & 0.367 & 0.089 & -0.09 & 1.95 \\
			&  & privacy preservers - reward seekers & 0.2 & 0.321 & 0.971 & -0.69 & 1.09 \\
			&  & privacy preservers - privacy neutrals  & 0.097 & 0.33 & 0.998 & -0.82 & 1.01 \\
			&  & reward opportunists - privacy ignorants & 0.817 & 0.296 & 0.052 & 0.0 & 1.64 \\
			&  & reward opportunists - reward seekers  & 0.083 & 0.236 & 0.997 & -0.57 & 0.74 \\
			&  & reward opportunists - privacy neutrals  & -0.02 & 0.248 & 1.0 & -0.71 & 0.67 \\
			&  & reward opportunists - privacy preservers & -0.117 & 0.339 & 0.997 & -1.06 & 0.82 \\\midrule
			&  \multirow{10}{*}{Light} & reward seekers - privacy ignorants  & 0.023 & 0.285 & 1.0 & -0.77 & 0.81 \\
			&  & privacy neutrals - privacy ignorants  & -0.044 & 0.296 & 1.0 & -0.87 & 0.78 \\
			&  & privacy neutrals - reward seekers & -0.067 & 0.231 & 0.998 & -0.71 & 0.57 \\
			&  & privacy preservers - privacy ignorants  & 0.067 & 0.38 & 1.0 & -0.99 & 1.12 \\
			&  & privacy preservers - reward seekers & 0.043 & 0.332 & 1.0 & -0.88 & 0.96 \\
			&  & privacy preservers - privacy neutrals  & 0.11 & 0.342 & 0.998 & -0.84 & 1.06 \\
			&  & reward opportunists - privacy ignorants & -0.192 & 0.307 & 0.971 & -1.04 & 0.66 \\
			&  & reward opportunists - reward seekers  & -0.215 & 0.244 & 0.903 & -0.89 & 0.46 \\
			&  & reward opportunists - privacy neutrals  & -0.148 & 0.257 & 0.978 & -0.86 & 0.57 \\
			&  & reward opportunists - privacy preservers & -0.258 & 0.351 & 0.947 & -1.23 & 0.71 \\\midrule
			&  \multirow{10}{*}{Noise} & reward seekers - privacy ignorants  & 0.272& 0.375 & 0.95 & -0.77 & 1.31 \\
			&  & privacy neutrals - privacy ignorants  & 0.798 & 0.389 & 0.25 & -0.28 & 1.88 \\
			&  & privacy neutrals - reward seekers & 0.526 & 0.304 & 0.419 & -0.32 & 1.37 \\
			&  & privacy preservers - privacy ignorants  & 1.267 & 0.5 & 0.09 & -0.12 & 2.65 \\
			&  & privacy preservers - reward seekers & 0.995 & 0.436 & 0.159 & -0.22 & 2.2 \\
			&  & privacy preservers - privacy neutrals  & 0.469 & 0.449 & 0.834 & -0.78 & 1.71 \\
			&  & reward opportunists - privacy ignorants & 0.408 & 0.403 & 0.849 & -0.71 & 1.53 \\
			&  & reward opportunists - reward seekers  & 0.136 & 0.321 & 0.993 & -0.75 & 1.03 \\
			&  & reward opportunists - privacy neutrals  & -0.389 & 0.338 & 0.778 & -1.33 & 0.55 \\
			&  & reward opportunists - privacy preservers & -0.858 & 0.461 & 0.344 & -2.14 & 0.42 \\\bottomrule
		\end{tabular}
	}
\end{table}

\begin{table}[!htb]
	\caption{Post hoc Tukey's range test ($\alpha=0.05$) on data collectors explaining the privacy change when groups' data-sharing choices are rewarded.}\label{tab:ANOVA-posthoc-data collectors}
	\centering
	\resizebox{\columnwidth}{!}{%
		\begin{tabular}{llllllll}\toprule
			& &  &   &  &  & \multicolumn{2}{c}{95\% Confidence Interval} \\
			\multicolumn{2}{l}{Data Sharing Criterion} & & Sum of Squares & Degrees of Freedom ($df$) & Mean Squares & F-test & Significance \\\midrule
			\multirow{10}{*}{Data collectors} &  & reward seekers - privacy ignorants  & 0.502 & 0.283 & 0.394 & -0.28 & 1.29 \\
			&  & privacy neutrals - privacy ignorants  & 0.582 & 0.295 & 0.287 & -0.24 & 1.4 \\
			&  & privacy neutrals - reward seekers & 0.08 & 0.229 & 0.997 & -0.55 & 0.71 \\
			&  & privacy preservers - privacy ignorants  & 0.911 & 0.391 & 0.143 & -0.17 & 2.0 \\
			&  & privacy preservers - reward seekers & 0.409 & 0.344 & 0.757 & -0.54 & 1.36 \\
			&  & privacy preservers - privacy neutrals  & 0.33 & 0.354 & 0.884 & -0.65 & -1.31 \\
			&  & reward opportunists - privacy ignorants & 0.967 & 0.305 & 0.017 & 0.12 & 1.81 \\
			&  & reward opportunists - reward seekers  & 0.465& 0.242 & 0.312 & -0.21 & 1.14 \\
			&  & reward opportunists - privacy neutrals  & 0.385 & 0.256 & 0.562 & -0.32 & 1.09 \\
			&  & reward opportunists - privacy preservers & 0.056 & 0.363 & 1.0 & -0.95 & 1.06 \\\midrule
			&  \multirow{10}{*}{Corporations} & reward seekers - privacy ignorants   & 0.286 & 0.343 & 0.92 & -0.67 & 1.24 \\
			&  & privacy neutrals - privacy ignorants  & 0.285 & 0.358 & 0.931 & -0.71 & 1,28 \\
			&  & privacy neutrals - reward seekers & -0.001 & 0.277 & 1.0 & -0.77 & 0.77 \\
			&  & privacy preservers - privacy ignorants  & 0.433 & 0.459 & 0.879 & -0.84 & 1.71 \\
			&  & privacy preservers - reward seekers & 0.147 & 0.4 & 0.996 & -0.96 & 1.26 \\
			&  & privacy preservers - privacy neutrals  & 0.148 & 0.413 & 0.996 & -1.0 & 1.29 \\
			&  & reward opportunists - privacy ignorants & 0.442 & 0.37 & 0.756 & -0.59 & 1.47 \\
			&  & reward opportunists - reward seekers  & 0.156 & 0.293 & 0.984 & -0.66 & 0.97 \\
			&  & reward opportunists - privacy neutrals  & 0.157 & 0.311 & 0.987 & -0.7 & 1.02 \\
			&  & reward opportunists - privacy preservers & 0.008 & 0.424 & 1.0 & -1.17 & 1.18 \\\midrule
			&  \multirow{3}{*}{NGOs} & reward seekers - privacy ignorants  & 0.374 & 0.324 & 0.778 & -0.53 & 1.27 \\
			&  & privacy neutrals - privacy ignorants  & 0.607 & 0.338 & 0.382 & -0.33 & 1.54 \\
			&  & privacy neutrals - reward seekers & 0.233 & 0.262 & 0.9 & -0.49 & 0.96 \\
			&  & privacy preservers - privacy ignorants  & 0.8 & 0.434 & 0.355 & -0.4 & 2.0 \\
			&  & privacy preservers - reward seekers & 0.426 & 0.378 & 0.792 & -0.62 & 1.47 \\
			&  & privacy preservers - privacy neutrals  & 0.193 & 0.39 & 0.988 & -0.89 & 1.27\\
			&  & reward opportunists - privacy ignorants & 0.317& 0.35 & 0.895 & -0.65 & 1.29 \\
			&  & reward opportunists - reward seekers  & -0.057 & 0.277 & 1.0 & -0.83 & 0.71 \\
			&  & reward opportunists - privacy neutrals  & -0.29 & 0.293 & 0.86 & -1.1 & 0.52 \\
			&  & reward opportunists - privacy preservers & -0.483 & 0.4 & 0.747 & -1.59 & 0.63 \\\midrule
			&  \multirow{3}{*}{Goverments} & reward seekers - privacy ignorants  & 0.263 & 0.38 & 0.958 & -0.79 & 1.32 \\
			&  & privacy neutrals - privacy ignorants  & 0.552 & 0.397 & 0.635 & -0.55 & 1.65 \\
			&  & privacy neutrals - reward seekers & 0.289 & 0.307 & 0.881 & -0.56 & 1.14 \\
			&  & privacy preservers - privacy ignorants  & 0.4 & 0.509 & 0.934 & -1.01 & 1.81 \\
			&  & privacy preservers - reward seekers & 0.137 & 0.443 & 0.998 & -1.09 & 1.37 \\
			&  & privacy preservers - privacy neutrals  & -0.152 & 0.457 & 0.997 & -1.42 & 1.12 \\
			&  & reward opportunists - privacy ignorants & 0.375 & 0.41 & 0.891 & -0.76 & 1.51 \\
			&  & reward opportunists - reward seekers  & 0.112 & 0.325 & 0.997 & -0.79 & 1.01 \\
			&  & reward opportunists - privacy neutrals  & -0.177 & 0.344 & 0.986 & -1.13 & 0.78 \\
			&  & reward opportunists - privacy preservers & -0.025 & 0.496 & 1.0 & -1.33 & 1.28 \\\midrule
			&  \multirow{3}{*}{Educational institutes} & reward seekers - privacy ignorants  & 0.528 & 0.284 & 0.346 & -0.26 & 1.32 \\
			&  & privacy neutrals - privacy ignorants  & 0.237 & 0.296 & 0.93 & -0.58 & 1.06 \\
			&  & privacy neutrals - reward seekers & -0.291 & 0.23 & 0.711 & -0.93 & 0.35 \\
			&  & privacy preservers - privacy ignorants  & 1.033 & 0.38 & 0.058 & -0.02 & 2.09 \\
			&  & privacy preservers - reward seekers & 0.505 & 0.331 & 0.548 & -0.41 & 1.42 \\
			&  & privacy preservers - privacy neutrals  & 0.797 & 0.342 & 0.143 & -0.15 & 1.74 \\
			&  & reward opportunists - privacy ignorants & 0.342 & 0.307 & 0.799 & -0.51 & 1.19 \\
			&  & reward opportunists - reward seekers  & -0.186 & 0.243 & 0.939 & -0.86 & 0.49 \\
			&  & reward opportunists - privacy neutrals  & 0.105 & 0.257 & 0.994 & -0.61 & 0.82 \\
			&  & reward opportunists - privacy preservers & -0.692 & 0.351 & 0.286 & -1.66 & 0.28 \\\bottomrule
		\end{tabular}
	}
\end{table}

\begin{table}[!htb]
	\caption{Post hoc Tukey's range test ($\alpha=0.05$) on data-sharing context/purpose explaining the privacy change when groups' data-sharing choices are rewarded.}\label{tab:ANOVA-posthoc-context}
	\centering
	\resizebox{\columnwidth}{!}{%
		\begin{tabular}{llllllll}\toprule
			& &  &   &  &  & \multicolumn{2}{c}{95\% Confidence Interval} \\
			\multicolumn{2}{l}{Data Sharing Criterion} & & Sum of Squares & Degrees of Freedom ($df$) & Mean Squares & F-test & Significance \\\midrule
			\multirow{10}{*}{Context/purpose} &  & reward seekers - privacy ignorants  & 0.119 & 0.3 & 0.995 & -0.71 & 0.95 \\
			&  & privacy neutrals - privacy ignorants  & 0.17 & 0.313 & 0.982 & -0.7 & 1.04 \\
			&  & privacy neutrals - reward seekers & 0.051 & 0.242 & 1.0 & -0.62 & 0.72 \\
			&  & privacy preservers - privacy ignorants  & 0.511 & 0.414 & 0.732 & -0.64 & 1.66 \\
			&  & privacy preservers - reward seekers & 0.392 & 0.364 & 0.819 & -0.62 & 1.4 \\
			&  & privacy preservers - privacy neutrals  & 0.341 & 0.375 & 0.893 & -0.7 & 1.38 \\
			&  & reward opportunists - privacy ignorants & 0.317 & 0.323 & 0.864 & -0.58 & 1.21 \\
			&  & reward opportunists - reward seekers  & 0.197 & 0.256 & 0.939 & -0.51 & 0.91 \\
			&  & reward opportunists - privacy neutrals  & 0.147 & 0.271 & 0.983 & -0.61 & 0.9 \\
			&  & reward opportunists - privacy preservers & -0.194 & 0.384 & 0.987 & -1.26 & 0.87 \\\midrule
			&  \multirow{10}{*}{Health/fitness} & reward seekers - privacy ignorants  & 0.765 & 0.36 & 0.216 & -0.23 & 1.76 \\
			&  & privacy neutrals - privacy ignorants  & 1.064 & 0.375 & 0.042 & 0.02 & 2.1 \\
			&  & privacy neutrals - reward seekers & 0.299 & 0.291 & 0.841 & -0.51 & 1.11 \\
			&  & privacy preservers - privacy ignorants  & 0.833 & 0.482 & 0.42 & -0.5 & 2.17 \\
			&  & privacy preservers - reward seekers & 0.068 & 0.419 & 1.0 & -1.09 & 1.23 \\
			&  & privacy preservers - privacy neutrals  & -0.231 & 0.433 & 0.984 & -1.43 & 0.97 \\
			&  & reward opportunists - privacy ignorants & 0.925 & 0.388 & 0.128 & -0.15 & 2.0 \\
			&  & reward opportunists - reward seekers  & 0.16 & 0.308 & 0.985 & -0.69 & 1.01 \\
			&  & reward opportunists - privacy neutrals  & -0.139 & 0.326 & 0.993 & -1.04 & 0.76 \\
			&  & reward opportunists - privacy preservers & 0.092 & 0.444 & 1.0 & -1.14 & 1.32 \\\midrule
			&  \multirow{10}{*}{Social networking} & reward seekers - privacy ignorants  & 0.318 & 0.263 & 0.746 & -0.41 & 1.05 \\
			&  & privacy neutrals - privacy ignorants  & 0.271 & 0.274 & 0.859 & -0.49 & 1.03 \\
			&  & privacy neutrals - reward seekers & -0.46 & 0.212 & 0.999 & -0.64 & 0.54 \\
			&  & privacy preservers - privacy ignorants  & 0.433 & 0.352 & 0.732 & -0.54 & 1.41 \\
			&  & privacy preservers - reward seekers & 0.116 & 0.306 & 0.996 & -0.73 & 0.96 \\
			&  & privacy preservers - privacy neutrals  & 0.162 & 0.316 & 0.986 & -0.71 & 1.04 \\
			&  & reward opportunists - privacy ignorants & 0.3 & 0.283 & 0.827 & -0.49 & 1.09 \\
			&  & reward opportunists - reward seekers  & -0.018 & 0.225 & 1.0 & -0.64 & 0.61 \\
			&  & reward opportunists - privacy neutrals  & 0.029 & 0.238& 1.0 & -0.63 & 0.69 \\
			&  & reward opportunists - privacy preservers & -0.133 & 0.324 & 0.994 & -1.03 & 0.77 \\\midrule
			&  \multirow{10}{*}{Environment} & reward seekers - privacy ignorants  & 0.486 & 0.292 & 0.459 & -0.32 & 1.29 \\
			&  & privacy neutrals - privacy ignorants  & 0.657 & 0.304 & 0.202 & -0.19 & 1.5 \\
			&  & privacy neutrals - reward seekers & 0.172 & 0.236 & 0.95 & -0.48 & 0.83 \\
			&  & privacy preservers - privacy ignorants  & 0.933 & 0.39 & 0.125 & -0.15 & 2.02 \\
			&  & privacy preservers - reward seekers & 0.447 & 0.34 & 0.681 & -0.49 & 1.39 \\
			&  & privacy preservers - privacy neutrals  & 0.276 & 0.351 & 0.934 & -0.7 & 1.25 \\
			&  & reward opportunists - privacy ignorants & 0.767 & 0.315 & 0.113 & -0.11 & 1.64 \\
			&  & reward opportunists - reward seekers  & 0.281 & 0.249 & 0.792 & -0.41 & 0.97 \\
			&  & reward opportunists - privacy neutrals  & 0.109 & 0.264 & 0.994 & -0.62 & 0.84 \\
			&  & reward opportunists - privacy preservers & -0.167 & 0.36 & 0.99 & -1.16 & 0.83 \\\midrule
			&  \multirow{10}{*}{Transport} & reward seekers - privacy ignorants  & 0.351 & 0.325 & 0.816 & -0.55 & 1.25 \\
			&  & privacy neutrals - privacy ignorants  & 0.908 & 0.338 & 0.063 & -0.03 & 1.85 \\
			&  & privacy neutrals - reward seekers & 0.557 & 0.262 & 0.218 & -0.17 & 1.28 \\
			&  & privacy preservers - privacy ignorants  & 0.767 & 0.434 & 0.399 & -0.44 & 1.97 \\
			&  & privacy preservers - reward seekers & 0.416 & 0.378 & 0.807 & -0.63 & 1.46 \\
			&  & privacy preservers - privacy neutrals  & -0.141 & 0.39 & 0.996 & -1.22 & 0.94 \\
			&  & reward opportunists - privacy ignorants & 0.875 & 0.35 & 0.098 & -0.1 & 1.85 \\
			&  & reward opportunists - reward seekers  & 0.524 & 0.277 & 0.329 & -0.25 & 1.29 \\
			&  & reward opportunists - privacy neutrals  & -0.033 & 0.294 & 1.0 & -0.85 & 0.78 \\
			&  & reward opportunists - privacy preservers & 0.108 & 0.401 & 0.999 & -1.0 & 1.22 \\\bottomrule
		\end{tabular}
	}
\end{table}

\bibliography{ms}
%
\bibliographystyle{naturemag}
%
%
\makeatletter\@input{xx.tex}\makeatother